\documentclass[journal]{IEEEtran}

\usepackage{amsmath,amsfonts}
\usepackage{algorithm}
\usepackage{algorithmic}
\usepackage{array}
\usepackage[caption=false,font=normalsize,labelfont=sf,textfont=sf]{subfig}
\usepackage{pstool}
\usepackage{textcomp}
\usepackage{stfloats}
\usepackage{url}
\usepackage{verbatim}
\usepackage{graphicx}
\usepackage{cite}

\usepackage[dvipsnames]{xcolor}

\usepackage{booktabs}
\usepackage{makecell}
\usepackage{multirow}

\usepackage{colortbl}
\usepackage{adjustbox}

\usepackage[capitalize]{cleveref}
\usepackage{siunitx}

\newcommand*{\V}{\mathbf}

\definecolor{mybronze}{RGB}{194, 151, 44} 

\begin{document}

\title{Uni-Fusion: Universal Continuous Mapping}

\author{Yijun Yuan and Andreas N\"uchter
\thanks{The authors are with Informatics XVII -- Robotics at Julius-Maximilians-University of W\"urzburg, Germany.
  {\tt\small \{yijun.yuan|andreas. nuechter\}@uni-wuerzburg.de}~ 

This work was in parts supported by the Federal Ministry for Economic Affairs and Climate Action (BMWK) on the basis of a decision by the German Bundestag und the grant number KK5150104GM1. We also acknowledge the support by the Elite Network Bavaria (ENB) through the ``Satellite Technology'' academic program.
}
} 

\maketitle

\begin{abstract}
We present Uni-Fusion, a universal continuous mapping framework for surfaces, surface properties (color, infrared, etc.) and more (latent features in CLIP embedding space, etc.).
We propose the first universal implicit encoding model that supports encoding of both geometry and different types of properties (RGB, infrared, features, etc.) without requiring any training.
Based on this, our framework divides the point cloud into regular grid voxels and generates a latent feature in each voxel to form a Latent Implicit Map (LIM) for geometries and arbitrary properties.
Then, by fusing a local LIM frame-wisely into a global LIM, an incremental reconstruction is achieved.
Encoded with corresponding types of data, our Latent Implicit Map is capable of generating continuous surfaces, surface property fields, surface feature fields, and all other possible options. 
To demonstrate the capabilities of our model, we implement three applications:
(1) incremental reconstruction for surfaces and color 
(2) 2D-to-3D transfer of fabricated properties
(3) open-vocabulary scene understanding by creating a text CLIP feature field on surfaces. 
We evaluate Uni-Fusion by comparing it in corresponding applications, from which Uni-Fusion shows high-flexibility in various applications while performing best or being competitive.
The project page of Uni-Fusion is available at {\tt\small\protect\url{https://jarrome.github.io/Uni-Fusion/}}. 
\end{abstract}


\section{Introduction} 
\IEEEPARstart{I}{n} robotics, 3D perception plays an important role in enabling robots to interact with their surroundings.
To achieve this, robots must use different types of sensors and employ techniques such as reconstruction and scene understanding. 
These tasks require the processing of various types of information, including geometry and surface properties.
However, algorithms must be designed specifically to handle these different types of data.

Therefore, at the outset, we pose the \textbf{following question}: \textit{Is it feasible to handle all these information with a single, universal mapping model?}

Reconstruction, as one of the most prominent topics in the field, has been developing for decades. 
During this period, numerous works have pushed the limits~\cite{o2012gaussian,kim2013continuous,ghaffari2018gaussian,yuan2018fast,martens2016geometric,lee2019online,wu2021faithful,ivan2022online,curless1996volumetric,izadi2011kinectfusion,dai2017bundlefusion,huang2021di,yuan2022algorithm,sucar2021imap,zhu2022nice}. 
Reconstruction models aim to extract the zero-level surface given a set of points.
These approaches are typically based on occupancy grids and signed distance functions (SDFs).
Occupancy grids are primarily used in 2D and object-level shape reconstruction, where Gaussian Process Occupancy Maps (GPOM), Gaussian Process Implicit Surface (GPIS), and Hilbert Maps have been developed to generate continuous probabilistic occupancy maps.
\begin{figure}[t]
	\centering
	\includegraphics[width=1\linewidth]{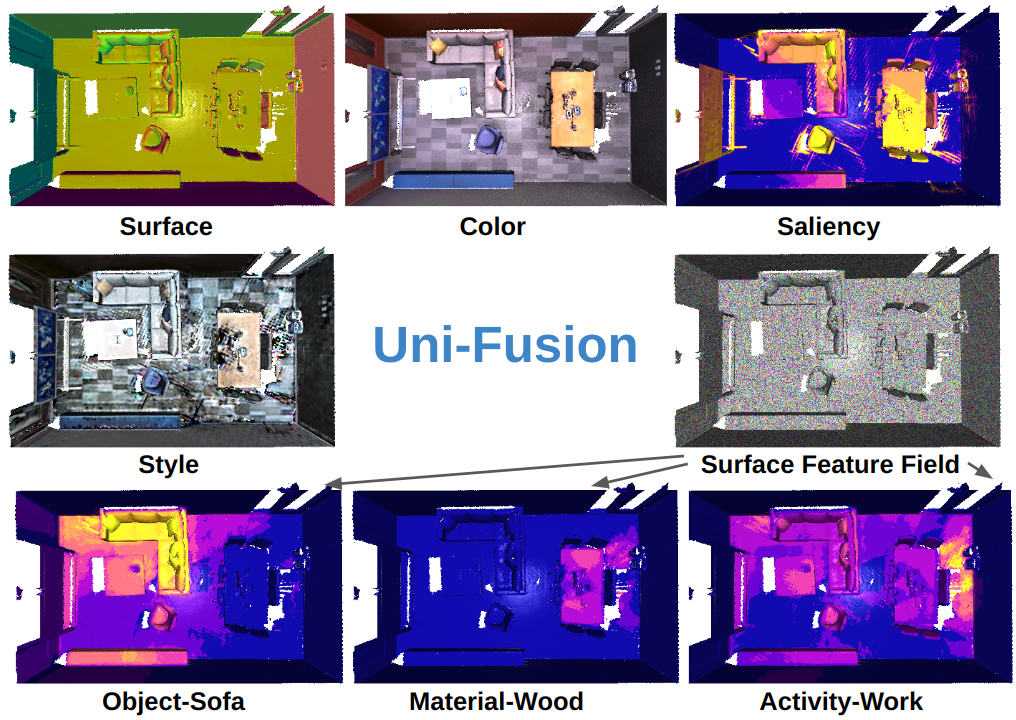}
	\caption{One Universal Continuous Mapping for all reconstructions. Such as surface, properties including RGB, saliency, style and $...$, even high dimensional features (CLIP embeddings, and etc). A rendered video is available on youtube\protect\footnotemark.}
	\label{fig:cover}
	\vspace{-.3cm}
\end{figure}
\footnotetext{\url{https://www.youtube.com/watch?v=4Z-u7yU2ARU}}
For scene-level reconstruction, most recent works rely on SDFs.
TSDF-Fusion~\cite{curless1996volumetric}, as the most widely used reconstruction model, has facilitated real-time 3D reconstruction.
With the rapid development of RGB-D sensors, such as the Kinect and RealSense series, standard models have been invented~\cite{izadi2011kinectfusion,dai2017bundlefusion}.
These models utilize discretized SDFs and the Marching Cubes algorithm~\cite{lorensen1987marching} to generate surface meshes. 

To address the high memory consumption associated with such a representation, recent techniques rely on deep neural networks to encode the geometry~\cite{huang2021di,yuan2022algorithm,li2022bnv,zhu2022nice}.
By using sparse voxels in the whole scene, high-dimensional vectors are extracted for each local geometry.
With this representation, researchers have proposed to fuse the map of latent vectors instead of the explicit field.
The explicit field is then extracted with arbitrary resolution.
Thus, such a form of representation produces a continuous mapping.
These methods are called Neural Implicit Maps (NIMs).

Previous studies have demonstrated it with encoding functions
pre-trained on object datasets~\cite{huang2021di,yuan2022algorithm,li2022bnv}.
Pre-training on color or other point properties becomes impractical due to the much higher complexity of the context pattern, compared to shape.
Currently, the only recent solutions for continuous color reconstruction are based on back-propagation to update local latent features~\cite{zhu2022nice} or Neural Radiance Fields (NeRF)~\cite{rosinol2022nerf}.
However, these methods require a significant number of training epochs and are not suitable for real-time applications.
Therefore, this paper aims to fill the gap by introducing a universal model that directly encodes arbitrary properties without the need for time-consuming learning or training.

Our model uses Gaussian Process Regression (GPR) as its basis.
Firstly, we propose to decouple the GPR by utilizing the approximation of the kernel function.
Which supports the encoding of the point cluster into a single feature vector.
Then, by leveraging the
sparse voxel structure, we construct a local feature vector within each voxel to form a map of latents.
Since our encoding-decoding function uses GPR, the entire model does not pre-touch any format of data properties.
Therefore, our Uni-Fusion is applicable to arbitrary reconstruction applications.

We believe that there dose not exist such a model that can handle every aspect of robot perception.
Therefore, based on this encoder, we introduce Uni-Fusion, a universal model for all types of data that generates continuous maps.
A selected set of examples of ``what Uni-Fusion can do'' is presented in~\cref{fig:cover}.
With various input data, our Uni-Fusion model efficiently encodes and generates continuous maps for surfaces, colors, styles, and more.
To further explore the potential of this model, we even construct a field for high-dimensional CLIP embedding~\cite{ghiasi2022scaling}.

The contributions of this work are as follows:
\begin{itemize}
	\item
	we propose a universal encoding-decoding model for local information that does not require any training,
	\item
	we present the first universal continuous mapping model, capable of constructing continuous surfaces and various surface property fields, including high-dimensional features such as CLIP embeddings.
	\item
	we implement applications to demonstrate the capabilities of our proposed model, including: (1) incremental surface \& color reconstruction, (2) 2D-to-3D transfer on a 3D canvas, (3) open-vocabulary scene understanding.
\end{itemize}
\noindent \textit{Content Overview:} 
In \cref{sec:related_works}, we provide an overview of related works. 
Then, in~\cref{sec:UE} and~\cref{sec:UCM}, we present the Universal Encoder and the Uni-Fusion model, a universal continuous mapping framework built upon that encoder.
Afterwards, in~\cref{sec:apps}, we showcase the broad applicability of Uni-Fusion by presenting a number of applications in different scenarios.
The capabilities of Uni-Fusion are evaluated in~\cref{sec:exp}.
%
Finally, we outline the future directions and conclude this paper.

\section{Related Works} 
\label{sec:related_works}

We first discuss the development of continuous mapping from 2D scenes to 3D scenes. 
Then we explore the importance of recent state-of-the-art (SOTA) reconstructions utilizing neural implicit models.
Next, we examine the development of kernel methods that are closely related to our approach.

\subsection{Continuous Mapping}

The practice of continuous mapping originated in 2D scenarios.
Gaussian Process Occupancy Maps (GPOM)~\cite{o2012gaussian} use Gaussian Process Regression (GPR) to predict a continuous representation, allowing the construction of maps at arbitrary resolutions. 
To improve the scalability, Kim et al.~\cite{kim2013continuous} employ a divide and conquer strategy where GP is applied in each cluster.
Then, the incremental approach to GPOM is employed by leveraging the Bayesian Committee Machine (BCM) technique~\cite{ghaffari2018gaussian,yuan2018fast}.

Meanwhile, the Hilbert Maps approach~\cite{senanayake2017bayesian,zhi2019continuous} has been introduced. This approach is a mapping technique that does not require an explicit formulation.
Hilbert Maps achieve continuous mapping by continuously optimizing parameters. 

Due to the growing popularity of 3D sensors, there has been a shift in focus towards 3D scenarios.
Gaussian Process Implicit Surface (GPIS), has emerged in the field~\cite{martens2016geometric,lee2019online,wu2021faithful,ivan2022online}.
These methods,
given zero-level surface points, either sample points along the normal direction and assign distance values, or directly use derivative models with normals as labels (we will discuss this in more detail in our mapping~\cref{sec:surface_mapping}).
However, these methods focus primarily on shapes rather than entire scenes.
This is due to the fact that GPR-based methods naturally incur high computational costs when 
dealing with large amounts of data.
This is especially true when moving from 2D to 3D testing scenarios.
Therefore, we leverage the concept of Neural Implicit Maps~(\cref{sec:related:NIM}) to address this challenge and focus on scene reconstruction.

\subsection{Neural implicit based Reconstruction}
\label{sec:related:NIM}

Neural implicit reconstruction was originally introduced for SDF and occupancy-based reconstruction.
The seminal work by Park et al.~\cite{park2019deepsdf}, known as DeepSDF, employs a deep model to encode geometry prior using multi-layer perceptrons (MLPs) and extracts discretized SDF through decoding queries.
It then uses an algorithm similar to Marching Cubes~\cite{lorensen1987marching} for mesh extraction. 
For the latter, Occupancy Networks~\cite{mescheder2019occupancy} learn to estimate the occupancy probability at positions using an implicit function.
The Multiresolution IsoSurface Extraction (MISE) technique is used to generate meshes.
\begin{figure}[b]
	\centering
	\includegraphics[width=.8\linewidth]{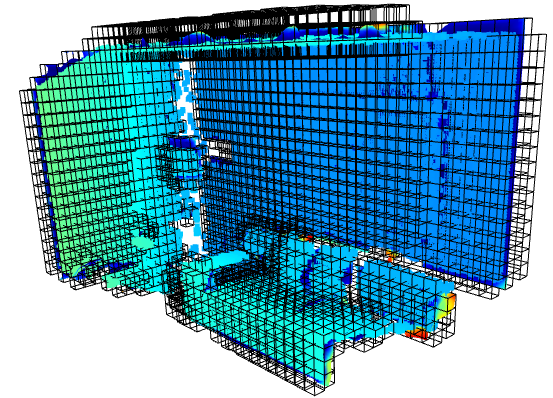}
	\caption{Uniform, sparse voxels. Each voxel is encoded into one feature vector $\V F_m$~\cite{huang2021di}.}
	\label{fig:PLV}
\end{figure}

To enhance the efficiency of the reconstruction, 
DeepLS~\cite{chabra2020deep} uses multiple local deep priors and reconstructs based on a set of local SDFs. 
Jiang et al.~\cite{jiang2020local} additionally proposes the use of a local implicit grid to further simplify the model complexity of the encoding.
Conversely, Convolutional Occupancy Networks~\cite{peng2020convolutional} explore the local latent by replacing the encoder-decoder with optimization on a grid of local features, thereby alleviating the burden of MLPs. 

More recently, Neural Implicit Maps (NIM) have achieved real-time 3D reconstruction of real scenes.
Huang et al. propose DI-Fusion~\cite{huang2021di}, a neural implicit mapping method that performs incremental scene reconstruction.
DI-Fusion achieves high memory efficiency and produces improved reconstructions, by fusing on maps of latent features.
Similarly, NeuralBlox~\cite{lionar2021neuralblox} fuses a grid of latent features with known external pose estimation.
BNV-Fusion~\cite{li2022bnv} further improves feature quality by incorporating post-fusion optimization.
To address large scene reconstructions that may involve loops, NIM-REM~\cite{yuan2022algorithm} introduces an SE(3)-transformation algorithm for NIM and develops a mapping\&remapping model that works in conjunction with bundle adjustment and loop closing.
Thinking outside the box, Sucar et al.~\cite{sucar2021imap} propose iMAP, a novel SLAM pipeline that incorporates neural implicit-based incremental reconstruction.
Using a differentiable rendering model, iMAP performs online optimization of the reconstruction by minimizing the image distances.
Zhu et al. present NICE-SLAM~\cite{zhu2022nice}, which replaces MLPs with a convolutional occupancy grid to further improve the efficiency and quality of the reconstruction.

In this work, our model also uses regularly spaced voxels, as shown in~\cref{fig:PLV}, for local encoding-decoding. 
Since we propose to use a single model for all properties, pre-training such a model is not feasible.
Therefore, we employ a kernel method to achieve this objective.

\subsection{Kernel Function Approximation}

Kernel methods suffer from a scalability issue due to the $\mathcal{O}(n^3)$ time complexity required during regression.
One possible solution is kernel matrix approximation, but this is beyond the scope of this paper.
Another solution, kernel function approximation, aims to enhance the scalability of kernel methods by employing explicit vector maps. For example, kernel function $k(\V x_1, \V x_2) \approx   v(\V x_1)^T  v(\V x_2)$ with mapping function 
$v:\mathbb{X}\rightarrow \mathbb{R}^l $~\cite{deng2022neuralef}.
Two approaches are commonly used for approximation: Random Fourier Features (RFFs)~\cite{rahimi2007random,rahimi2008weighted,yu2016orthogonal,munkhoeva2018quadrature,francis2021major} and Nystr\"om methods~\cite{francis2021major,williams2000using,deng2022neuralef}. 

RFFs explicitly handle the shift-invariant kernels by mapping the data using the Fourier transform technique~\cite{francis2021major}.
However, RFFs are primarily employed for shift-invariant kernels and require a large $l$ for their operation.
Furthermore, since RFFs are data-independent, they exhibit significantly worse generalization performance compared to Nystr\"om methods~\cite{yang2012nystrom}.
In contrast, Nystr\"om methods can approximate any positive definite kernel~\cite{deng2022neuralef}.
It relies on finding the eigenfunctions to form the approximation:
\begin{align*}
	k(\V x_1, \V x_2) = \sum_{i\ge 1} \mu_i \varphi_i(\V x_1)\varphi_i(\V x_2),
\end{align*}
where $\varphi_i$ and $\mu_i\ge 0$ are eigenfunctions and eigenvalues of kernel function $k$ with respect to the probability measure $q$.
With the top-$l$ eigenpairs, the Nystr\"om method approximates the kernel function with $ k(\V x_1, \V x_2) = \sum^l_{i\ge 1} \mu_i \varphi_i(\V x_1)\varphi_i(\V x_2)$.

However, Nystr\"om methods are computationally expensive for medium-sized training data and require evaluating each sample $N$ times, which is inefficient for direct regression of a continuous map.

Instead, we encode the local geometry and properties, eliminating the need for a full-space approximation.
By restricting the approximation to the limited space $\V x\in [-0.5,0.5]^3$, and applying the function in each local region, we reduce the computational burden.
\section{Decoupled Regression as an Universal Encoding}
\label{sec:UE}

In this section, we propose a universal encoding model for point clouds.
Building on this encoder, in~\cref{sec:UCM}, we introduce Universal Continuous Mapping.

\subsection{Decoupled Gaussian Process Regression as Encoder-Decoder}
\label{sec:encoder}

Inspired by DI-Fusion~\cite{huang2021di}, which introduces latent feature maps for continuous surface prediction, our goal is to generate a latent vector for each local patch.
%
First, a universal encoder design comes from the Gaussian Process Regression (GPR), which is a widely used technique for low-dimensional regression.
It has been used in various mapping models~\cite{yuan2018fast,martens2016geometric}.
Given a set of $N$ observation points $\{(\V x_n, \V y_n)\}_{n=1}^{N}$ where point positions are $\V x_n \in \mathbb R^d$ ($d=3$ in this paper) and point properties are $\V y_n \in \mathbb R^c$, GPR is used to regress a function $f$ that best explains the data.
The $N$ points are aggregated in the $N\times d$ matrix $\V X$, and the targets are collected in the $N\times c$ matrix $\V Y$.

The Gaussian process assumes that the data is sampled from a multivariate Gaussian distribution, i.e.,
\begin{equation}
	\begin{bmatrix}
		\V Y\\
		\V Y_{*}
	\end{bmatrix}
	\sim \mathcal{N}(\V 0, 
	\begin{bmatrix}
		k(\V X, \V X)\ k(\V X, \V X_{*})\\
		k(\V X_{*}, \V X)\ k(\V X_{*}, \V X_{*})
	\end{bmatrix}	 
	).
\end{equation}
where $(\V X, \V Y)$ and $(\V X_*, \V Y_*)$ represent the observation and inference pairs. 
For simplicity, we denote the matrix $\V K=k(\V X, \V X)$, $\V K_{*}=k(\V X, \V X_*)$, $\V K_{**}=k(\V X_*, \V X_*)$.

With the derivative in~\cite{williams2006gaussian}, we obtain 
\begin{equation}
	\V Y_{*} | \V X, \V Y, \V X_{*} \sim \mathcal{N} (\V K_{*}^T\V K^{-1}\V Y, \V K_{**}-\V K_{*}^T\V K^{-1}\V K_{*}).
\end{equation}
When an additional Gaussian error is introduced as $\V y=f(\V x)+\epsilon$, the covariance of $\V X$ is rewritten as $\V K+\delta^2_n\V I$.
The regression result is typically considered to be the mean
\begin{equation}
	\label{eq:mean}
	\V Y_{*} = \V K_{*}^T(\V K+\delta^2_n\V I)^{-1}\V Y.
\end{equation}
This well illustrates the challenge of using Gaussian process regression directly in large-scale reconstructions due to its $\mathcal{O}(n^3)$ time complexity, which is not feasible.
Furthermore, the formula (\cref{eq:mean}) is impractical since it requires the large input point cloud data to be maintained for the $ \V K_{*}$ computation. 

To address the issue of high complexity and avoid the need to maintain the entire point cloud, we propose to decouple the GPR to obtain the low-dimensional latent vectors for local regions.
The decoupling process involves approximating the kernel function as
\begin{equation}
	k(\V X,\V X_{*} ) \approx f_{posi}(\V X) f_\text{posi}(\V X_{*})^T
	\label{eq:k_approx}
\end{equation}
where $f_\text{posi}: \mathbb{R}^3\rightarrow \mathbb{R}^l$.
We refer to the function $f_\text{posi}$ as the position encoding function, consistent with the Neural Implicit Maps model, Di-Fusion~\cite{huang2021di}.
Thus, \cref{eq:mean} is rewritten as
\begin{equation}
	\label{eq:approx_K_mean}
	\begin{split}
		\V Y_{*} 
		=f_\text{posi}(\V X_*)f_\text{enc}(\V X, \V Y)
	\end{split}
\end{equation}
where the content encoder function is
\begin{equation}
	\label{eq:encode}
	f_\text{enc}(\V X, \V Y) = f_\text{posi}(\V X)^T(f_\text{posi}(\V X)f_\text{posi}(\V X)^T +\delta^2_n\V I)^{-1}\V Y \in \mathbb{R}^{l \times c}.
\end{equation}

The encoded feature is denoted by $ \V F_{(\V X, \V Y)} = f_{enc}(\V X, \V Y) \in \mathbb{R}^{l\times c}$.
which serves as the basis for the construction of latent maps in the subsequent universal continuous mapping model (\cref{sec:UCM}).
Specifically, for geometry encoding we set $l=20$ and $c=1$, resulting in a $20$-dimensional vector feature.
Similarly, for color encoding, we have $l=20$ and $c=3$.

The decoding value for the inferred point $\V x_*$ is expressed as
\begin{equation}
	\label{eq:decode}
	f_{dec}(\V x_{*},\V F_{(\V X, \V Y)}) = f_{posi}(\V x_{*}) \V F_{(\V X, \V Y)}.
\end{equation}
Thus, a signed distance field or surface property field is approached.

In the following we derive the approximation function.

\begin{figure}[]
	\centering
	\psfragfig[width=1\linewidth]{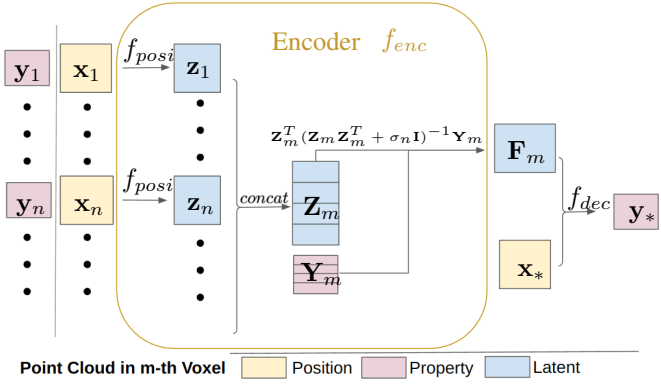}{
		\psfrag{y1}{$\V y_1$}
		\psfrag{x1}{$\V x_1$}
		\psfrag{yn}{$\V y_n$}
		\psfrag{xn}{$\V x_n$}
		\psfrag{z1}{$\V z_1$}
		\psfrag{zn}{$\V z_n$}
		\psfrag{fp}{$f_{posi}$}
		\psfrag{fd}{$f_{dec}$}
		\psfrag{fe}{\color{mybronze}{$f_{enc}$}}
		\psfrag{ec}{\color{mybronze}{{Encoder}}}
		\psfrag{zm}{$\V Z_m$}
		\psfrag{ym}{$\V Y_m$}
		\psfrag{Fm}{$\V F_m$}
		\psfrag{x}{$\V x_{*}$}
		\psfrag{y}{$\V y_{*}$}
		\psfrag{f}{\tiny $\V Z_m^T(\V Z_m\V Z_m^T+\sigma_n\V I)^{-1}\V Y_m$}
	}

	\caption{Interpreting formula with a graph that is coherent to the Encoder-decoder structure in Neural Implicit Maps~\cite{huang2021di}. }
	\label{fig:encoder}
	\vspace{-.3cm}
\end{figure}

\subsection{Position Encoding with Approximated Kernel Function} 
\label{sec:posi_encode}

Considering that our mapping needs to encode the local geometry \& property, the encoding function only requires to touch points in a limited region ($[-.5,.5]^3$ in our case).

As we discussed in related work, Nytr\"om methods offer greater accuracy than RFFs, because they depend on the given points. This property is well-suited to our application.

The Nystr\"om method for kernel approximation begins with the use of eigenfunctions according to Mercer's theorem:
\begin{equation}
	k(\V x_1, \V x_2) = \sum_{i\ge 1} \mu_i \psi_i(\V x_1)^T\psi_i(\V x_2)
	\label{eq:eigen}
\end{equation}
where $\psi_i$ and $\mu_i\ge 0$ are eigenfunctions and eigenvalues of kernel function $k$ with respect to the probability measure $q$.

Given a set of anchor samples $\hat{\V X}=\{\hat{\V x}_1,\cdots,\hat{\V x}_N\}$, 
we perform eigen-decomposition on the matrix $k(\hat{\V X}, \hat{\V X})$ to obtain its eigenpairs $\{(\lambda_i,\V u_i)\}_{i\in\{1,\cdots,l\}}$ with rank $l$.

Subsequently, Nystr\"om method produces
\begin{equation}
	\label{eq:nytrom_1}
	\psi_i(\V x) = \sum_{n}k(\V x, \hat{\V x}_n)\V u_{i,n}, i= 1,\cdots,l.
\end{equation}




To simplify, we express the eigenfunction as 
\begin{equation}
	\psi_i(\V x) = k(\V x, \hat{\V X})\V u_i .
	\label{eq:nytrom_2}
\end{equation}
Similarly, the eigenvalue is written as $\mu_i=\frac{1}{\lambda_i}$.

For clarity, we introduce the notation $ \pmb \mu=[{\mu_1},\cdots,{\mu_l}]$ and
${\Psi}=[\psi_1,\cdots,{\psi_l}]^T$
based on \cref{eq:eigen} where
${\Psi}:\mathbb{R}^3\rightarrow \mathbb{R}^l$.

To maintain consistency with \cref{eq:k_approx}, we set 
\begin{equation}
	f_{posi}(\V x) = diag(\sqrt{\pmb\mu})\Psi(\V x) 
	\label{eq:f_posi}
\end{equation}
where $diag$ produces diagonal matrix.
$f_{posi}$ above refers to the position encoder in \cref{eq:encode}.

In this paper, we employ the Mat\'ern kernel function~\cite{genton2001classes}\footnote{\url{https://en.wikipedia.org/wiki/Mat\'ern\_covariance\_function}}:
\begin{equation}
	\label{eq:matern}
	k(\V x_1, \V x_2) = \sigma^2 \frac{2^{1-\nu}}{\Gamma(\nu)}(\sqrt{2\nu}\frac{dist(\V x_1, \V x_2)}{\rho})K_{\nu}(\sqrt{2\nu}\frac{dist(\V x_1, \V x_2)}{\rho})
\end{equation}
where $\Gamma$ presents the gamma function, $K_{\nu}$ is the modified Bessel function of the second kind, $dist$ denotes the Euclidean distance, and $\sigma$ and $\rho$ are hyperparameters of the kernel function.
We utilize the half integer $\nu=3+\frac{1}{2}$, which results in the specific function:
\begin{equation}
	\label{eq:matern_specific}
	k(d) = \sigma^2(1+\frac{\sqrt{7}d}{\rho}+\frac{2}{5}(\frac{\sqrt{7}d}{\rho})^2+\frac{1}{15}(\frac{\sqrt{7}d}{\rho})^3)exp(-\frac{\sqrt{7}d}{\rho})
\end{equation}
where $d=dist(\V x_1, \V x_2)$ for short.

To approximate the above kernel function, anchor points are required.
We sample $N_a=256$ points uniformly 
from $[-.5,.5]^3$ cube (as $\hat{\V X}$) to compute the kernel matrix $\V K_{a}=k(\hat{\V X},\hat{\V X})$.
Subsequently, we perform an eigendecomposition on this kernel matrix, resulting in $\V K_{a}=\V U \Lambda \V U^T$. 
Lastly, $\V U$ and $\Lambda$ are utilized in \cref{eq:f_posi} and~\cref{eq:k_approx} to form the kernel function approximation.

It is important to note that the dimension of the encoded feature, $l$, used in \cref{sec:encoder}, depends on $\V U$. 
It further determines the size of the map in the next section (~\cref{sec:UCM}).
\begin{figure}[htbp]
	\centering
	\psfragfig[width=1\linewidth]{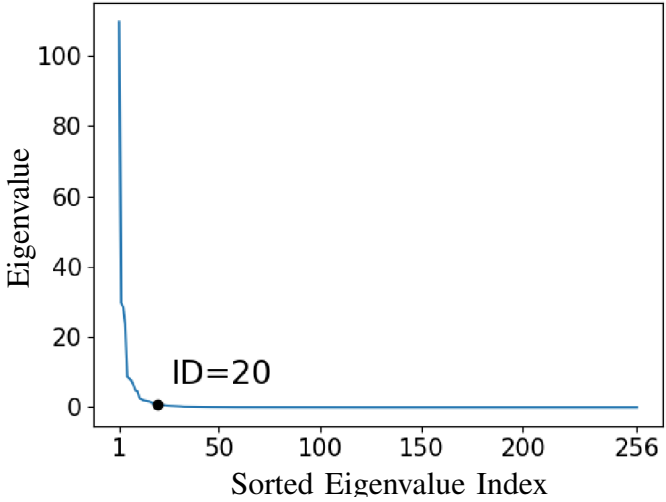}{
		\psfrag{a}{\large{Sorted Eigenvalue Index}}
		\psfrag{b}{\large{\rotatebox{90}{Eigenvalue}}}
	}
	\caption{Sorted eigenvalues for $\V K_{a}$'s eigendecomposition.}
	\label{fig:eigenvalues}
\end{figure}

The eigenvalues of $\Lambda$ are plotted in \cref{fig:eigenvalues}, revealing that the matrix is primarily influenced by a small number of pairs with significant eigenvalues.
Most of the eigenvalues are less than 1. 
Therefore, we choose $l=20$ which is about $0.8$ in this plot to approximate the kernel while maintaining a compact feature dimension.

Note that we perform a single sampling and decomposition step. Subsequently, the encoding-decoding process in \cref{fig:encoder} only requires loading and reusing the parameters for $f_{posi}$, $f_{enc}$ and $f_{dec}$. In contrast, related works often involve pre-training on large datasets of objects~\cite{huang2021di,li2022bnv} or indoor scenes~\cite{zhu2022nice}. 
For our model, however, \textbf{no training is needed}.


\section{Universal Continuous Mapping} 
\label{sec:UCM}

The previous~\cref{sec:UE} suggests a universal encoding model for different types of data.
Based on this function, in this section, our Universal Continuous Mapping produces a map of latents to implicitly represent the scene.
We refer to this scene representation as \textbf{Latent Implicit Maps (LIM)},
which supports surfaces, surface properties, and high-dimensional surface features. 
%
\begin{figure}[htbp]
	\centering
	\psfragfig[width=1\linewidth]{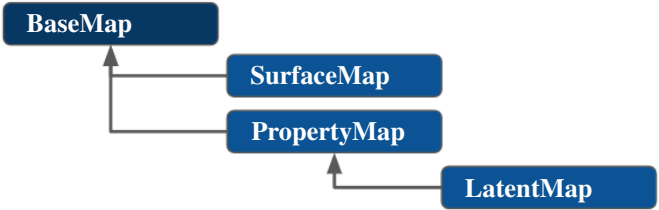}{
		\psfrag{A}{\color{white}{\textbf{BaseMap}}}
		\psfrag{B}{\color{white}{\textbf{SurfaceMap}}}
		\psfrag{C}{\color{white}{\textbf{PropertyMap}}}
		\psfrag{D}{\color{white}{\textbf{LatentMap}}}
	}
	\caption{Inheritance graph for the class of Latent Implicit Maps (LIM).}
	\label{fig:inherit}
\end{figure}


The inheritance graph is shown in \cref{fig:inherit}.
We introduce a BaseMap to handle the voxel structure (\cref{sec:map_rep}), dynamically allocate space and fuse maps (\cref{sec:fuse}).
The SurfaceMap (\cref{sec:surface_mapping}) and the PropertyMap (\cref{sec:context_fields}) are derived from the BaseMap and are designed to handle specific data and applications.
The LatentMap (\cref{sec:latent_fields}) is derived from PropertyMap.
The main difference is that the PropertyMap works primarily with low-dimensional properties, such as color ($c=3$), infrared ($c=1$) data, and so on. 
On the other hand, the LatentMap handles much higher dimensional features, such as CLIP embeddings~\cite{ghiasi2022scaling} ($c=768$), depending on the specific application requirements.

\subsection{Map Representation}
\label{sec:map_rep}

We follow Neural Implicit Maps (\cite{huang2021di,yuan2022algorithm,li2022bnv}) to use uniformly spaced voxels to sparsely represent the scene. The scene is denoted as $\V V=\{\V v_m = (\V c_m, \V F_m, w_m)\}$, where $m$ is the voxel index. Each voxel $\V v_m$
comprises the
voxel center $\V c_m\in \mathbb R^3$, voxel latent feature $\V F_m\in \mathbb R^{l \times c}$, and the count of observed points $w_m\in\mathbb N$.

Given a sequence of incremental frames as input, our model constructs local LIMs (\cref{sec:surface_mapping}, \cref{sec:context_fields}, \cref{sec:latent_fields}) and fuses (\cref{sec:fuse}) them into a global LIM.
Then, we derive the explicit map from the global LIM.

\subsection{Surface Mapping}
\label{sec:surface_mapping}

Because the input point cloud $\V X$ is located on zero-level surface, it is not adequate to recover a 3D field of scene, $f_{SDF}:\mathbb R^3 \rightarrow \mathbb R$.
Therefore, we use the concept of Gaussian Process Implicit Surfaces (GPIS)~\cite{martens2016geometric,lee2019online,wu2021faithful,ivan2022online} to incorporate derivatives into kernel or to sample additional non-zero level points.
Both derivative-based and sample-based GPISs approaches utilize normal information.
Hence, we first preprocess $\V X$ to obtain normals $\V S$. 

\subsubsection{Using Derivatives based GPIS}
\label{sec:GPIS:deri}

From~\cite{martens2016geometric}, the derivatives of a GP are also Gaussian. 
Therefore, the covariance between data and derivatives is computed by differentiating the covariance function~\cite{solak2002derivative}. 
Specifically:
\begin{equation}
	\begin{split}
		cov(\frac{\partial f_{SDF}(\V x) }{\partial x_i}, f_{SDF}(\V x^{'})) &= \frac{\partial k(\V x, \V x^{'})}{\partial x_i}\\
		&=\frac{\partial}{\partial x_i}[f_{posi}(\V x)]f_{posi}(\V x^{'})^T.
	\end{split}
\end{equation}
Additionally,
\begin{equation}
	\begin{split}
		cov(\frac{\partial f_{SDF}(\V x) }{\partial x_i}, \frac{\partial f_{SDF}(\V x^{'}) }{\partial x_j}) &= \frac{\partial^2 k(\V x, \V x^{'})}{\partial x_i\partial x^{'}_j}\\
		&=\frac{\partial}{\partial x_i}[f_{posi}(\V x)]\frac{\partial}{\partial x^{'}_j}[f_{posi}(\V x^{'})]^T.
	\end{split}
\end{equation}
Given points $\{\V x_n\}^N_{n=1}$ with normals $\{\V s_n\}^N_{n=1}$ and field values $\{\V y_n=0\}^N_{n=1}$,
the position encoding function for derivatives is $f_{posi, deri}(\V x, i) = \frac{\partial}{\partial x_i}[f_{posi}(\V x)]$.
Its corresponding field value is the normal value $s_i$ on the axis $i$.
Therefore, we define
\begin{multline}
	f_{posi, gpis}(\V X)=\\ [f_{posi}(\V X), f_{posi, deri}(\V X, 1), f_{posi, deri}(\V X, 2), f_{posi, deri}(\V X, 3)]
\end{multline}
with regression values $\V Y_{gpis}=[\V 0, \V s_{\cdot,1}, \V s_{\cdot,2}, \V s_{\cdot,3}]^T$, where $\V 0 = zeros(1,N)$, $\V s_{\cdot,i}=[s_{1,i},\cdots,s_{N,i}]$.

Then the local geometric encoding function is defined as
\begin{multline}
	f_{enc, gpis}(\V X, \V Y, \V S)=\\
	f_{posi, gpis}(\V X)^T(f_{posi, gpis}(\V X)f_{posi, gpis}(\V X)^T + \sigma_n^2\V I)^{-1}\V Y_{gpis}.
\end{multline}
By introducing derivatives into the kernels, the matrix size is increased by a factor of 15, 
while the encoded feature dimension remains low at $l$:
$\V F_{(\V X, \V Y, \V S)}=f_{enc, gpis}(\V X, \V Y, \V S)\in \mathbb R^{l\times 1}$.

For inference with the points $\V x_*$, the predictions are consistent with \cref{eq:decode}, $\V y_*=f_{posi}(\V x_*) \V F_{(\V X, \V Y, \V S)}$.

\subsubsection{Using Sample based GPIS}
\label{sec:GPIS:sample}

Sample-based GPIS is commonly used in GPIS research.
This method avoids the computation of Jacobians and allows for smaller kernel sizes, resulting in significant reductions in computational costs in terms of time and memory.

In sample-based GPIS, given points $\{\V x_n\}^N_{n=1}$ with normals $\{\V s_n\}^N_{n=1}$ and field value $\{\V y_n=0\}^N_{n=1}$, the dataset is extended by sampling points along the normal direction.
The corresponding field values are the signed distances as the sampled points move along the normal.
Then, \cref{eq:encode} is applied to the extended points and distances ($\V X_{ext}$, $\V Y_{ext}$).
The inference process of this model is the same as derivative-based GPIS in~\cref{sec:GPIS:deri}.

\bigskip

For each frame, points are assigned to their corresponding voxel.
Then, we encode the local geometry within each voxel using methods described in~\cref{sec:surface_mapping}, either derivative-based GPIS or sample-based GPIS to obtain the voxel representation $\V v=(\V c, \V F, w)$ where $\V F\in\mathbb R ^ {l\times 1}$ represents the geometric latent vector.
Subsequently, the local LIMs are fused into a global LIM according to the fusion procedure outlined in~\cref{sec:fuse}.

To visualize the surface result, we
construct the signed distance field from the global LIM by performing inference on sample points.
The sample points are generated on a grid within each voxel, with a certain resolution.
By applying the Marching Cubes algorithm to the SDF, we obtain a surface mesh that represents the reconstructed surface.

\subsection{Surface Property Fields}
\label{sec:context_fields}

The previous surface mapping approach discussed in~\cref{sec:surface_mapping} can be viewed as a special case of surface property mapping.
However, it is important to note that these two mappings operate in different spaces. 
Specifically, in our implementation, we do not derive the SurfaceMap class from PropertyMap. 
Instead, we introduce a BaseMap that performs common operations and allows them to be specific to local map construction and visualization, e.g., meshing and coloring.

We introduce the more general mapping of surface properties.
Since all points lie on the zero level of the signed distance fields, the PropertyMap naturally operates within a subspace of $\mathbb R^3$, the surface $\mathcal{S}$.
A surface property in this context refers to any property associated with each point, such as color, infrared values, and so on.
These properties are represented by the $\V y$ value in the encoder diagram shown in~\cref{fig:encoder}, with a dimensionality of $c$.

As an example, we take the most commonly used surface property, color.
Given an observed colored point cloud $\{(\V x_n, \V q_n)\}^N_{n=1}$ as input, where $\V q_n$ denote the RGB color values.
The corresponding surface property values are $\{\V y_n=\V q_n\}^N_{n=1}$.
We aggregate these values into two $N\times 3$ matrices $\V X$ and $\V Q$.
Therefore, the encoded feature for this point cloud is obtained using the following equation: 
\begin{equation}
	\V F_{color}=f_{posi}(\V X)^T(f_{posi}(\V X)f_{posi}(\V X)^T +\delta^2_n\V I)^{-1}\V Q.
\end{equation}
Here, $\V F_{color}\in \mathbb R^{l\times 3}$ represents the color feature.

Since we use $l=20$ for the approximation function in our experiments, the color map only needs to store $20\times3$ float values in each voxel to represent a continuous color field.
It is important to note that our model requires no training and can be applied directly to different types of data.
During inference, since the field is in the surface space $\mathcal{S}$, we sample points $\V x_*$ at arbitrary resolutions, either from a known mesh or a surface constructed from previous surface mapping (\cref{sec:surface_mapping}).
Following~\cref{eq:decode}, the inference point $\V x_*$ is position encoded and multiplied by the color feature $\V F_{color,m}$ in the corresponding voxel $m$ to obtain its value $\V q_{*}=f_{dec}(\V x_{*}, \V F_{color,m})$.

\subsection{Surface Feature Fields}
\label{sec:latent_fields}

Surface feature fields represent an extension of the previous surface property fields discussed in~\cref{sec:context_fields}.
In this extension, we broaden the scope of surface properties to include features. 
This demonstrates the versatility of our mapping model, as it is applied directly without the need of any training.

We start by considering an embedding function $f_{im}:\mathbb R^{N\times 3}\rightarrow \mathbb R^{N\times c}$ that processes the input data $\V X$. We treat the feature of each point as a surface property, where
$\{\V y_n=f_{im} (\V X)_{\V x_n}\in \mathbb R^{l\times c}\}^N_{n=1}$.

Following the encoding and fusion steps described in~\cref{sec:context_fields} and~\cref{sec:fuse}, we construct latent implicit maps for the surface feature fields.
As a result, we extract maps of features at arbitrary resolutions using the function $f_{dec}(\cdot, \V F_\cdot)$.

\begin{figure}[t]
	\centering
	\psfragfig[width=1\linewidth]{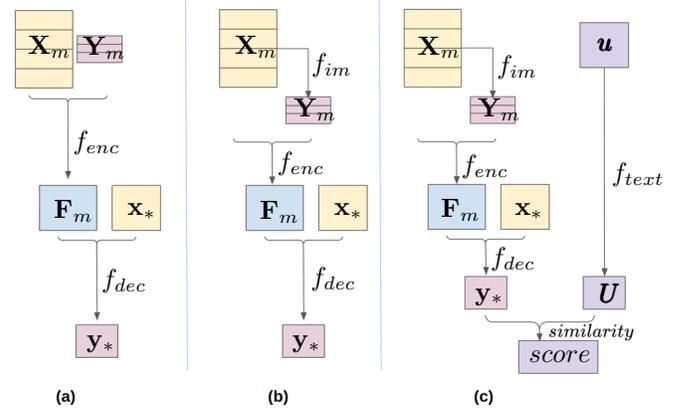}{
		\psfrag{t}{$f_{text}$}
		\psfrag{xm}{$\V X_m$}
		\psfrag{ym}{$\V Y_m$}
		\psfrag{fe}{$f_{enc}$}
		\psfrag{fm}{$f_{im}$}
		\psfrag{Fm}{$\V F_m$}
		\psfrag{x}{$\V x_{*}$}
		\psfrag{y}{$\V y_{*}$}
		\psfrag{fd}{$f_{dec}$}
		\psfrag{u}{$\pmb u$}
		\psfrag{U}{$\pmb U$}
		\psfrag{s}{$score$}
	}
	\caption{Encoding-decoding diagram in various applications. (a) fold applications obtain point properties ($\V Y_\text{m}$) directly from the sensor. (b) fold applications derive point properties using a function $f_{im}$ that captures style, saliency and etc. (c) fold applications utilize feature as $\V Y_\text{m}$ to construct a LIM for a (CLIP) feature field. Then, a text command is used to extract the semantic information.}
	\label{fig:latent_diff}
\end{figure}

We illustrate an application in~\cref{sec:openvoc}, specifically in the context of open-vocabulary scene understanding. 
In this application, our model constructs a CLIP space feature field on the surface,
enabling it to respond to textual input.
The key difference compared to surface property fields is demonstrated in~\cref{fig:latent_diff}.
In~\cref{fig:latent_diff}(c), a CLIP text encoder $f_{text}$ is added to encode the text command $\pmb u$ into the CLIP feature $\pmb U$.
By leveraging the surface field for CLIP embeddings generated by the left branch,
our model identifies the desired region by computing the similarity between features.

\subsection{Map Fusion}
\label{sec:fuse}

We adopt the voxel-to-voxel fusion approach from Neural Implicit Maps~\cite{huang2021di} to update the LIM. The fusion operation is performed as follows:
\begin{equation}
	\label{eq:fuse}
	\V F_m \leftarrow \frac{\V F_m w_m+\V F^{'}_m w^{'}_m}{w_m+w^{'}_m}, w_m\leftarrow w_m+w^{'}_m,
\end{equation}
where $\V v_m=(\V c_m, \V F_m, w_m)$ represents the voxel $m$ from the global LIM, and $\V v^{'}_m=(\V c^{'}_m, \V F^{'}_m, w^{'}_m)$ represents the voxel $m$ from  the local LIM.

\section{Applications}
\label{sec:apps}
To demonstrate the wide range of applications of our model, we have implemented the following series of applications:
\begin{enumerate}
	\item Incremental surface \& color reconstruction
	\item 3D saliency detection
	\item Open-vocabulary scene understanding
	\item Surface infrared field
	\item 3D style transfer
\end{enumerate}
Starting from our motivation in inspection and service robotics, we implement 1) incremental surface \& color reconstruction to visualize the robot environment.
For robot exploration, we implement 2) 3D saliency detection to indicate the salient regions in maps.
To recover object-level semantic information in the environment, we implement 3) open-vocabulary scene understanding to yield the regions containing the objects.
Furthermore, to demonstrate the flexibility, we implement 4) surface infrared fields and 5) 3D style transfer for artistic purposes. 

In~\cref{fig:latent_diff}, we classify these applications into 3 categories: (a) obtains properties directly from sensor observations, such as applications 1) and 4). (b) processes sensor data and predicts properties, such as applications 2), 5). (c) extends (b) by operating on high dimensional features, specifically application 3).

Applications 1) and 4) belong to the first category. We primarily describe 1) incremental surface \& color reconstruction (\cref{sec:incremental_reconstruction}), while for 4) we can easily replace color with infrared.
For the second with 2) and 5) in~\cref{sec:fabircated_prop}, we mainly discuss the usage of fabricated properties and do not provide detailed explanations of the mapping part, as it is already covered in the previous category.
Finally, we delve into the third category with application 3), which involves mapping a LIM for high-dimensional latent fields.
We demonstrate the flexibility and capability of this application
in~\cref{sec:openvoc}.

\subsection{Application: Incremental Reconstruction}
\label{sec:incremental_reconstruction}

\begin{figure*}[t]
	\centering
	\psfragfig[width=1\textwidth]{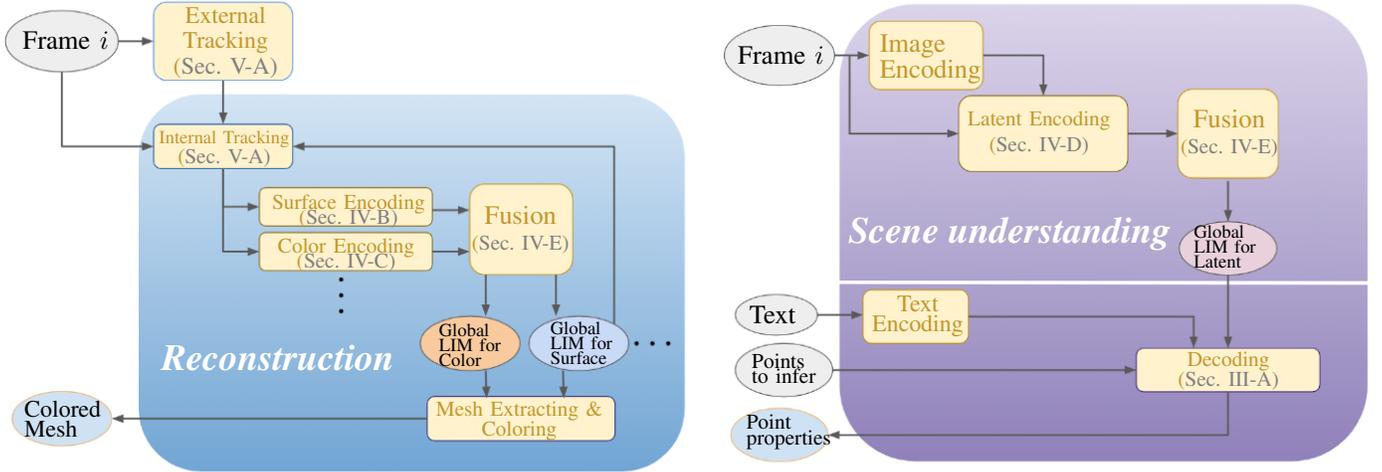}{
		\psfrag{A}{Frame $i$}
		\psfrag{B}{\begin{tabular}{@{}c@{}}
				\small\color{mybronze}{External}\\[-.5ex]
				\small\color{mybronze}{Tracking}\\[-.5ex]
				\small\color{mybronze}{(\color{gray}{Sec. V-A})}
			\end{tabular}}
		\psfrag{C}{\begin{tabular}{@{}c@{}}
				\scriptsize	\color{mybronze}{Internal Tracking}\\[-1ex]
				\footnotesize\color{mybronze}{(\color{gray}{Sec. V-A})}
		\end{tabular}}
		\psfrag{D}{\begin{tabular}{@{}c@{}}
				\footnotesize\color{mybronze}{Surface Encoding}\\[-1.4ex]
				\footnotesize\color{mybronze}{(\color{gray}{Sec. IV-B})}
		\end{tabular}}
		\psfrag{E}{\begin{tabular}{@{}c@{}}
				\footnotesize\color{mybronze}{Color Encoding}\\[-1.4ex]
				\footnotesize\color{mybronze}{(\color{gray}{Sec. IV-C})}
		\end{tabular}}
		\psfrag{F}{\begin{tabular}{@{}c@{}}
				\color{mybronze}{Fusion}\\[-.5ex]
				\footnotesize\color{mybronze}{(\color{gray}{Sec. IV-E})}
		\end{tabular}}
		\psfrag{G}{\begin{tabular}{@{}l@{}}
				\scriptsize	Global\\[-1.4ex]
				\scriptsize	LIM for\\[-1.4ex]
				\scriptsize	Color
			\end{tabular}}
		\psfrag{H}{\begin{tabular}{@{}l@{}}
				\scriptsize	Global\\[-1.4ex]
				\scriptsize	LIM for\\[-1.4ex]
				\scriptsize	Surface
		\end{tabular}}
		\psfrag{I}{\color{white}{\Large\textit{\textbf{Reconstruction}}}}
		\psfrag{J}{\begin{tabular}{@{}l@{}}
				\small Colored\\[-1.ex]
				\small Mesh
		\end{tabular}}
		\psfrag{K}{\begin{tabular}{@{}c@{}}
				\footnotesize\color{mybronze}{Mesh Extracting \&}\\[-1.ex]
				\footnotesize\color{mybronze}{Coloring}
		\end{tabular}}
		\psfrag{a}{Frame $i$}
		\psfrag{b}{\begin{tabular}{@{}l@{}}
				\color{mybronze}{Image}\\[-.5ex]
				\color{mybronze}{Encoding}
		\end{tabular}}
		\psfrag{c}{\begin{tabular}{@{}c@{}}
				\footnotesize\color{mybronze}{Latent Encoding}\\[-.5ex]
				\footnotesize\color{mybronze}{(\color{gray}{Sec. IV-D})}
		\end{tabular}}
		\psfrag{d}{\begin{tabular}{@{}c@{}}
				\color{mybronze}{Fusion}\\[-.5ex]
				\footnotesize\color{mybronze}{(\color{gray}{Sec. IV-E})}
		\end{tabular}}
		\psfrag{e}{\begin{tabular}{@{}l@{}}
				\scriptsize Global\\[-1.4ex]
				\scriptsize LIM for\\[-1.4ex]
				\scriptsize Latent
		\end{tabular}}
		\psfrag{f}{\color{white}{\Large\textit{\textbf{Scene understanding}}}}
		\psfrag{g}{\begin{tabular}{@{}c@{}}
				\small\color{mybronze}{Text}\\[-1.ex]
				\small\color{mybronze}{Encoding}
		\end{tabular}}
		\psfrag{h}{Text}
		\psfrag{i}{\begin{tabular}{@{}l@{}}
				\footnotesize Points\\[-1.4ex]
				\footnotesize to infer
		\end{tabular}}
		\psfrag{j}{\begin{tabular}{@{}l@{}}
				\footnotesize Point\\[-1.4ex]
				\footnotesize properties
			\end{tabular}}
		\psfrag{k}{\begin{tabular}{@{}c@{}}
				\footnotesize\color{mybronze}{Decoding}\\[-1ex]
				\footnotesize\color{mybronze}{(\color{gray}{Sec. III-A})}
		\end{tabular}}		
	}
	\caption{Reconstruction and scene understanding applications' pipeline. On the left incremental reconstruction application, external tracking runs in parallel to reconstruction to provide coarse poses. While doing reconstruction, internal tracking refines the pose estimation fur a better surface fit. ``$\cdots$'' means that we can add more other properties from \cref{sec:recons:other} and \cref{sec:fabircated_prop} into this pipeline. On the right scene understanding application, we assume that the pose of the frame is pre-known. The upper part of the white line is the fusion of LIM for the feature field. The lower part infers specific semantic information along with the text command.}
	\label{fig:recons_and_scene_understanding}
\end{figure*}

\label{sec:app_recons}

In this section, we present an application of incremental 3D reconstruction using RGB-D sequences.
Since RGB-D sequences provide both point positions and color values, it allows us to construct two types of LIMs: one for surface (\cref{sec:surface_mapping}), and one for color (\cref{sec:context_fields}).

The pipeline is illustrated in \cref{fig:recons_and_scene_understanding}.
When an RGB-D frame $i$ is fed into the framework, it is firstly converted into a colored point cloud ($\V X\in\mathbb R^{N\times 3}, \V Q\in\mathbb R^{N\times 3}$).
The tracking module takes $(\V X, \V Q)$ to estimate the current pose $\V T$.
Next, the transformed point cloud $\V X \V T^T$ is used as input to the surface mapping (\cref{sec:surface_mapping}), while the colored point cloud $(\V X \V T^T, \V Q)$ is used as input to the surface color mapping (\cref{sec:context_fields}), resulting in the generation of local LIMs.
Using the fusion operation in~\cref{eq:fuse}, the local LIM is fused into the global LIM on a voxel by voxel basis.

For visualization purposes, we first sample a grid within each voxel and infer using the global surface LIM to obtain the SDF.
The Marching Cube algorithm is then applied to extract the mesh.

Once the surface is reconstructed, we can sample points from it at arbitrary resolution and perform inference using the global color LIM to reconstruct the surface color.

\subsubsection{Tracking}
\label{sec:tracking}

According to Zhu et al.~\cite{zhu2022nice}, the current implicit scene representation-based tracking models, such as iMAP, DI-Fusion and NICE-SLAM, still have a performance gap compared to state-of-the-art tracking approaches such as BAD-SLAM and ORB-SLAM2.
Therefore, instead of following the neural implicit models to track with frame-to-model or ray-tracing based optimization, we incorporate ORB-SLAM2 in a separate thread to provide a pose prior. 
Hence, we call this external tracking.

Note that the primary focus of ORB-SLAM2 is on localization not scene reconstruction.
This means that direct use of ORB-SLAM2 provides coarse surface reconstruction.
We further use colored point cloud registration (CPCR)~\cite{park2017colored} as a tracking refiner.

In our implementation, ORB-SLAM2 runs independently over all frames.
Every few frames, CPCR tracks an initially posed colored point cloud to compute the odometry within a local window.
Mapping is then done in the same thread.
Therefore, the latter is called internal tracking.

\subsubsection{Other types of datas}
\label{sec:recons:other}

Application 4) uses a point cloud with infrared information, which is a straightforward modification of the color-based approach.
LIM feature dimension is also correspondingly reduced.
This flexibility allows different types of point cloud properties to be integrated into the continuous mapping pipeline.

\subsection{Application: 2D-to-3D Transfer}
\label{sec:fabircated_prop}

Applications such as 2) and 4) can be easily integrated with application 1) incremental reconstruction (\cref{sec:incremental_reconstruction}) by incorporating the fabricated result together with the point cloud.
For instance, given RGB-D frames, we detect saliency or transfer image styles to generate a fabricated $X$ image. Here, $X$ represents saliency, style, or other properties. 
By combining $X$ with depth information through unprojection,
we assign
the fabricated values to corresponding points, resulting in point pairs ($\V X$, $\V Q_{X}$).

Similar to the reconstruction pipeline in~\cref{fig:recons_and_scene_understanding}, we employ encoding (\cref{sec:encoder}) and fusion (\cref{eq:fuse}) to construct a global LIM for the fabricated properties $X$.
This global LIM represents a surface $X$ fields that is utilized for subsequent inference.

While it is possible to similarly transfer a 2D semantic image to 3D,
it may not be feasible in practice due to the need for multiple passes of different categories of semantic information 
on the same dataset (such as object, usability, etc.).
Therefore, in the following section, we demonstrate the construction of a surface feature field for scene understanding application that satisfies various 
requirements through a single mapping pass.

\subsection{Application: Open-vocabulary Scene Understanding}
\label{sec:openvoc}

This application follows OpenScene and CLIP-Field~\cite{peng2022openscene,shafiullah2022clip}, which learn to predict dense features for 3D scene points,
where the features are co-embedded with text and image pixels in CLIP feature space.
Inspired by this, we design the mapping for surface feature fields (\cref{sec:latent_fields}). 
The distinction between surface property fields is illustrated in~\cref{fig:latent_diff}.
We use the pre-trained OpenSeg model~\cite{ghiasi2022scaling} to obtain the function
$f_{im}$ that produce CLIP feature for each image point.
Then we encodes the voxel latent $\V F_m$ from the CLIP features $\V Q_m$ and their corresponding positions $\V X_m$ .
During inference, given an open-vocabulary text input $\pmb u$ and a position input $\V x_*$, we obtain the CLIP space features
and determine the semantic property of the point based on similarity computation. 
The pipeline of this application is illustrated in ~\cref{fig:recons_and_scene_understanding}.

In this application, the image encoding function $f_{im}$ and the text encoding function $f_{text}$ are obtained from pre-trained model, while $f_{enc}$ and $f_{dec}$ in our Uni-Fusion framework are deterministic functions.
With these functions, our model constructs a continuous field for the CLIP feature on surface.

A very interesting and relevant work for Uni-Fusion's scene understanding application is VLMaps~\cite{huang2023visual}.
While VLMaps produces a 2D map, our model produces a surface CLIP feature field in 3D space.

\section{Experiments}
\label{sec:exp}

In this section, we demonstrate the wide range of applications and the high capabilities of Uni-Fusion. 
First, we evaluate Uni-Fusion in application 1) Incremental surface and color reconstruction, comparing its performance with SOTAs.
For applications 2) and 5), which are new topics, no specific benchmarks are available. 
Therefore, we showcase the performance on existing results.
Next, we implement application 3) and compare it with SOTA zero-shot semantic segmentation models.
Finally, for application 4), since infrared data is not commonly used, we collect our own dataset containing infrared values and show all applications on this data.

\subsection{Implementation Details}
\label{sec:exp:details}

In the experiments, we use our sample-based GPIS for local geometry encoding.
For each point, two additional points are sampled along normal direction, one positive and one negative, with distance $d_s=0.1$ in the local voxel's normalized space. 
Compared to derivative-based GPIS, our sample-based GPIS is more efficient in both space and time. 
For the encoder, we randomly sample $256$ anchor points from the range $[-0.5,0.5]^3$.
We utilize the first $20$ eigenpairs, resulting in a feature dimension of $20$.
The model selection process is discussed in the ablation study.

Different latent maps use different granularities.
For the surface LIM, we use a voxel size of $5\si{\centi\meter}$. 
For color which requires later comparison to NeRF, we use a voxel size of $2\si{\centi\meter}$.
For other property LIM and feature LIM, we use a voxel size of $10\si{\centi\meter}$.

For smooth reconstruction, the encoded voxel is designed overlapped following~\cite{huang2021di}.
The encoded voxel uses twice the voxel size, resulting in a half-space overlap with each neighboring voxel.
During meshing, SDFs are retrieved and interpolated from the overlapped voxels~\cite{huang2021di}.
While for the remaining properties, we sample only from its own voxel part.

The implementation runs on a PC with AMD Ryzen 9 5950X 16-core CPU and an Nvidia Geforce RTX 3090 GPU (24 GB).
Online collaboration with OpenSeg (takes 15 GB GPU memory) utilizes one other 3090 GPU solely for OpenSeg to avoid out of memory.

\subsection{Datasets}

We evaluate incremental reconstruction on the ScanNet dataset~\cite{dai2017scannet}, TUM RGB-D dataset~\cite{sturm2012benchmark}, and Replica dataset~\cite{sucar2021imap}.
Using MSG-Net~\cite{zhang2018multi}'s material set, we transfer styles to the 3D canvas.
For open-vocabulary scene understanding, we evaluate on ScanNet segmentation data~\cite{qi2017pointnet++} and S3DIS dataset~\cite{armeni20163d}.

\subsubsection{ScanNet~\cite{dai2017scannet}}

ScanNet is a densely annotated RGB-D video dataset.
It is captured with the structure sensor~\cite{occipital} and contains 1513 scenes for training and validation.
For each scene, both images and a 3D mesh is provided, along with their 2D and 3D semantic annotations. 

ScanNet provides 312 scenes for validation, which contains a wide range of different room structures.
It has now been widely used in the thorough evaluation of the performance of reconstruction and semantic segmentation.

\subsubsection{TUM RGB-D~\cite{sturm2012benchmark}}

TUM RGB-D is a benchmark to mainly evaluate the tracking performance.
It is captured with Microsoft Kinect sensor together with ground-truth trajectory from the sensor.

\subsubsection{Replica~\cite{sucar2021imap}}

The Replica dataset refers to iMAP's pre-processed dataset~\cite{sucar2021imap}.
It is a synthetic rendered RGB-D dataset from given 3D models.
The advantage of including this dataset is that Replica does not have motion blur. 
This is better to evaluate the capability of the algorithms on reconstructing surface color.

\subsubsection{MSG-Net Style~\cite{zhang2018multi}}

MSG-Net provides material images for transfering the styles.
We select 21style fold for demonstration.
These images are given in \cref{fig:style} together with our result.

\subsubsection{ScanNet Point Cloud Segmentation Data~\cite{qi2017pointnet++}}

For point cloud semantic segmentation benchmarking, PointNet++~\cite{qi2017pointnet++} preprocesses the original ScanNet~\cite{dai2017scannet} and generates subsampled point clouds and corresponding annotations for each scene.

\subsubsection{S3DIS~\cite{armeni20163d} and 2D-3D-S~\cite{armeni2017joint}}

S3DIS is a semantic segmentation dataset for 3D point clouds.
Which is also a subset of the 2D-3D-S dataset.
The 2D-3D-S dataset is a multi-modality dataset containing 2D, 2.5D and 3D domains. 
This dataset is densely annotated with semantic labels.

Note that 2D-3D-S's 2D captures is not a RGB-D video as ScanNet.
2D-3D-S's images only have small overlap. 
Therefore, it is only suitable for semantic segmentation and not for incremental reconstruction.

\subsection{Baselines}

For online surface mapping evaluation, we select TSDF-Fusion~\cite{curless1996volumetric}, iMAP~\cite{sucar2021imap}, SOTA DI-Fusion~\cite{huang2021di} and BNV-Fusion~\cite{li2022bnv} as four baseline methods.

For the color field, we choose TSDF-Fusion~\cite{curless1996volumetric}, $\sigma$-Fusion~\cite{rosinol2023probabilistic}, iMAP~\cite{sucar2021imap}, NICE-SLAM~\cite{zhu2022nice} and even the recent hot Neural Radiance Fields model NeRF-SLAM~\cite{rosinol2022nerf} as five baselines.
While including NeRF in the comparison may not be entirely fair, we want to show how Uni-Fusion narrows the performance gap.

For the scene understanding application, we evaluate generalized zero-shot point cloud semantic segmentation with ZSLPC~\cite{cheraghian2019zero}, DeViSe~\cite{frome2013devise} and SOTA 3DGenZ~\cite{michele2021generative} for comparison.

\subsection{Metrics}

For incremental reconstruction, we evaluate the geometric reconstruction using \textbf{Accuracy}, \textbf{Completeness}, and \textbf{F1 score} according to SOTA BNV-Fusion. It firstly uniformly samples $100,000$ points from the reconstruction and ground truth meshes respectively.
Then \textbf{Accuracy} (\textbf{Completeness}) measures the percentage of reconstruction-to-groundtruth (groundtruth-to-reconstruction) distances that are lower than $2.5\si{\centi\meter}$ threshold. \textbf{F1 score} is the harmonic mean of accuracy and completeness.
For tracking performance, we use \textbf{ATE RMSE}.

To evaluate color reconstruction, we follow SOTA on this topic, NeRF to render both depth and RGB images to compare the image level \textbf{Depth L1} and \textbf{RGB PSNR}.

To compare scene understanding, we follow zero-shot point cloud semantic segmentation SOTA 3DGenZ to evaluate the \textbf{Intersection-of-Union (IoU)} and \textbf{Accuracy}.

\subsection{Reconstruction Results}

In our evaluation, we first use the ScanNet validation set with 312 sequences to thoroughly test the geometric reconstruction on a wide variety of scenes.
Then, we use TUM RGB-D to compare our modified tracking model with related works.
Since the tracking part is not the contribution of this paper, we give a brief overview of the tracking results.
To fairly evaluate the color reconstruction, we compare with related works, including NeRF, on the high-quality rendered Replica dataset.
\begin{table*}[t!]
	\centering
	\caption{Surface comparison on ScanNet~\cite{dai2017scannet}.
		Scores are fetched from~\cite{li2022bnv}.
		$^*$ indicates the result from our runs of the official BNV-Fusion code.}
	\small
	\setlength{\tabcolsep}{0.6em}
		\begin{tabular}{l  c c c| c c c }
			\toprule
			Method & \begin{tabular}{@{}c@{}}Pre-Train\\ with extra dataset\end{tabular} & \begin{tabular}{@{}c@{}}Train \\ with sequences\end{tabular} & Real-time & Accuracy (\%) $\uparrow$ & Completeness (\%) $\uparrow$ & F1 score $\uparrow$ \\
			\midrule
			TSDF Fusion~\cite{zhou2018open3d} & None & None & $\checkmark$ &73.83 & 85.85 & 78.84 \\
			iMAP~\cite{sucar2021imap} & None & Online train& &68.96 & 82.12 & 74.96 \\
			DI-Fusion~\cite{huang2021di} &Object Pretrain & None & $\checkmark$&66.34 & 79.65 & 72.97 \\
			BNV-Fusion~\cite{li2022bnv} &Object Pretrain &  Post Optimization& &{74.90} & \textbf{88.12} & {80.56} \\
			BNV-Fusion$^{*}$~\cite{li2022bnv} &Object Pretrain & Post Optimization &&{73.42} & {81.75} & {77.18} \\
			\textbf{Uni-Fusion (Ours)} &None &None &$\checkmark$&\textbf{80.43} & {84.91} & \textbf{82.44} \\
			\bottomrule
		\end{tabular}
	\label{tab:scannet}
	\vspace{-0.3cm}
\end{table*}

\begin{figure*}[t!]
	\subfloat[width=.3\textwidth][Accuracy]{
		\centering
		\includegraphics[width=.2\linewidth]{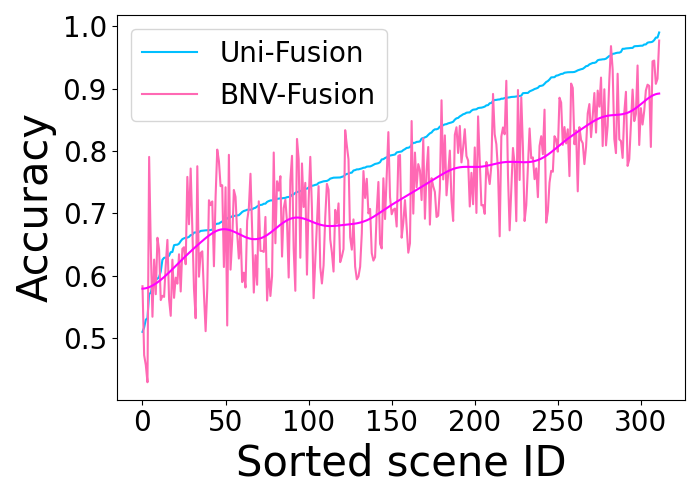}
		\includegraphics[width=.1\linewidth]{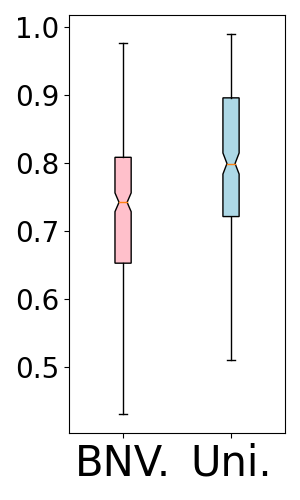}
	}
	\subfloat[width=.3\textwidth][Completeness]{
		\centering
		\includegraphics[width=.2\linewidth]{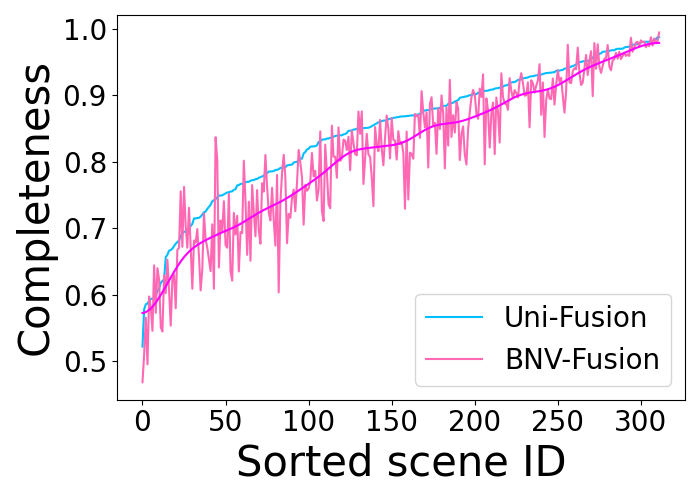}
		\includegraphics[width=.1\linewidth]{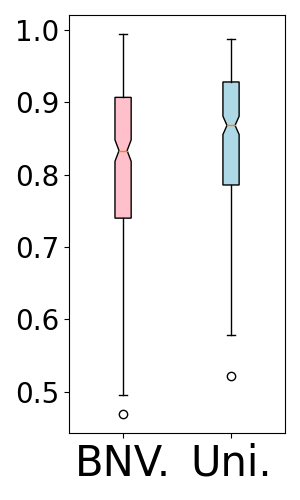}
	}
	\subfloat[width=.3\textwidth][F1 score]{
		\centering
		\includegraphics[width=.2\linewidth]{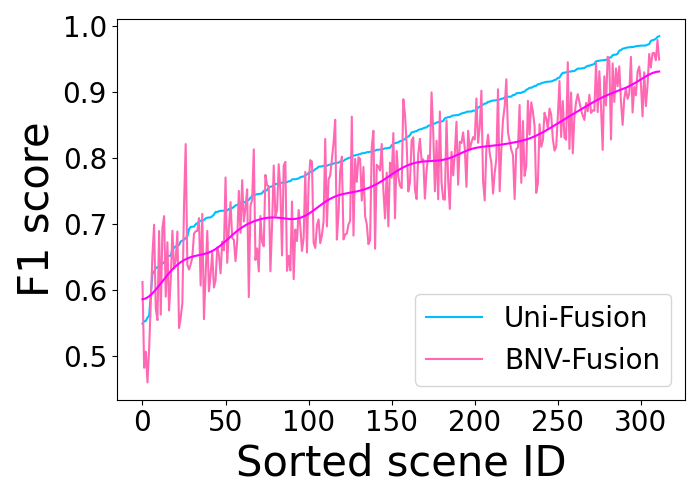}
		\includegraphics[width=.1\linewidth]{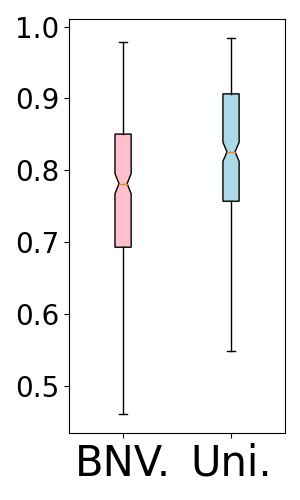}
	}
	\label{fig:recon:scannet:elementwise}
	\caption{Quantitative comparison on 312 scenes of the ScanNet validation set.
		We demonstrate the performance of SOTA BNV-Fusion and our Uni-Fusion.
		We sort Uni-Fusion's evaluation value and reordered all of the scores.
		The zigzag pink is the BNV-Fusion result;
		we also plot a deep-pink smoothed curve for better visualization.}
	\vspace{-0.3cm}
\end{figure*}

\subsubsection{Evaluation on ScanNet Dataset~\cite{dai2017scannet}}
\label{sec:exp:scannet}

We use the 312 different scenes from the ScanNet validation set to evaluate the performance of surface reconstruction. 
We follow the pure mapping SOTA BNV-Fusion to take every 10th posed frame as input. 
Without using any learning (as iMAP, DI-Fusion, and BNV-Fusion do) or any post-optimization (as BNV-Fusion does), our Uni-Fusion is capable to achieve precise continuous mapping performance.

The results presented in~\cref{tab:scannet} demonstrate that our Uni-Fusion outperforms the SOTA method BNV-Fusion in terms of \textbf{$+6$ higher accuracy}. 
However, our model does not outperform BNV-Fusion in terms of completeness, since BNV-Fusion incorporates completion in post-optimization.
Nevertheless, Uni-Fusion's completion is still much higher than one other optimization based model iMAP.
Overall, our Uni-Fusion model achieves higher F1-scores, which quantifies the overall quality of reconstruction.

It is important to note that the SOTA BNV-Fusion is not capable of real-time performance as it requires post-optimization of all fed frames.
On the other hand, the real-time model Di-Fusion exhibits much worse results without using post-optimization.
In contrast, our \textbf{real-time} model, \textbf{Uni-Fusion} achieves \textbf{much better} reconstruction quality than these approaches even without post-optimization. 

We additionally run BNV-Fusion's official implementation (emphasized with $^*$) on the 312 videos from ScanNet and conduct a scene-wise comparison in \cref{fig:recon:scannet:elementwise}. 
Our result is shown with the {\color{Cyan}light blue} curve, BNV-Fusion's result is shown with {\color{Lavender}pink}.
The scene index is sorted based on the score value of Uni-Fusion.
To enhance visual clarity, we apply a smoothing technique to BNV-Fusion's curve and presented it with dark pink color.
The comparison clearly shows that Uni-Fusion is overall performing better than BNV-Fusion. 
In addition, we use box-plots to analyze the statistics alongside the curve plot. Uni-Fusion's scores show a higher concentration on the plots. 
While the difference in completeness may be less pronounced,
Uni-Fusion's box plot is smaller and positioned relatively higher. This indicates that Uni-Fusion achieves a more stable completeness result, while BNV-Fusion is more likely to achieve low completeness in some cases.

In summary, our Uni-Fusion model achieves superior results across almost all 312 scenes in terms of accuracy, completeness and F1-score.
This finding is consistent with the observation presented in~\cref{tab:scannet} with BNV-Fusion$^*$, that Uni-Fusion outperforms the official implementation of BNV-Fusion in all metrics.

We show reconstruction on selected scenes from ScanNet in~\cref{fig:recons:scannet_demo}. 
Both BNV-Fusion and our Uni-Fusion are able to produce high quality reconstructions.
However, we observe that BNV-Fusion generates numerous small meshes on walls, resulting in the appearance of small particles in the reconstruction. 
We attribute this behavior to BNV-Fusion's use of very small voxel size ($0.02\si{\meter}$) to achieve a high score.
This is further supported by the fact that their mesh averages \textbf{\SI{247}{MB}}, while ours averages only \textbf{\SI{54}{Mb}}.
Furthermore, our Uni-Fusion's mesh is smoother and
also provides highly-accurate color to the mesh which is not available for this surface SOTA.

Besides, we evaluate the storage cost of the latent representation. BNV-Fusion's latents require an average storage of \textbf{\SI{228}{MB}} across the 312 scenes of ScanNet, while Uni-Fusion achieves significantly lower storage requirements with an average of only \textbf{\SI{9}{MB}}.

\newcommand{\scannetImSize}{.15}
\begin{figure*}[t!]
	\centering
	\setlength{\tabcolsep}{0.1em}
	\renewcommand{\arraystretch}{.1}
	\begin{tabular}{|c | c |c |||c |c | c|}
		\toprule	
		{\Large{BNV-Fusion}} & {\Large{Uni-Fusion}} &{\Large{Ground Truth}} & {\Large{BNV-Fusion}} &{\Large{Uni-Fusion}} & {\Large{Ground Truth}} \\ 
		\midrule
		
		\includegraphics[width=\scannetImSize\linewidth]{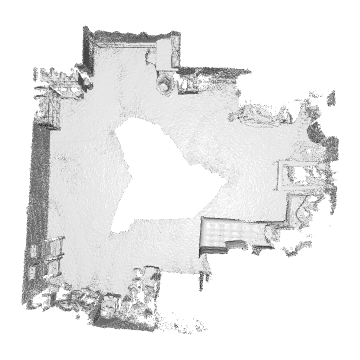}
		&\includegraphics[width=\scannetImSize\linewidth]{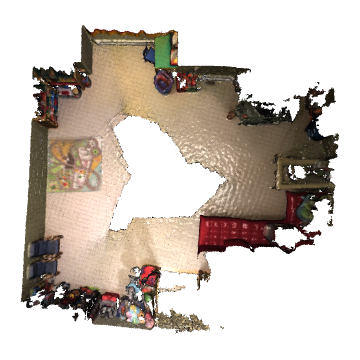}
		&\includegraphics[width=\scannetImSize\linewidth]{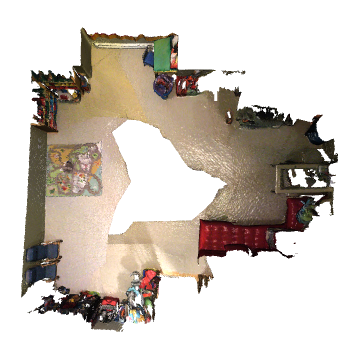}
		&		\includegraphics[width=\scannetImSize\linewidth]{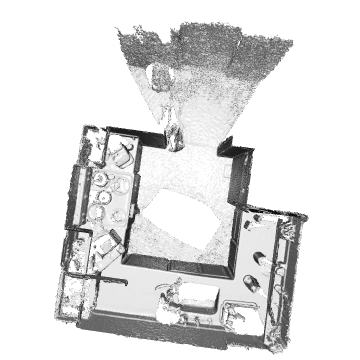}
		&\includegraphics[width=\scannetImSize\linewidth]{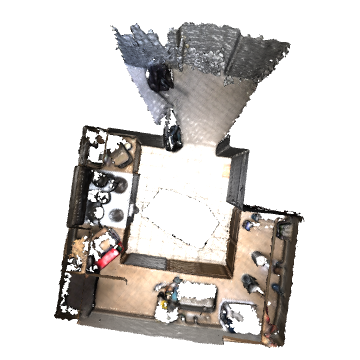}
		&\includegraphics[width=\scannetImSize\linewidth]{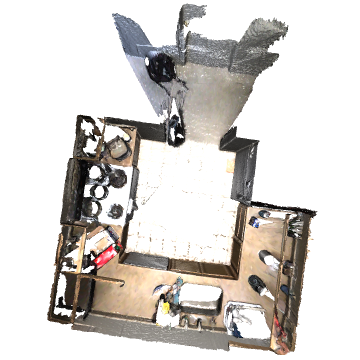}\\
		
		\includegraphics[width=\scannetImSize\linewidth]{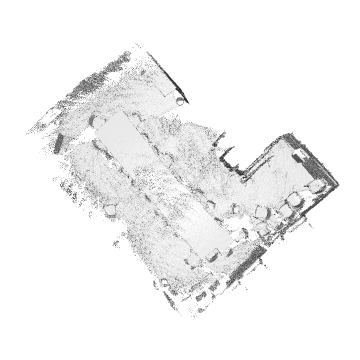}
		&\includegraphics[width=\scannetImSize\linewidth]{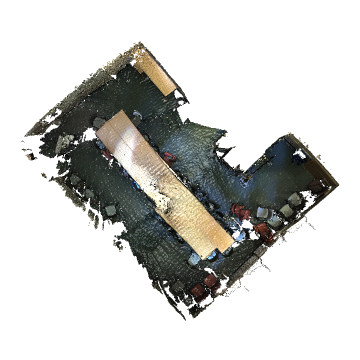}
		&\includegraphics[width=\scannetImSize\linewidth]{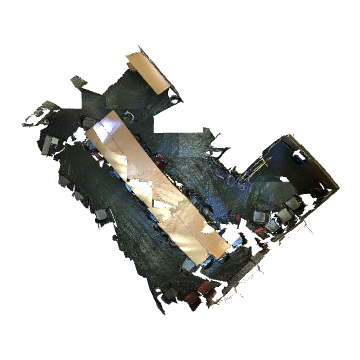}
		&		\includegraphics[width=\scannetImSize\linewidth]{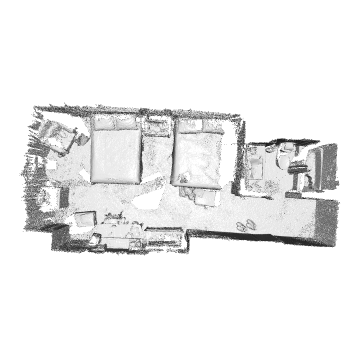}
		&\includegraphics[width=\scannetImSize\linewidth]{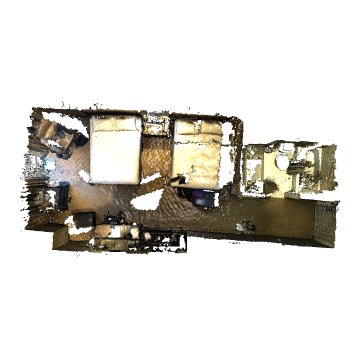}
		&\includegraphics[width=\scannetImSize\linewidth]{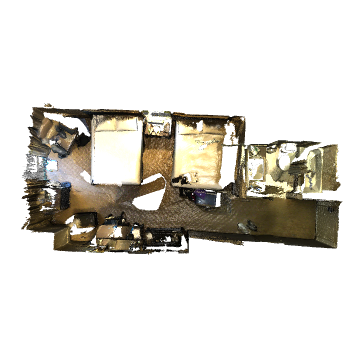}\\
		
		\includegraphics[width=\scannetImSize\linewidth]{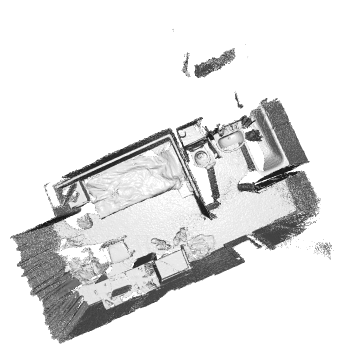}
		&\includegraphics[width=\scannetImSize\linewidth]{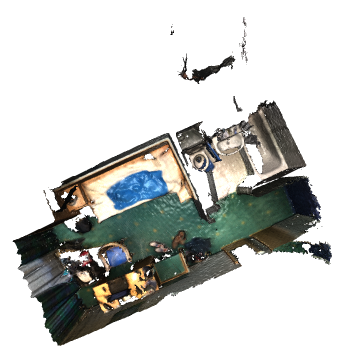}
		&\includegraphics[width=\scannetImSize\linewidth]{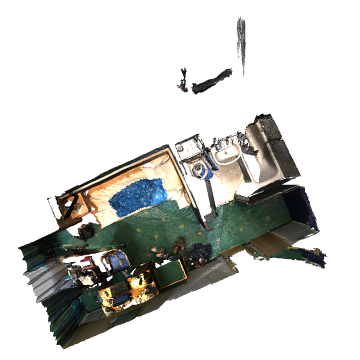}
		&		\includegraphics[width=\scannetImSize\linewidth]{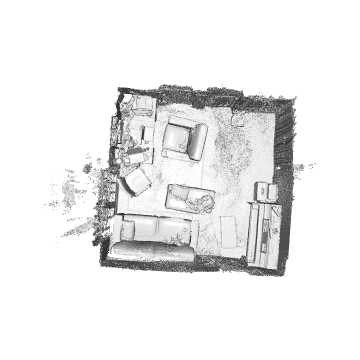}
		&\includegraphics[width=\scannetImSize\linewidth]{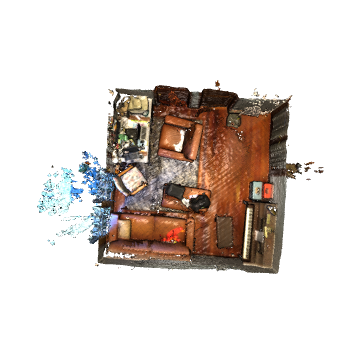}
		&\includegraphics[width=\scannetImSize\linewidth]{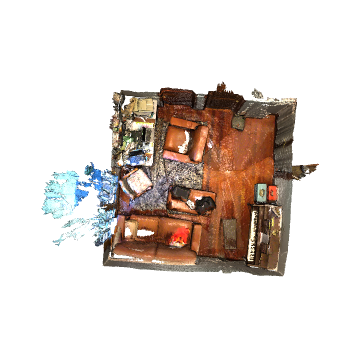}\\
		\hline
	\end{tabular}
	\caption{Surface reconstruction on ScanNet dataset.}
	\label{fig:recons:scannet_demo}
\end{figure*}

\subsubsection{Tracking Evaluation on TUM RGB-D Dataset~\cite{sturm2012benchmark}}
In the above test, we evaluate the performance of pure mapping.
Although tracking is not the contribution focus in our paper, it is still part of the incremental reconstruction model. We follow the novel incremental reconstruction model NICE-SLAM~\cite{zhu2022nice} to evaluate the camera tracking on the small scale TUM RGB-D dataset.
Our Uni-Fusion uses a coarse-to-fine strategy for 3D reconstruction tracking.
From~\cref{tab:tum_rmse}, it demonstrates overall better ATE RMSE than other implicit representation models.
\begin{table}[htbp]
	\caption{Tracking on TUM RGB-D~\cite{sturm2012benchmark}.
		ATE RMSE [$\si{\centi\meter}$] ($\downarrow$) is used as the evaluation metric.
	}
	\centering
	\footnotesize
	\setlength{\tabcolsep}{0.7em}
	\resizebox{\linewidth}{!}{
		\begin{tabular}{l|ccc}
			\hline
			& \tt{fr1/desk} &  \tt{fr2/xyz} &  \tt{fr3/office} \\
			
			\hline
			{iMAP}~\cite{sucar2021imap}      & 4.9 & 2.0 & 5.8  \\
			{DI-Fusion~\cite{huang2021di}} & 4.4 & 2.3 & 15.6 \\
			NICE-SLAM~\cite{zhu2022nice}           & 2.7 & 1.8 & 3.0 \\
			Ours& 1.8& 0.5& 2.1 \\
			\hline
			{BAD-SLAM}\cite{schops2019bad} & 1.7  & 1.1  & 1.7 \\
			{Kintinuous}\cite{whelan2012kintinuous} & 3.7  &  2.9  & 3.0 \\
			{ORB-SLAM2}\cite{mur2017orb} & \bf 1.6  & \bf 0.4  & \bf 1.0 \\
			\hline
	\end{tabular}}
	\vspace{2pt}
	
	\label{tab:tum_rmse}
\end{table}

On the other hand, there also exist high accuracy algorithms from SLAM. 
By additionally using bundle adjustment and loop-closing techniques, their tracking quality is much better than all of the implicit based models.
Our coarse-to-fine strategy obtains a good score because, first it ensures it does not easly lose track. Second, it is more suitable for surface fitting.

This further supports our test on the Replica dataset.

\begin{table*}[t!]
	\centering
	\caption{Geometric (L1) and Photometric (PSNR) evaluation on the Replica dataset~\cite{sucar2021imap}.}
	\footnotesize
	\setlength{\tabcolsep}{0.36em}
	\renewcommand{\arraystretch}{1.2}
	\begin{tabular}{clcccccccccccccccccc}
		\toprule
		& & \multicolumn{1}{c}{\makecell{\tt{office-0}}} & \multicolumn{1}{c}{\makecell{\tt{office-1}}} & \multicolumn{1}{c}{\makecell{\tt{office-2}}}& \multicolumn{1}{c}{\makecell{\tt{office-3}}} & \multicolumn{1}{c}{\makecell{\tt{office-4}}} & \multicolumn{1}{c}{\makecell{\tt{room-0}}} & \multicolumn{1}{c}{\makecell{\tt{room-1}}} &  \multicolumn{1}{c}{\makecell{\tt{room-2}}} & Avg. \\
		\midrule
		\multicolumn{5}{l}{\textit{Non-continuous mapping method}}\\
		\multirow{2}{*}{\makecell{\textbf{TSDF-Fusion}~\cite{curless1996volumetric}}}
		& {\bf Depth L1} [$\si{\centi\meter}$] $\downarrow$
		& 14.11 & 10.50 & 30.89 & 28.92 & 22.83	& 23.51 & 20.94 & 23.34 & 21.88 \\
		& {\bf PSNR } [$\si{\dB}$] $\uparrow$
		& 11.16 & 15.92 & 4.86 & 5.68 & 5.46 & 3.43 & 4.51 & 5.57 & 7.07 \\
		
		\midrule
		\multirow{2}{*}{\makecell{\textbf{$\sigma$-Fusion}\cite{rosinol2023probabilistic} }}
		& {\bf Depth L1} [$\si{\centi\meter}$] $\downarrow$
		& 13.80 & 10.21 & 22.27 & 28.70 & 22.21& 21.92 & 19.28 & 22.40 & 20.10 \\
		& {\bf PSNR } [$\si{\dB}$] $\uparrow$
		& 11.16 & 15.92 & 4.86 & 5.69 & 5.46& 3.45  & 4.51 & 5.57 & 7.08 \\

		\midrule
		\midrule
		\multicolumn{5}{l}{\textit{Continuous mapping method}}\\
		\multirow{2}{*}{\makecell{\textbf{iMAP$^*$}~\cite{sucar2021imap}}}
		& {\bf Depth L1} [$\si{\centi\meter}$] $\downarrow$
		& 6.43 & 7.41 & 14.23 & 8.68 & 6.80& 5.70 & 4.93 & 6.94 & 7.64\\
		& {\bf PSNR } [$\si{\dB}$] $\uparrow$
		& 7.39 & 11.89 & 8.12 & 5.62 & 5.98& 5.66 & 5.31 & 5.64  & 6.95\\
		\midrule
		\multirow{2}{*}{{\makecell{\textbf{Nice-SLAM}~\cite{zhu2022nice} }}}
		& {\bf Depth L1} [$\si{\centi\meter}$] $\downarrow$
		& { 1.51 } & { 0.93 } & { 8.41 } & { 10.48 } & {2.43} & { 2.53 } & { 3.45 } & { 2.93 }  & { 4.08 } \\
		& {\bf PSNR } [$\si{\dB}$] $\uparrow$
		& { 22.44 } & { 25.22 } & { 22.79 } & { 22.94 } & { 24.72 } & \textbf{ 29.90 } & \textbf{ 29.12 } & { 19.80 }& { 24.61 } \\

		\midrule	
		\multirow{2}{*}{{\makecell{\textbf{Uni-Fusion} (Ours) }}}
		& {\bf Depth L1} [$\si{\centi\meter}$] $\downarrow$
		& \textbf{0.79}&\textbf{0.56}&\textbf{1.59}&\textbf{2.71}&\textbf{1.66}&\textbf{1.94}&\textbf{0.69}&\textbf{1.80}& \textbf{1.47}
		\\
		& {\bf PSNR } [$\si{\dB}$] $\uparrow$ &\textbf{33.88}&\textbf{33.31}&\textbf{25.84}&\textbf{26.01}&\textbf{28.14}&24.02&26.20&\textbf{27.17} &\textbf{28.07}
		\\
		
		\midrule
		\midrule
		\multicolumn{5}{l}{\textit{Neural radiance field method}}\\
		\multirow{2}{*}{{\makecell{\textbf{NeRF-SLAM}~\cite{rosinol2022nerf} }}}
		& {\bf Depth L1} [$\si{\centi\meter}$] $\downarrow$
		& {2.49}   & {1.98}  & {9.13}  & {10.58} & {3.59}	& {2.97}  & {2.63}  & {2.58}  & {4.49} \\
		& {\bf PSNR } [$\si{\dB}$] $\uparrow$
		& \textbf{48.07}  & \textbf{53.44} & \textbf{39.30} & \textbf{38.63} & \textbf{39.21} 	& \textbf{34.90} & \textbf{36.95} & \textbf{40.75}& \textbf{41.40} \\
		
		\bottomrule
	\end{tabular}%
	
	\label{tab:replica_per_scene}
\end{table*}

\begin{table*}[t!]
	\centering
	\caption{Differences among different Surface \& Color reconstruction models.}
	\small
	\setlength{\tabcolsep}{.3em}
			\begin{tabular}{l | c c c c c c }
				\toprule
				Method & 
				\begin{tabular}{@{}c@{}}Pre-Train\\ with extra dataset\end{tabular}
				& \begin{tabular}{@{}c@{}}Train\\ with sequences\end{tabular}
				& Real-time	
				& Direct Output &  \begin{tabular}{@{}c@{}}Light\\ direction\end{tabular} 
				&Render\\
				\hline 
				TSDF-Fusion & None & None & $\checkmark$& Discrete TSDF &  &Ray Rasterization\\\hline
				$\sigma$-Fusion & None & None &$\checkmark$&Discrete TSDF  && Ray Rasterization\\\hline
				iMAP & None & Online Train && MLPs  & &Volumetric Rendering\\\hline
				NICE-SLAM & \begin{tabular}{@{}c@{}}Pretrain\\ with indoor dataset\end{tabular} & Online Train&& Neural Implicit Grid&  & Volumetric Rendering\\\hline
				
				NeRF-SLAM & None & Train hundred epochs &-&NeRF & $\checkmark$ &Volumetric Rendering \\\hline
				
				\textbf{Uni-Fusion} & None & None&$\checkmark$& Latent Implicit Map && Ray Rasterization\\				
				\hline
			\end{tabular}
	\label{tab:replica_diff}
\end{table*}
\subsubsection{Evaluation on Replica RGB-D Dataset~\cite{sucar2021imap}}

\newcommand{\replicaImSize}{.23}
\begin{figure*}[b!]
	\centering
	\setlength{\tabcolsep}{0.1em}
	\renewcommand{\arraystretch}{.1}
	\begin{tabular}{|c | c |c |c| }
		\toprule
		{\Large{NICE-SLAM}} &{\Large{NeRF-SLAM}}&\textbf{\Large{Uni-Fusion}}&\Large{Ground Truth}\\
		
		\midrule
		\includegraphics[width=\replicaImSize\linewidth]{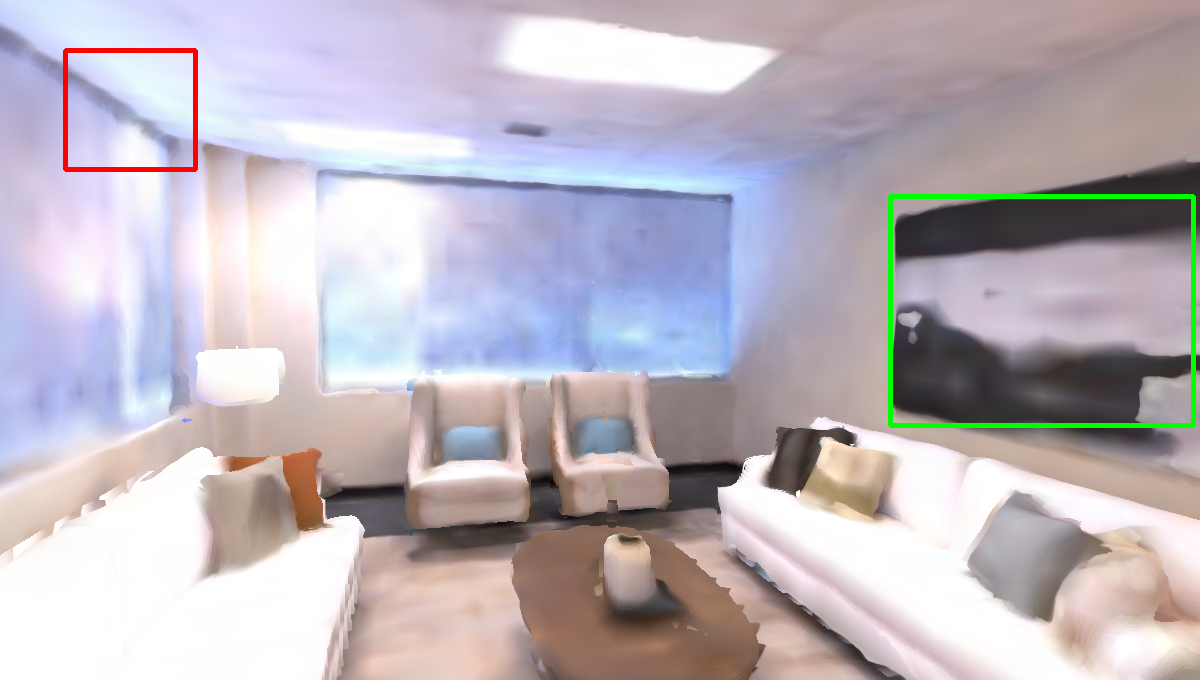} &
		\includegraphics[width=\replicaImSize\linewidth]{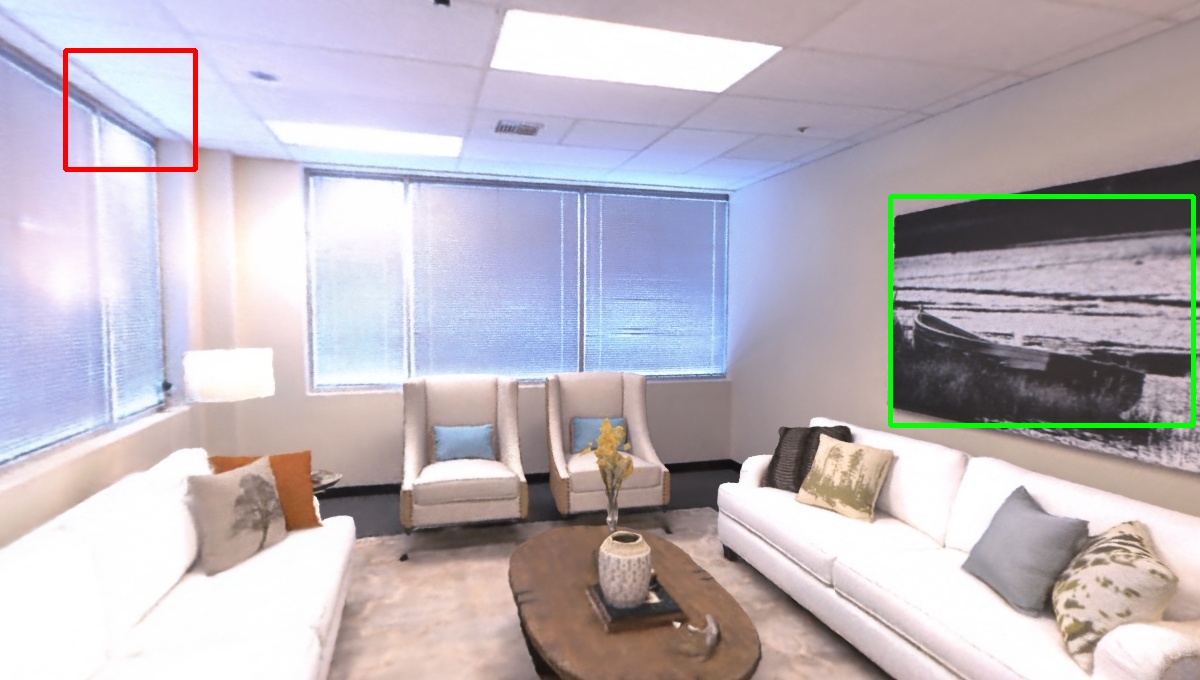} &
		\includegraphics[width=\replicaImSize\linewidth]{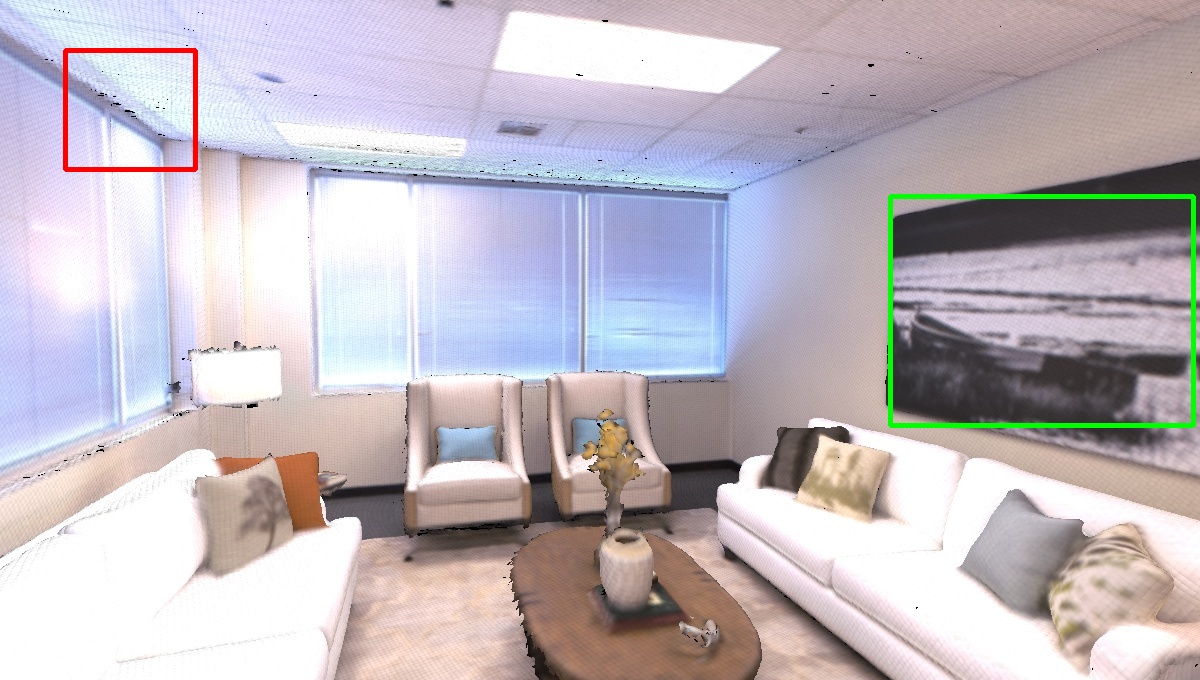} &
		\includegraphics[width=\replicaImSize\linewidth]{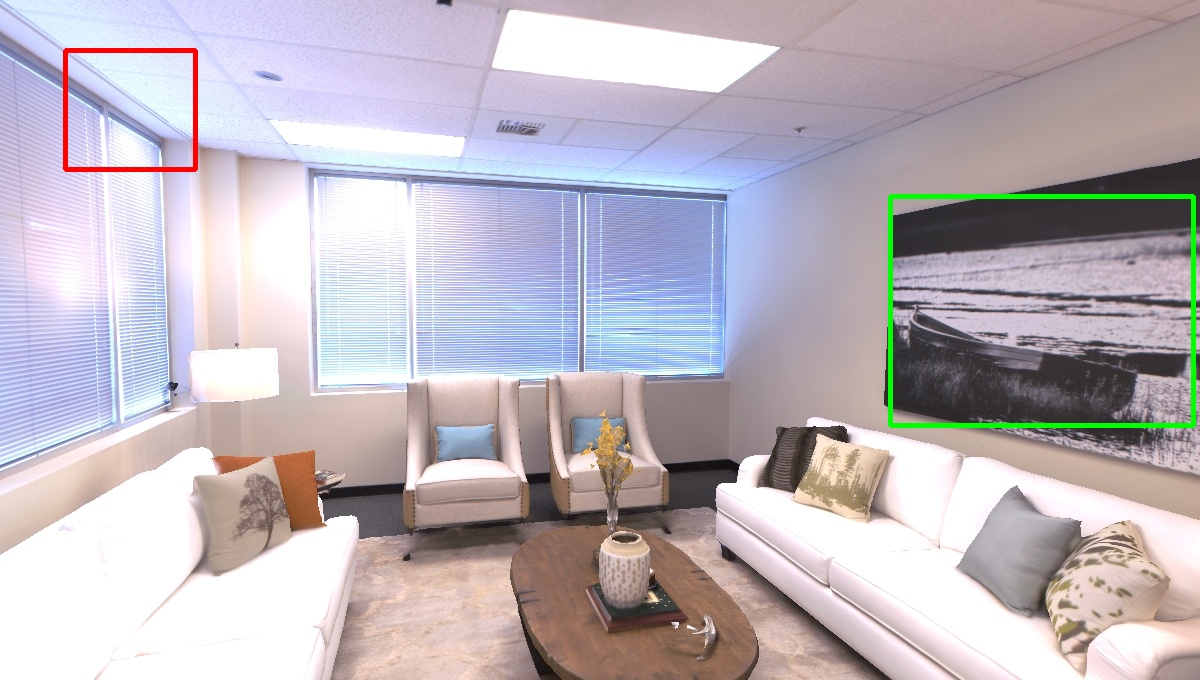} \\
		\hline
		\includegraphics[width=\replicaImSize\linewidth]{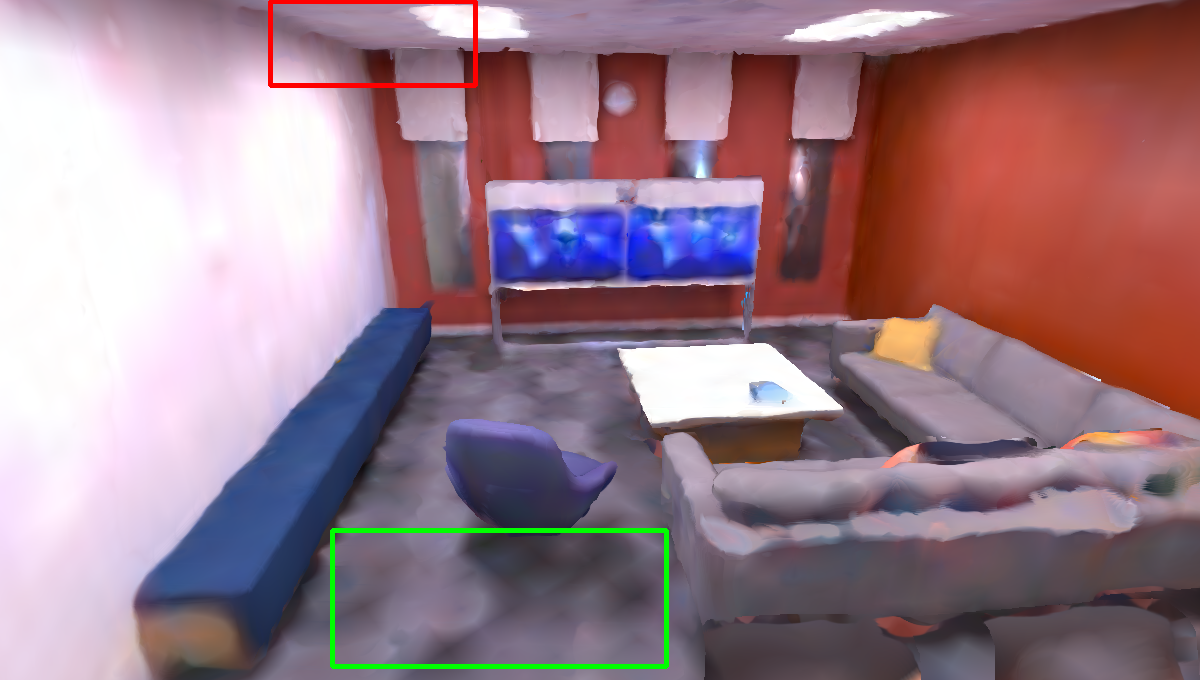} &
		\includegraphics[width=\replicaImSize\linewidth]{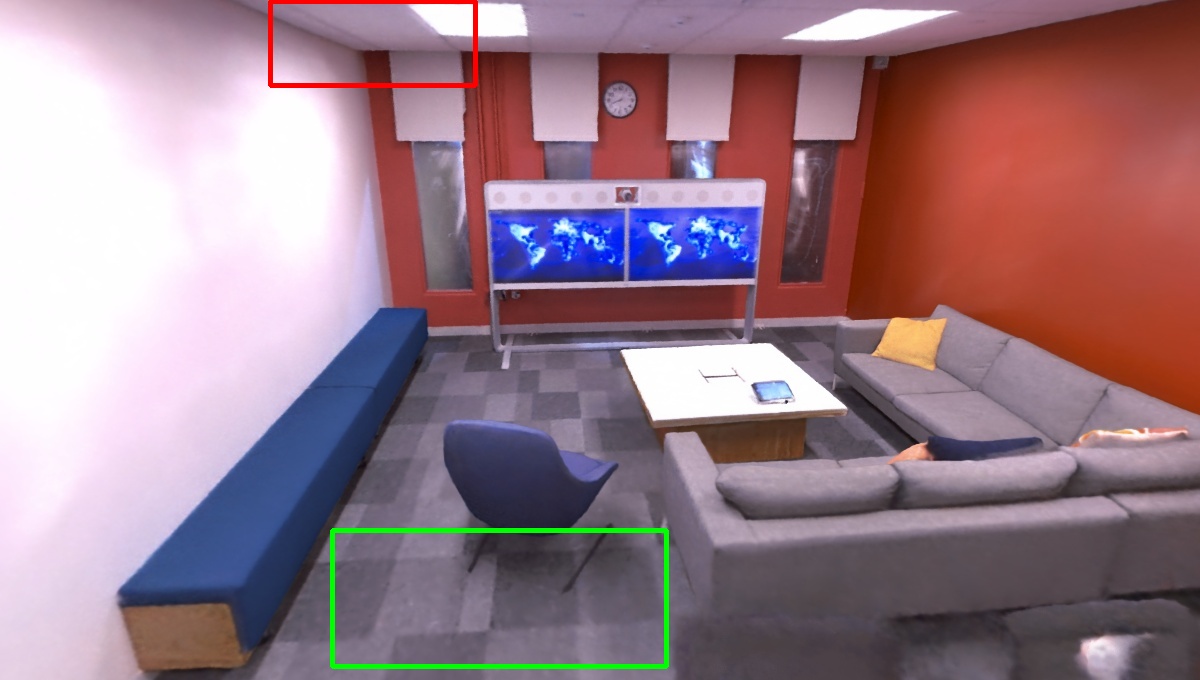} &
		\includegraphics[width=\replicaImSize\linewidth]{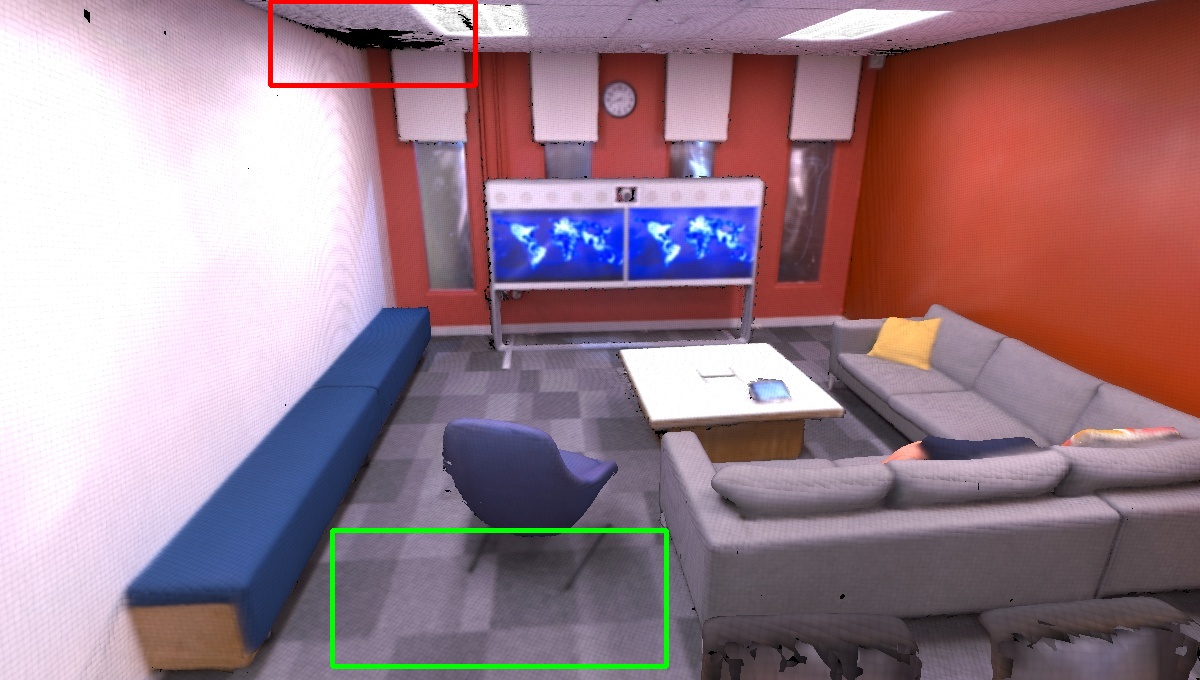} &
		\includegraphics[width=\replicaImSize\linewidth]{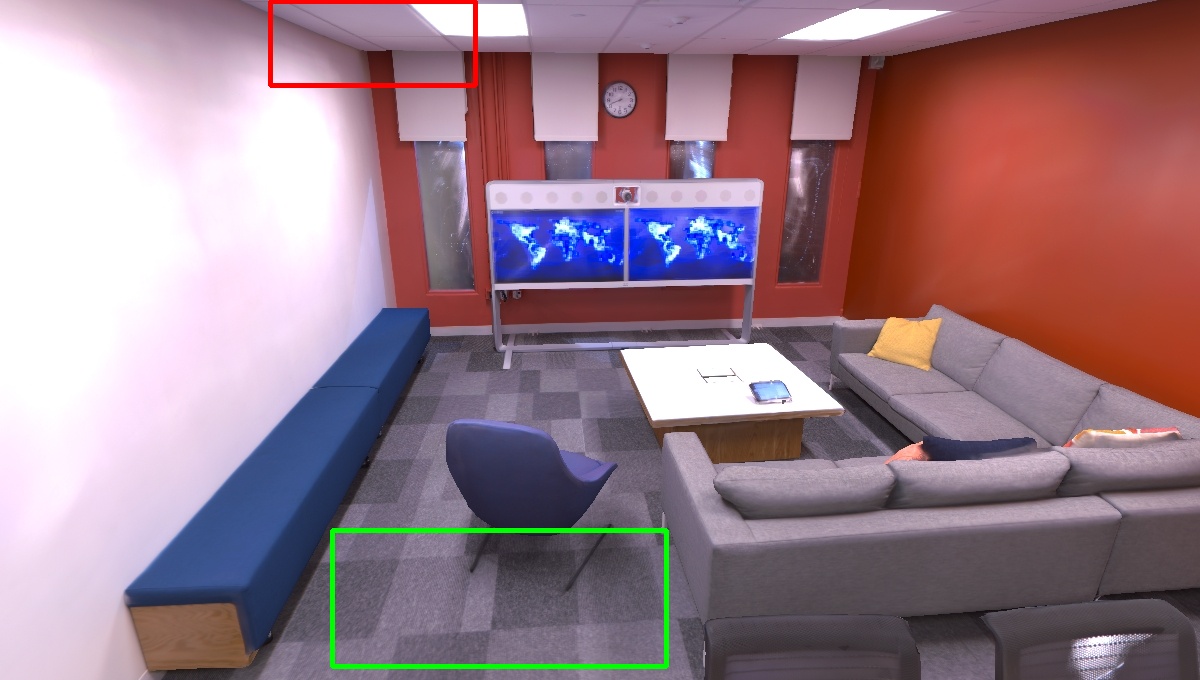} \\
		\hline
		\includegraphics[width=\replicaImSize\linewidth]{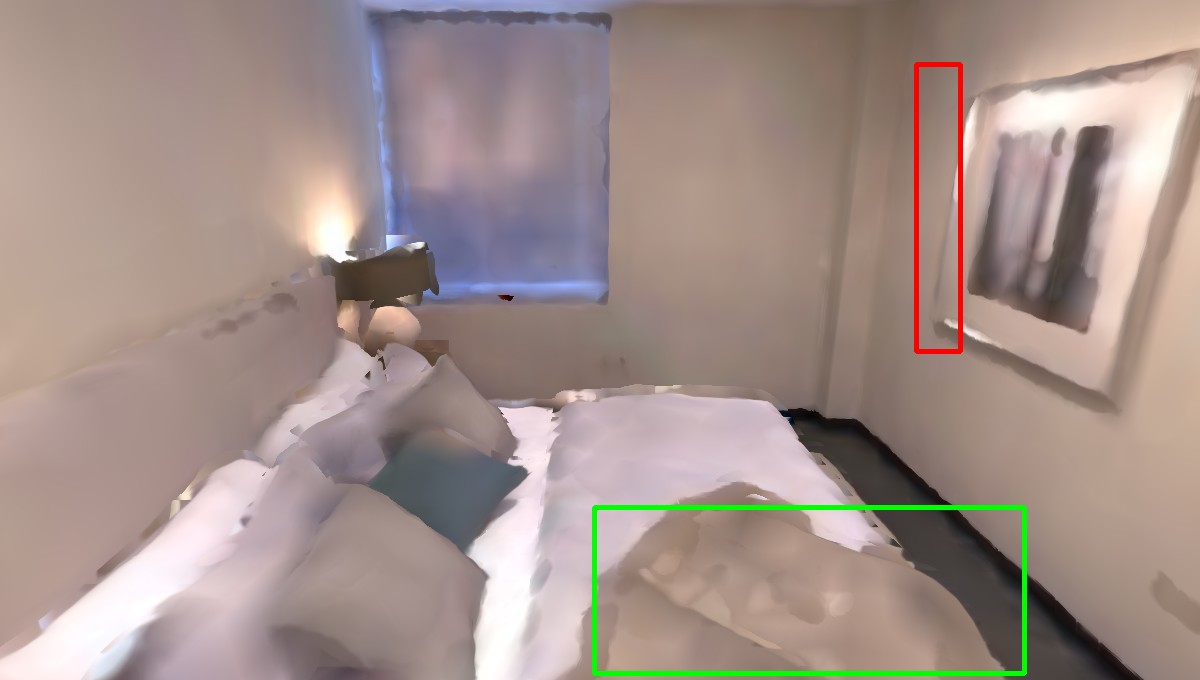} &
		\includegraphics[width=\replicaImSize\linewidth]{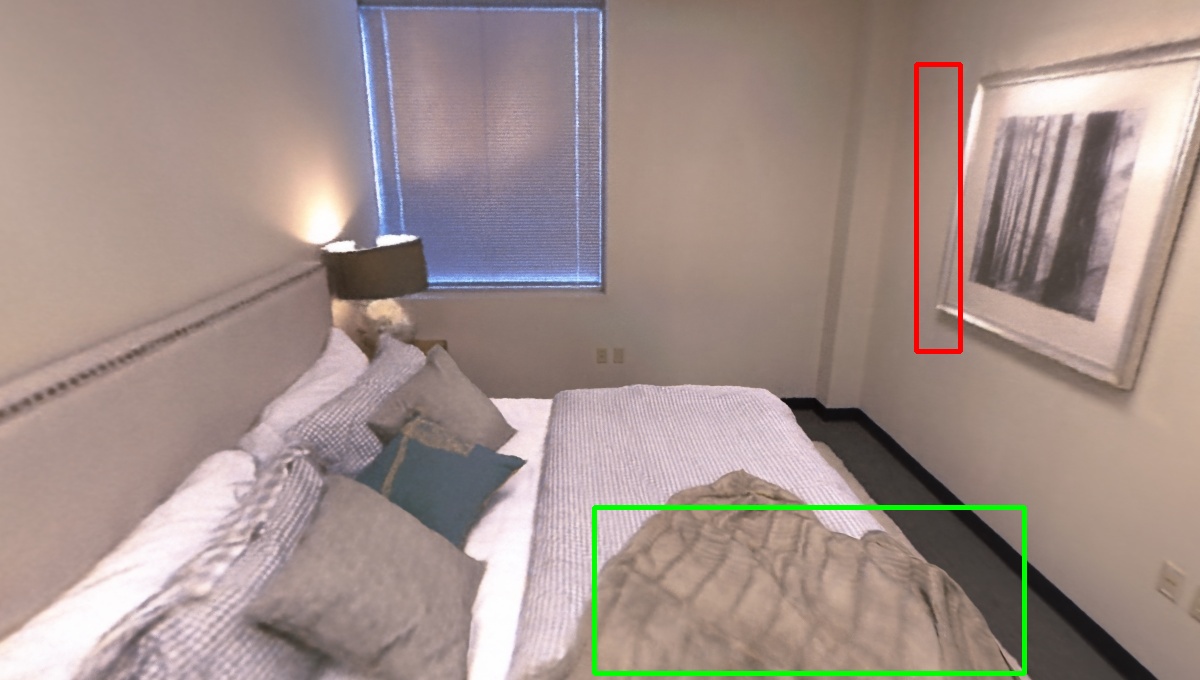} &
		\includegraphics[width=\replicaImSize\linewidth]{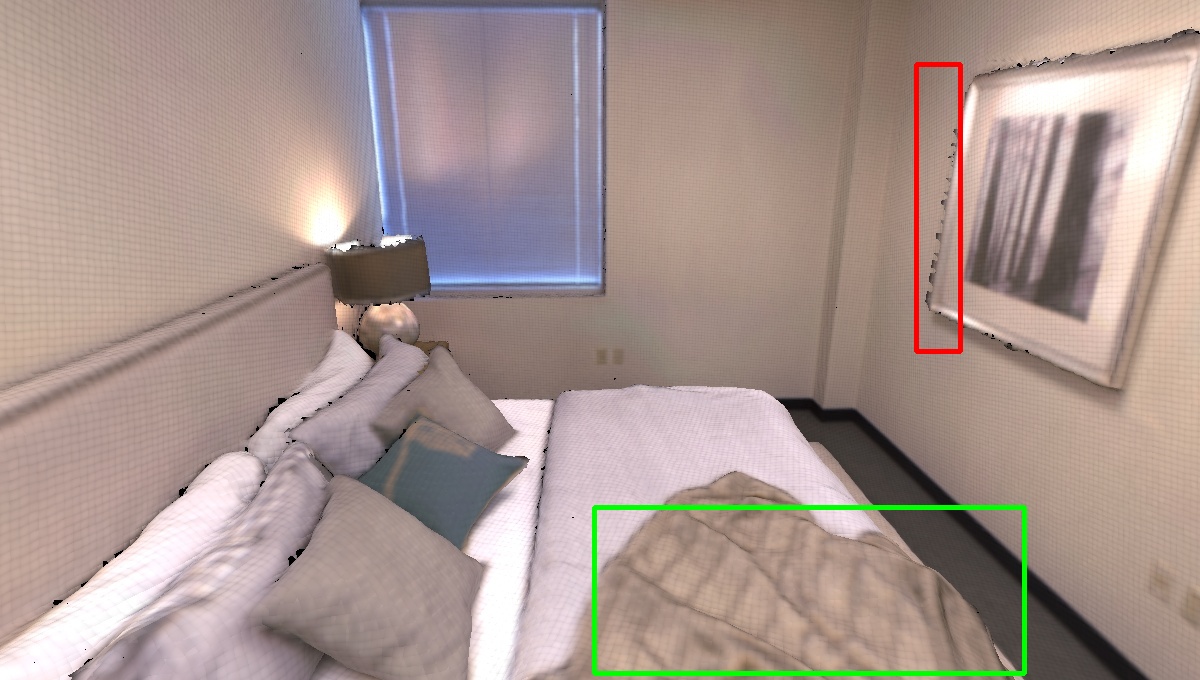} &
		\includegraphics[width=\replicaImSize\linewidth]{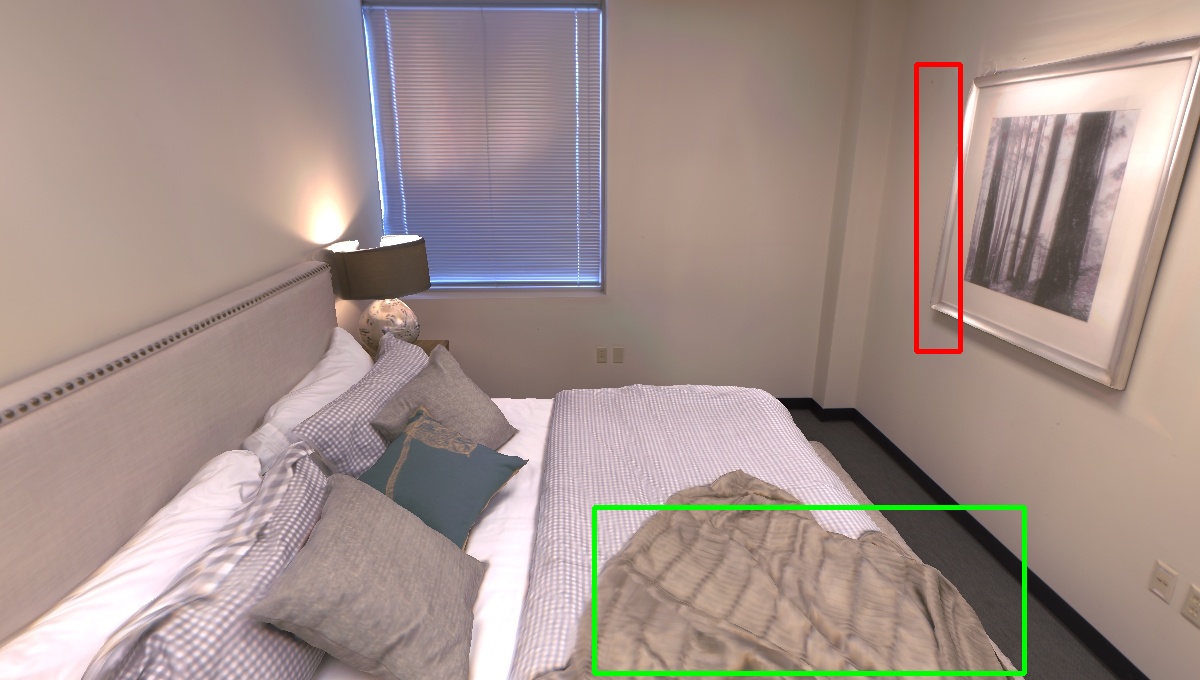} \\
		\hline
		\includegraphics[width=\replicaImSize\linewidth]{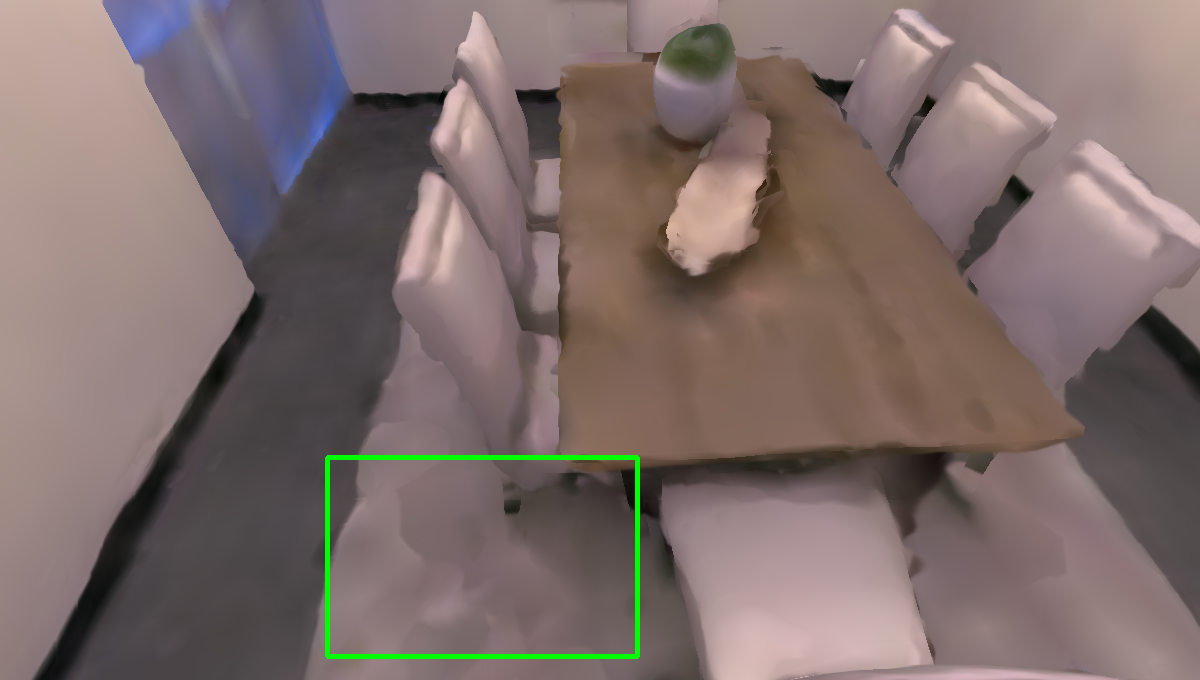} &
		\includegraphics[width=\replicaImSize\linewidth]{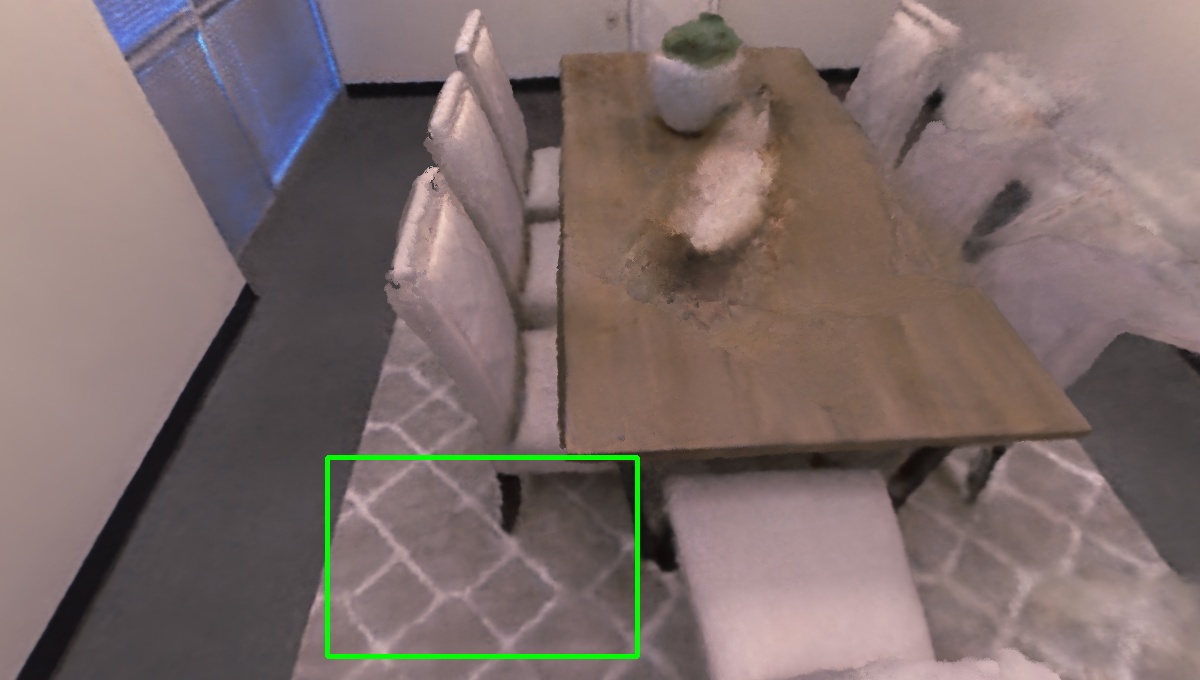} &
		\includegraphics[width=\replicaImSize\linewidth]{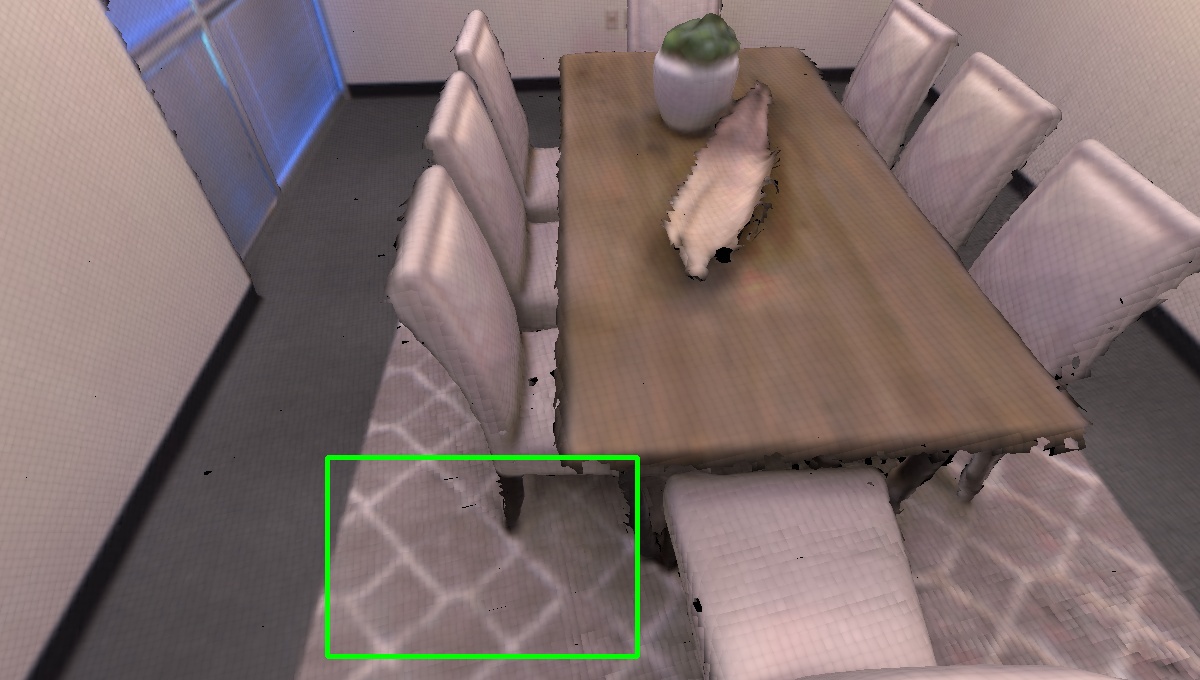} &
		\includegraphics[width=\replicaImSize\linewidth]{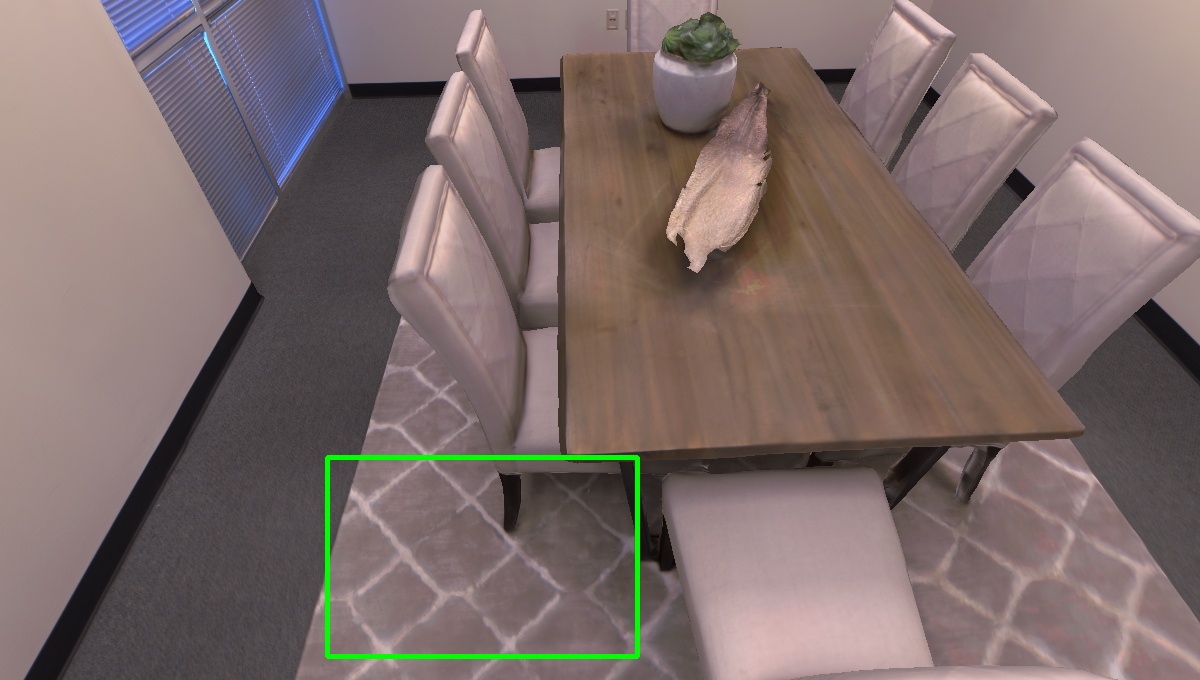} \\		
		
	\end{tabular}
	\caption{Demonstration of color rendering on the Replica dataset. Fine appearances are highlighted in {\color{green}green window}. Small defects are in a {\color{red}red} box.}
	\label{fig:replica_render}
	\vspace{-.5cm}
\end{figure*}
In this evaluation, we compare with implicit reconstruction (TSDF-Fusion, $\sigma$-Fusion) and latent implicit reconstruction models (iMAP, NICE-SLAM) that support color. 
Additionally,  we include a large-scale NeRF model, NeRF-SLAM, in the comparison.  
It is important to note that NeRF is SOTA in the view-synthesis task, which gives it an unfair advantage over other models because it learns light directions and does not really model a surface.
However, we include NeRF in this evaluation to demonstrate  that Uni-Fusion significantly reduces the gap.
Notably, NeRF-SLAM embeds external tracking model~\cite{teed2021droid,rosinol2023probabilistic} to provide poses while using SOTA NeRF implementation Instant-ngp~\cite{muller2022instant} for NeRF construction.

Uni-Fusion tracks and follows the same setting as in ScanNet test to take every 10 frames for mapping.
NICE-SLAM and NeRF-SLAM create depth and color using volumetric rendering.
In Uni-Fusion, we cast rays from the virtual camera onto our result surface mesh for the depth image. 
The cast points are then inferred using Uni-Fusion's color LIM to obtain color results.

According to~\cref{tab:replica_per_scene}, Uni-Fusion demonstrate
the best Depth L1 on all scenes with an average of \textbf{$\pmb{1.47}\si{\centi\meter}$ depth L1}. 
This is a \textbf{$\pmb{177\%}$ boost} compared to the second best model.

Moreover, excluding NeRF, our Uni-Fusion also shows the best performance in modeling the colors, achieving an average PSNR of $28.07$$\si{\dB}$.

However, it is strange that NICE-SLAM loses details while in two cases, it shows better PSNR than Uni-Fusion. 
To highlight the true result,
we provide rendered images in~\cref{fig:replica_render}.
It is evident that our Uni-Fusion accurately models the details of painting, carpet and quilt, while NICE-SLAM only roughly models the average color.

In addition, from~\cref{fig:replica_render}, Uni-Fusion's rendering quality is as precise as NeRF. 
The painting, carpet and quilt in Uni-Fusion's results are very similiar to the original appearance.
The {\color{green} green window} highlights the regions of interest.
Uni-Fusion reproduces the high-quality appearances that are very close to NeRF in terms of qualitative evaluation.
However, Uni-Fusion still has a quantitative score gap to NeRF's color rendering ($41.4\si{\dB}$), despite the highly comparable qualitative results to NeRF and ground truth.
We attribute this difference to three main factors: 
\textbf{1.} Uni-Fusion does not model the light directions to points, which is essential to NeRF.
\textbf{2.} NeRF optimizes image quality by focusing primarily on color rather than depth,
which is evident from its higher color rendering score but much worse depth rendering compared to Uni-Fusion.
\textbf{3.} Uni-Fusion does not support hole filling,
which results in black holes in the rendered images.

We summarize the differences between Uni-Fusion and other models in~\cref{tab:replica_diff}.
Similar to TSDF-Fusion and $\sigma$-Fusion, Uni-Fusion is a forward method that does not require any training of map representation, i.e., pre- or on-line training. 
It shares similarities with NICE-SLAM and NeRF-SLAM in producing an implicit map with a set of latents, that outputs results at arbitrary resolution.
However, Uni-Fusion differs in the extraction of the signed distance field,
as each query value is directly inferred using the corresponding ruling latent,
while NICE-SLAM and NeRF-SLAM use a much denser grid of features for interpolation during volumetric rendering based inference.

Like TSDF-Fusion and $\sigma$-Fusion, our Uni-Fusion is a real-time algorithm,
whereas iMAP, NICE-SLAM and NeRF-SLAM are not capable of running in real time.
NeRF-SLAM claims to be real-time, which is questionable as it still needs hundreds of epochs training after feeding the data.

However, optimization with backpropagation allows for pixel-to-pixel learning,
which is theoretically superior to the regression and fusion strategy. 
Although Uni-Fusion demonstrates its high ability to model colors, exploring NeRF-like post-optimization would be a promising direction for further improvements.

\begin{figure}[]
	\centering
	\includegraphics[width=.49\linewidth]{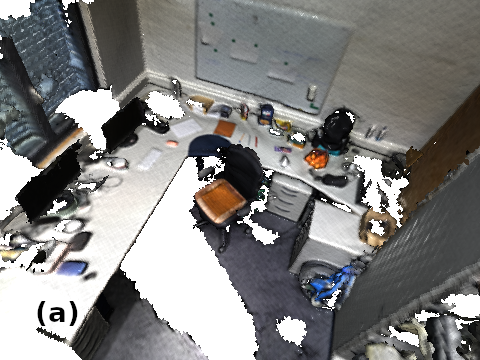}
	\includegraphics[width=.49\linewidth]{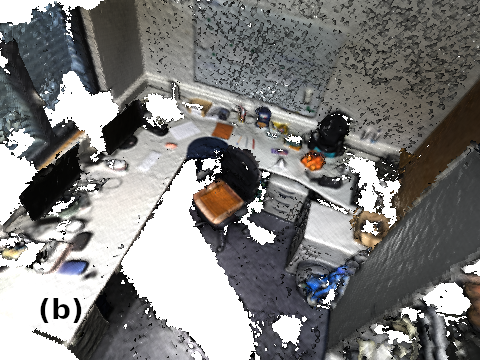}
	\caption{Ablation study on surface construction basis. (a) Sample based. (b) Derivative based.}
	\label{fig:ablation:GPIS}
\end{figure}


\begin{figure}
	\centering
	\psfragfig[width=.8\linewidth]{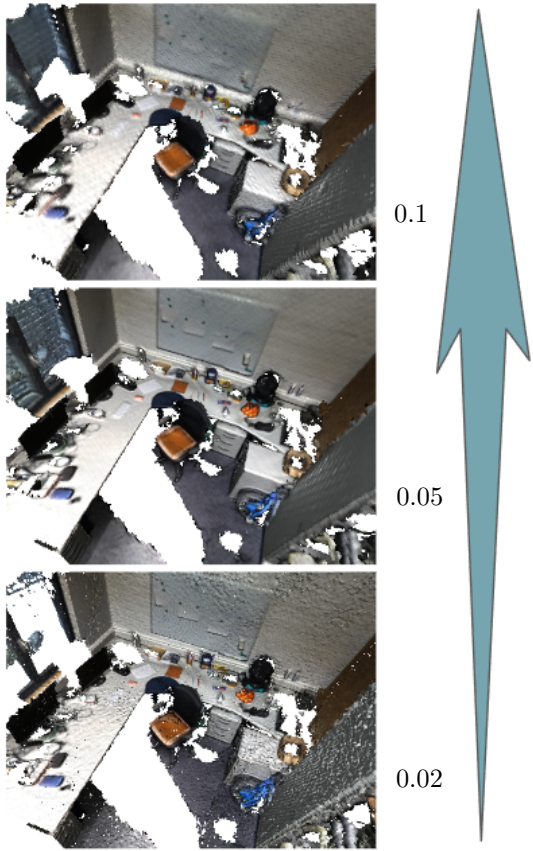}{
	\psfrag{a}{$ {0.1}$}
	\psfrag{b}{$ {0.05}$}
	\psfrag{c}{$ {0.02}$}
	}
	\caption{Ablation study on voxel size.}
	\label{fig:ablation:voxel_size}
\end{figure}
\begin{figure}
	\centering
	\includegraphics[width=1\linewidth]{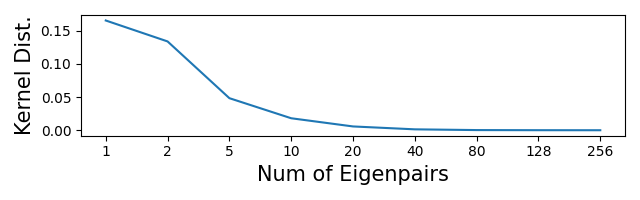}
	\includegraphics[width=1\linewidth]{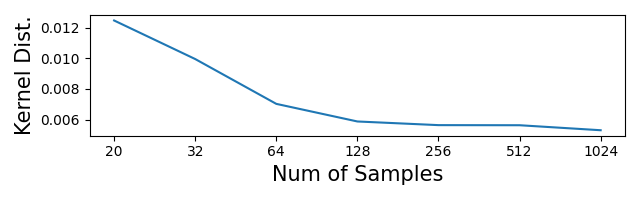}
	\caption{Ablation study on number of eigenpairs and number of anchor samples in kernel approximation.}
	\label{fig:ablation:approx}
\end{figure}

\newcommand{\styleImSize}{.2}
\begin{figure*}[]
	\centering
	\setlength{\tabcolsep}{0.01em}
	\renewcommand{\arraystretch}{.1}
	\resizebox{.95\textwidth}{!}{\begin{tabular}{ccccc}
			\toprule
			\includegraphics[width=\styleImSize\linewidth]{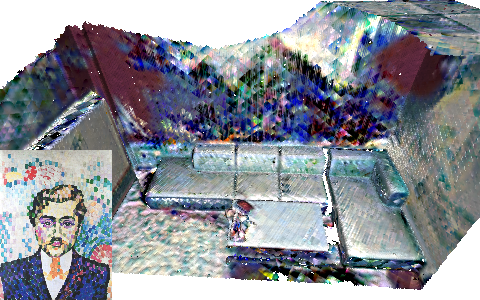} &
			\includegraphics[width=\styleImSize\linewidth]{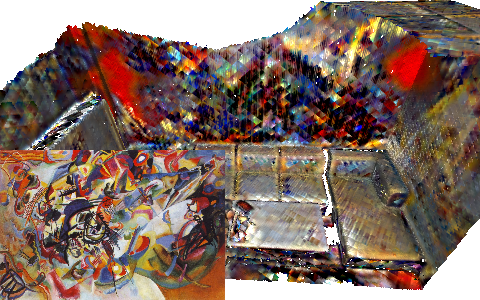} &
			\includegraphics[width=\styleImSize\linewidth]{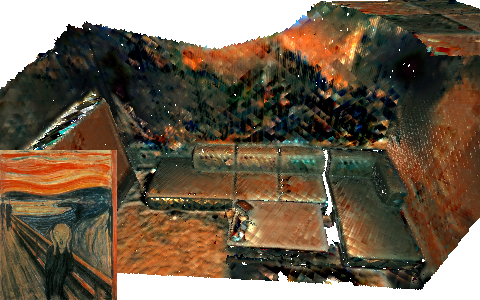} &
			\includegraphics[width=\styleImSize\linewidth]{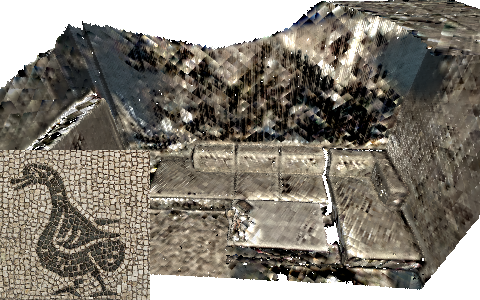} &
			\includegraphics[width=\styleImSize\linewidth]{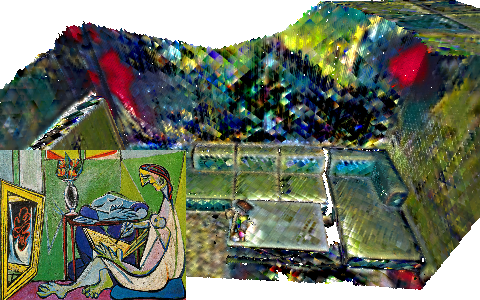} \\
			\includegraphics[width=\styleImSize\linewidth]{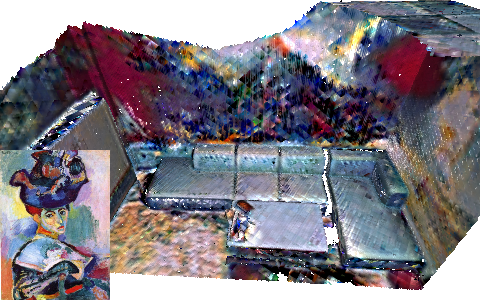} &
			\includegraphics[width=\styleImSize\linewidth]{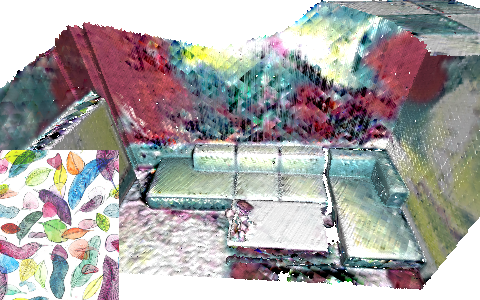} &
			\includegraphics[width=\styleImSize\linewidth]{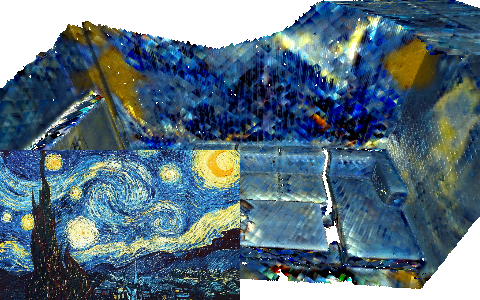} &
			\includegraphics[width=\styleImSize\linewidth]{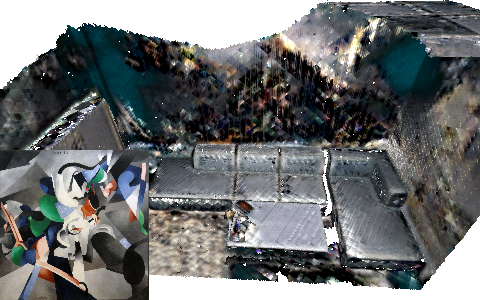} &
			\includegraphics[width=\styleImSize\linewidth]{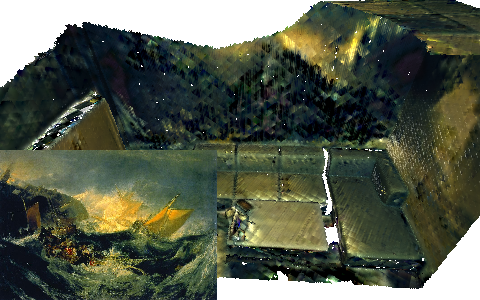} \\
			\includegraphics[width=\styleImSize\linewidth]{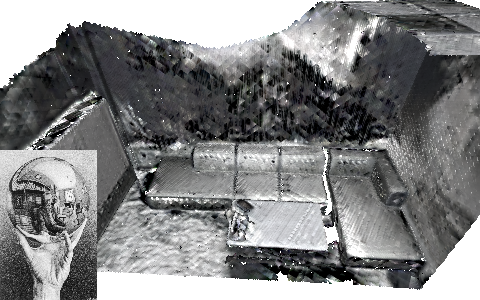} &
			\includegraphics[width=\styleImSize\linewidth]{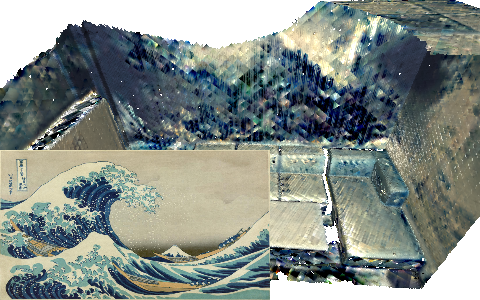} &
			\includegraphics[width=\styleImSize\linewidth]{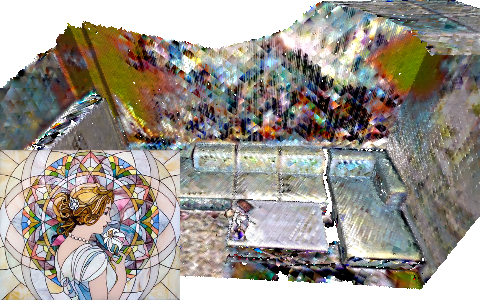} &
			\includegraphics[width=\styleImSize\linewidth]{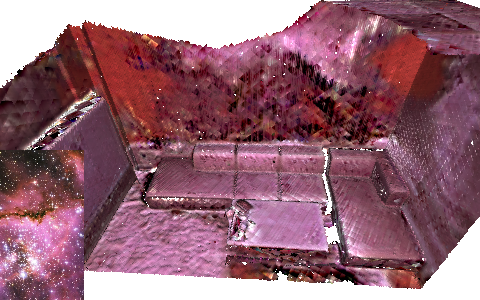} &
			\includegraphics[width=\styleImSize\linewidth]{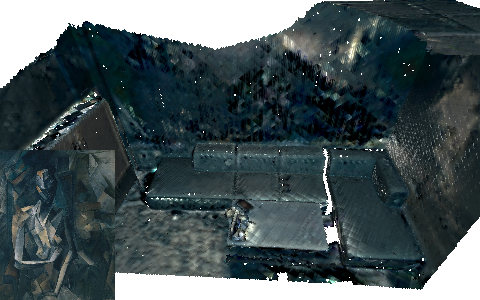} \\
			\includegraphics[width=\styleImSize\linewidth]{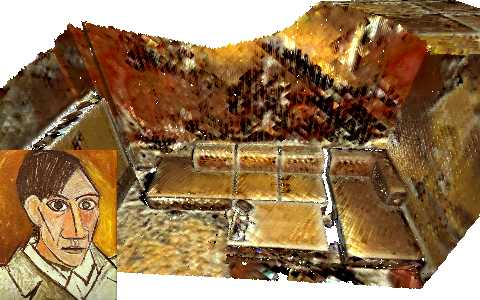} &
			\includegraphics[width=\styleImSize\linewidth]{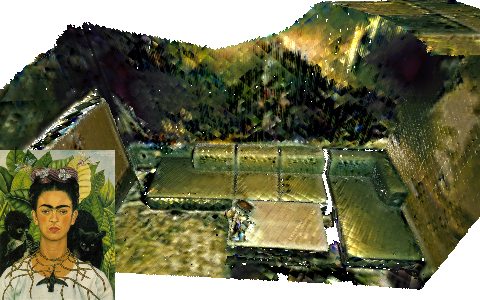} &
			\includegraphics[width=\styleImSize\linewidth]{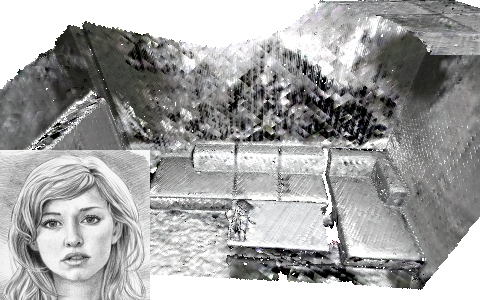} &
			\includegraphics[width=\styleImSize\linewidth]{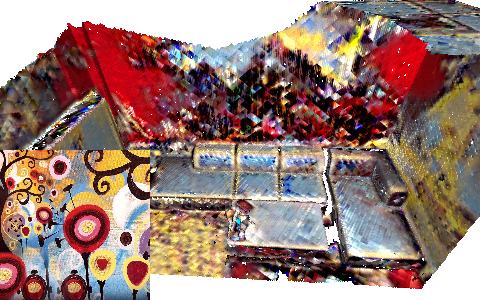} &
			\includegraphics[width=\styleImSize\linewidth]{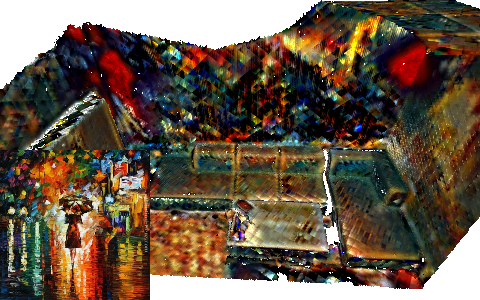} 
			\\	\hline
		\end{tabular}
	}
	\caption{Style transfer on 3D canvas.}
	\label{fig:style}
\end{figure*}
\subsection{Ablation study}
\label{exp:surface:ablation}

\subsubsection{Sample-based or Derivative-based}

We select the surface model with our own recorded sequences. 
All settings are detailed in \cref{sec:exp:details}.
As shown in~\cref{fig:ablation:GPIS}, reconstructions of Yijun's office are demonstrated. 
While both models are capable of construction, the derivative-based model introduces a lot of noise to the surface.
This issue arises, because, for smoothness purpose, we follow Di-Fusion~\cite{huang2021di} to build voxels that overlap with their neighbors, leading to redundant voxels near the surface.
For these redundant voxels, no center sample is provided and thus the derivative-based surface construction builds poor SDFs on unknown region of the voxels.

In contrast, the sample-based surface construction does not encounter this problem because it adds more points within the voxels, enabling the construction of very smooth surfaces.
We observe well-constructed and accurately colored objects such as the whiteboard, the chair, the school bag and even the oranges.

%
%
%

\subsubsection{Voxel size}

While testing of the office scene, we vary the voxel size from low to high. 
From~\cref{fig:ablation:voxel_size}, when a low voxel size $0.02\si{\meter}$ is used, the surface appears rough.
As the voxel size increases, the smoothness improves.
However, when a voxel size of $0.1\si{\meter}$ is employed, the surface appears blurry. 
Considering Uni-Fusion produces a surface color field, the quality of surface directly impacts the coloring.
Thus, further increasing the voxel size will result in deteriorated surface quality.

Therefore, in the above given experiments, $0.05\si{\meter}$ voxel size is used for surface construction.
Additionally, it should be clarified that each voxel used for encoding actually has a size of $0.1\si{\meter}$ due to the employment of overlapped voxels.

\subsubsection{Number of eigenpairs and anchor points}


The kernel approximation is affected by the number of eigenpairs and anchor points. 
Uni-Fusion treats the approximation module as a cohesive entity, expecting it to behave like a real kernel.
Therefore, we conduct the ablation study at the module level.

This feature dimension $l$ corresponds to the number of eigenpairs retained during kernel approximation. 
We employ a uniform sampling of $256$ anchor points for the approximation. 
To access the accuracy in recovering the original kernel, we randomly sample $2000$ test samples in $[-.5,.5]^3$ and compute the matrix $\V K$ using original Mat\'ern Kernel. Our approximation, denoted as $\hat{\V K}$, is then evaluated. 
We calculate the Mean Absolute Error (MAE) between the two kernels and
observe the curve presented in~\cref{fig:ablation:approx}.
From the figure, we find that the error decreases fast until $l$ reaches $20$, and beyond that point, the improvement becomes marginal. 
Considering that the result for $l=20$ is very close to that at $l=40$ while requiring only half the storage space, the optimal selection for $l$ is $20$.

Similarly, we perform a module-level ablation for number of anchor points.
With $l$ fixed at $20$, a minimum of $20$ samples is required.
\cref{fig:ablation:approx} demonstrates that the approximation shows minimal improvement beyond $256$ samples.
Additionally, since this number of anchor samples only affects the computation time and not the storage space,
it is advisable to select a the large value such as $256$, but not the largest, to improve efficiency without sacrificing accuracy.

\newcommand{\fabImSize}{.18}
\begin{figure*}[]
	\centering
	\setlength{\tabcolsep}{0.1em}
	\renewcommand{\arraystretch}{.1}
	\begin{tabular}{|c | c |c  | c | c | c |}
		\toprule 
		& \textbf{scene0568\_00} & \textbf{scene0249\_00} & \textbf{scene0435\_00} & \textbf{office3} & \textbf{room0}\\
		\midrule
		\rotatebox{90}{\textbf{Color}} &
		\raisebox{-.5\height}{\includegraphics[width=\fabImSize\linewidth]{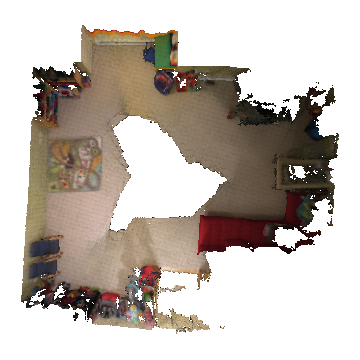}} & 
		\raisebox{-.5\height}{\includegraphics[width=\fabImSize\linewidth]{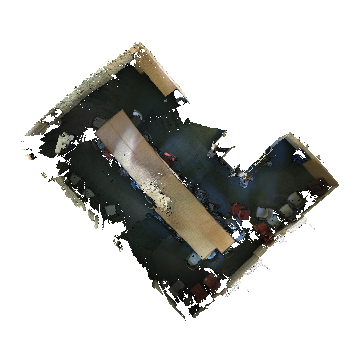}} & \raisebox{-.5\height}{\includegraphics[width=\fabImSize\linewidth]{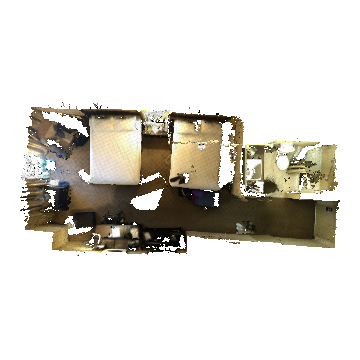}}
		& 
		\raisebox{-.5\height}{\includegraphics[width=\fabImSize\linewidth]{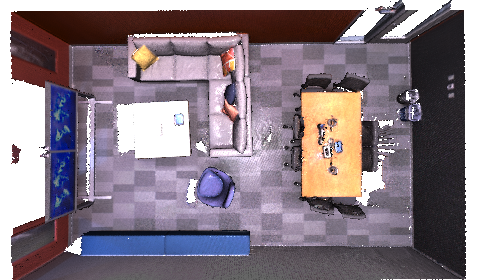}}&
		\raisebox{-.5\height}{\includegraphics[width=\fabImSize\linewidth]{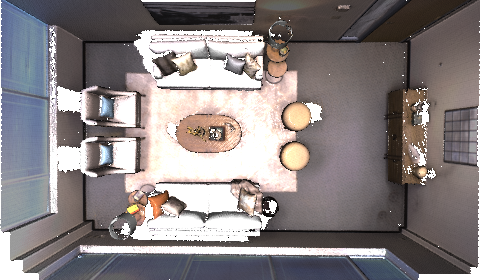}}
		\\	
		\midrule
		\rotatebox{90}{\textbf{Saliency}} &
		\raisebox{-.5\height}{\includegraphics[width=\fabImSize\linewidth]{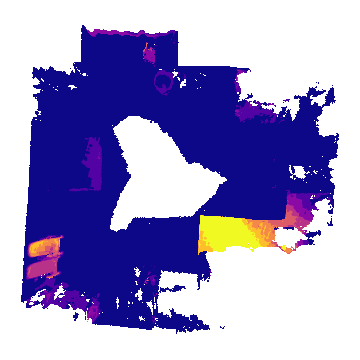}}&
		\raisebox{-.5\height}{\includegraphics[width=\fabImSize\linewidth]{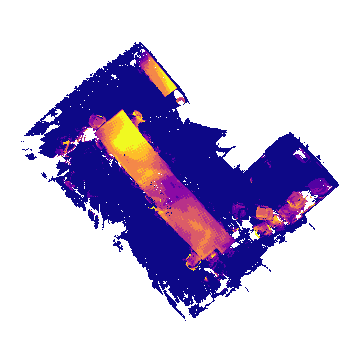}}&
		\raisebox{-.5\height}{\includegraphics[width=\fabImSize\linewidth]{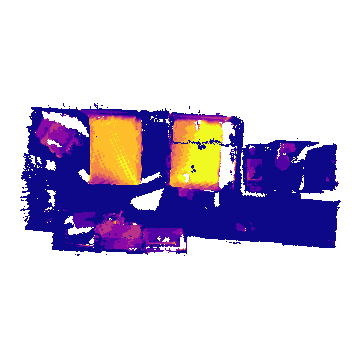}}&
		\raisebox{-.5\height}{\includegraphics[width=\fabImSize\linewidth]{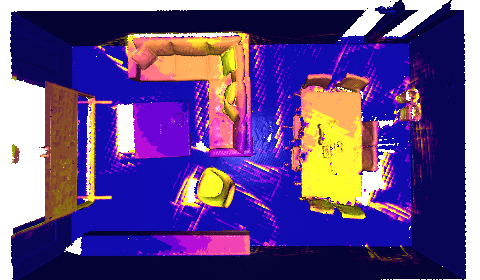}}&
		\raisebox{-.5\height}{\includegraphics[width=\fabImSize\linewidth]{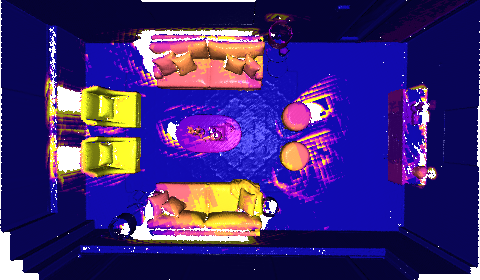}}
		\\
		
		\hline
	\end{tabular}
	\caption{Saliency transfer on the 3D canvas.
		The top row shows the result colored meshes.
		The bottom row shows the saliency meshes.}
	\label{fig:saliency}
\end{figure*}

\subsection{Results for 2D-to-3D transfer}

In addition to surface and color, Uni-Fusion also allows continuous mapping of other fabricated data.
Therefore, we apply Uni-Fusion to style data and saliency data in order to achieve style and saliency transfer on the 3D canvas.

In~\cref{fig:style}, we present artistic painting on the 3D canvas. 
Each frame undergoes style transfer using MSG-Net~\cite{zhang2018multi}, and Uni-Fusion is used to construct the style LIM for the surface style field.
The test scene is ``office0'' from the Replica dataset captured from a single custom view.
$20$ images are used to provide the style and are attached at lower left corner accordingly.
Our Uni-Fusion successfully constructs style meshes that closely resemble the taste of the supplied style images.
For instance, with pure style or abstract painting, the 3D ``canvas'' shows a very similar style. 
Among the style images, with our favorite style, located in the middle of the fourth row, Uni-Fusion produces a high quality 3D sketch painting. 

In~\cref{fig:saliency}, we demonstrate saliency detection in 3D.
In this figure we select three scenes from ScanNet and two scenes from Replica.
Similarly, we detect saliency on each frame using InSPyReNet\cite{kim2022revisiting} and construct saliency LIM for surface saliency field.

In the second row, high saliency regions are colored in yellow, indicating the object of interest. 
This information can be used to guide a robot's navigation in 3D scene.
For example, in the first column, the sofa, chair, curtain, and television in the room certainly attract more attention in daily life.
In the second column, a meeting room, the long desk and chairs are obviously the main components.
Similarly, in the third column showcasing a hotel room, the bed stands out, along with the sofa, desk, lamp, and TV in the fourth column, and the sofas and chairs in the last column.

\begin{table*}[b!]
	\caption{GZSL semantic segmentation results. Scores are in \%.
		$^\dagger$ indicate 3DGenZ's adaption of the method.
		Note that, Uni-Fusion-SU does not even train with the seen classes.}
	\centering
	\begin{tabular}{l|c|c |c ||ccc|ccc}
		\toprule
		\multicolumn{1}{c}{}& \multicolumn{2}{c|}{Training set} & Inference input &\multicolumn{3}{c|}{ScanNet } & \multicolumn{3}{c}{S3DIS}\\
		& Backbone & Classifier & &$Seen$& $Unseen$ & $All$&$Seen$& $Unseen$ & $All$
		\\
		\midrule
		
		\multicolumn{5}{l}{\textit{Supervised methods with different levels of supervision}}\\
		
		Full supervision & $seen \cup unseen$ & $seen \cup unseen$ & Point Cloud &43.3&51.9 &45.1&74.0&50.0&66.6 \\
		
		ZSL backbone & $seen$ & $seen \cup unseen$  &Point Cloud&41.5&39.2 & 40.3&60.9& 21.5&  48.7 \\
		
		ZSL-trivial & $seen$ & $seen$ &Point Cloud&39.2&0.0&31.3&70.2 &0.0&48.6  \\
		\midrule
		\multicolumn{5}{l}{\textit{Generalized zero-shot-learning methods}}\\
		
		ZSLPC-Seg~\cite{cheraghian2019zero}$^\dagger$ & $seen$ & $unseen$  &Point Cloud&28.2&0.0& 22.6&65.6 &0.0& 45.3\\
		
		DeViSe-3DSeg~\cite{frome2013devise}$^\dagger$ & $seen$ & $unseen$   &Point Cloud &20.0&0.0&16.0&70.2&0.0& 48.6\\ 
		3DGenZ~\cite{michele2021generative} & $seen$ & $seen \cup \hat{unseen}$  &Point Cloud &32.8&7.7& {27.8}&53.1&7.3&   \textbf{39.0} \\
		\midrule
		\multicolumn{5}{l}{\textit{Zero-shot learning + map fusion}}\\
		Uni-Fusion-SU (Ours) &None&None&Sparse Frames&31.0&\textbf{41.9}&\textbf{32.9} &31.3&\textbf{24.0}&29.0\\
		\bottomrule
		\multicolumn{1}{l}{}\\[-7pt]
	\end{tabular}
	
	\label{tab:sem_seg_overview}
\end{table*}

\begin{figure*}[t!]
	\centering
	\setlength{\tabcolsep}{0.1em}
	\renewcommand{\arraystretch}{.1}
	\begin{tabular}{|c | c |c |||c |c | c|}
		\toprule
		{\Large{3DGenZ}} & {\Large{Uni-Fusion}} &{\Large{Ground Truth}} & {\Large{3DGenZ}} &{\Large{Uni-Fusion-SU}} & {\Large{Ground Truth}} \\ \midrule
		
		\includegraphics[width=\scannetImSize\linewidth]{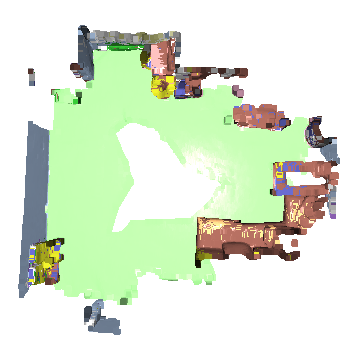}
		&\includegraphics[width=\scannetImSize\linewidth]{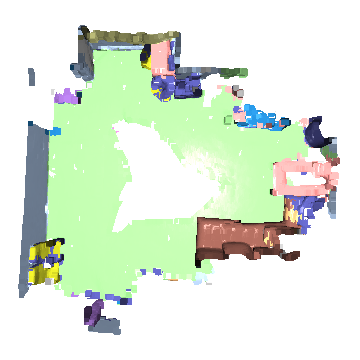}
		&\includegraphics[width=\scannetImSize\linewidth]{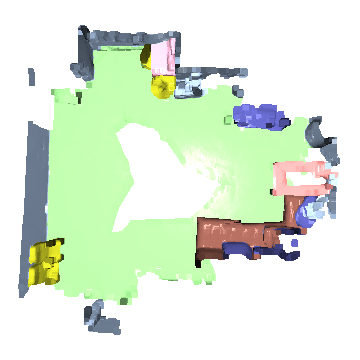}
		&		\includegraphics[width=\scannetImSize\linewidth]{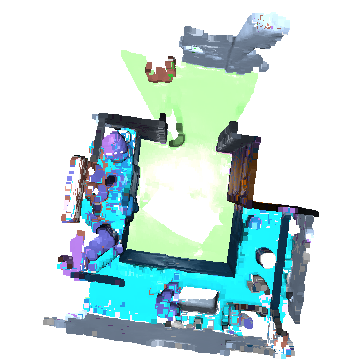}
		&\includegraphics[width=\scannetImSize\linewidth]{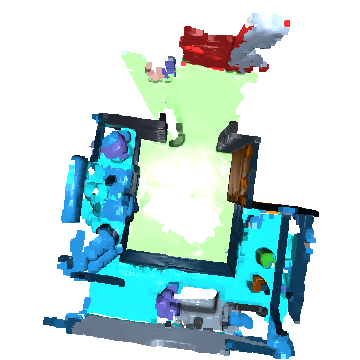}
		&\includegraphics[width=\scannetImSize\linewidth]{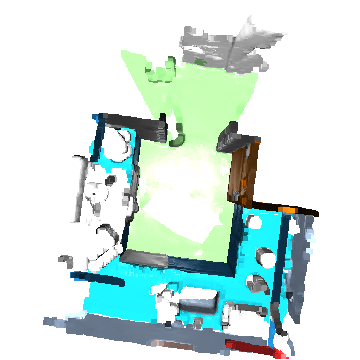}\\

		\includegraphics[width=\scannetImSize\linewidth]{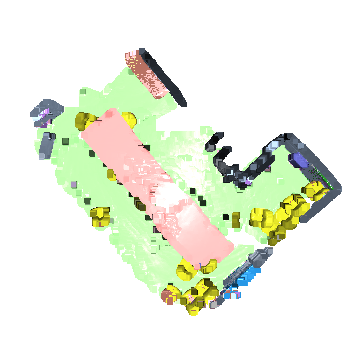}
		&\includegraphics[width=\scannetImSize\linewidth]{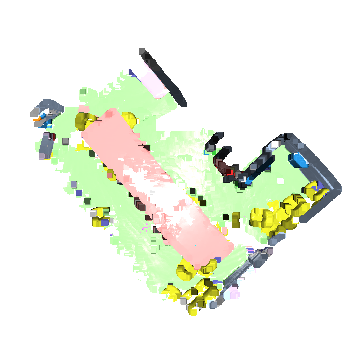}
		&\includegraphics[width=\scannetImSize\linewidth]{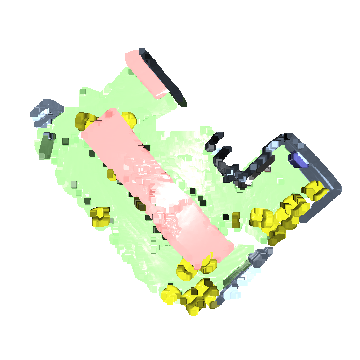}
		&		\includegraphics[width=\scannetImSize\linewidth]{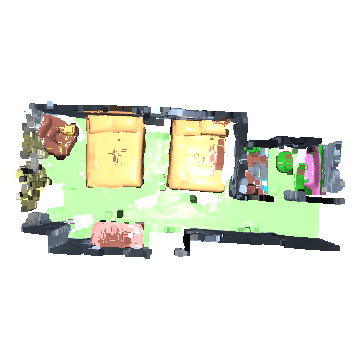}
		&\includegraphics[width=\scannetImSize\linewidth]{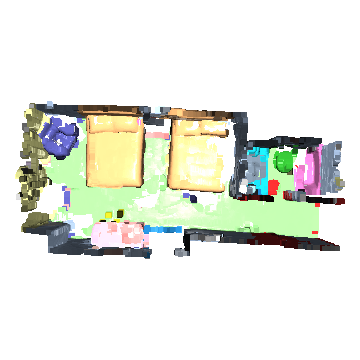}
		&\includegraphics[width=\scannetImSize\linewidth]{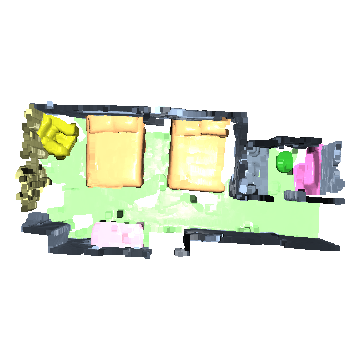}\\

		\includegraphics[width=\scannetImSize\linewidth]{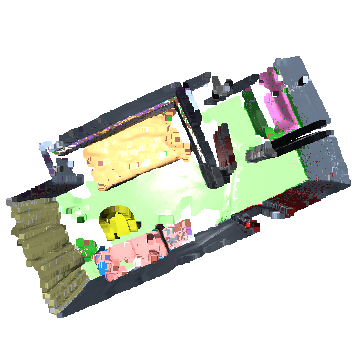}
		&\includegraphics[width=\scannetImSize\linewidth]{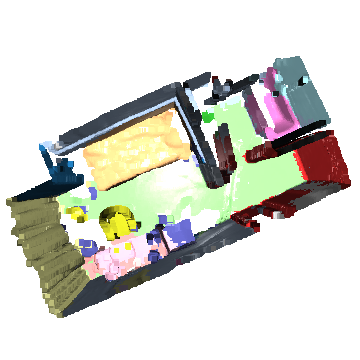}
		&\includegraphics[width=\scannetImSize\linewidth]{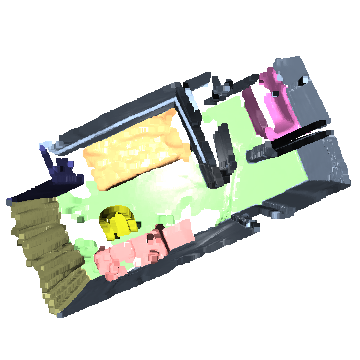}
		&		\includegraphics[width=\scannetImSize\linewidth]{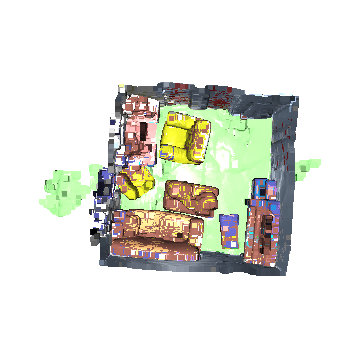}
		&\includegraphics[width=\scannetImSize\linewidth]{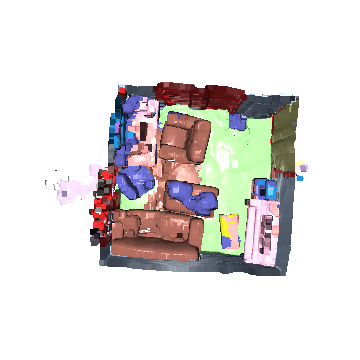}
		&\includegraphics[width=\scannetImSize\linewidth]{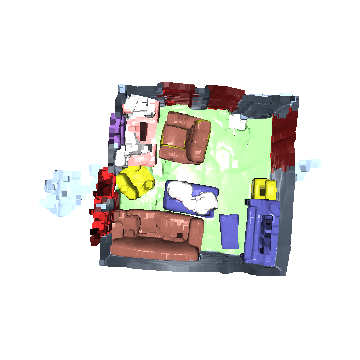}\\
		\bottomrule
		
	\end{tabular}
	\includegraphics[width=.95\linewidth]{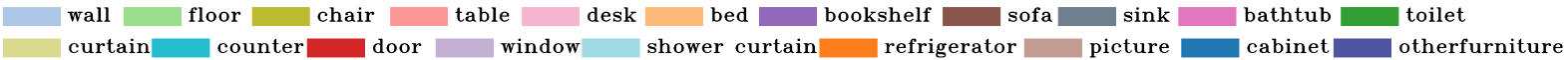}
	\caption{Demonstration of semantic segmentation on the ScanNet dataset.
		Selected scenes are consistent with~\cref{fig:recons:scannet_demo}}
	\label{fig:segmentation_demo}
	
\end{figure*}

\subsection{Scene Understanding Results}

Saliency detection effectively highlights the objects of interest.
This is also considered part of 3D semantic understanding.
However, as the semantics categories vary, fusing different categories of semantics into multiple LIMs can be inefficient.
Therefore, in this section, we utilize Uni-Fusion to fuse and construct a surface field for high-dimensional CLIP embeddings.
With a single LIM, we can generate different semantic results based on corresponding commands.
Since now our Uni-Fusion works with OpenSeg for scene understanding purposes, we call it Uni-Fusion-SU.

\subsubsection{Semantic Segmentation}
\label{sec:exp:semantic}

We first evaluate our model on generalized zero-shot point cloud semantic segmentation application.
Generalized Zero-Shot Learning (GZSL) differs from Zero-Shot Learning (ZSL) in that ZSL only predicts classes unseen during training, while GZSL predicts both unseen and seen classes~\cite{michele2021generative}.
Therefore, comparing our results with GZSL SOTAs provides a better understanding of the potential of Uni-Fusion-SU, as it does not train on both seen and unseen. 

This test uses ScanNet and S3DIS datasets for benchmarking. 
It is important to note that the \textbf{compared baselines are trained on the corresponding datasets}.
Our Uni-Fusion-SU uses OpenSeg to provide the 2D image level feature ebmedding.
Although \textbf{Uni-Fusion-SU} is also zero-shot, \textbf{it does not touch any ScanNet or S3DIS annotations}.

We demonstrate the mIoU scores in~\cref{tab:sem_seg_overview}.
In particular, our model achieves best results among the zero-shot learning methods on the ScanNet dataset and remains competitive with fully supervised methods.

Furthermore, we provide results specifically for the unseen classes in~\cref{sup:tab:sn_acc_miou}.
Although not as good as the fully supervised approach, Uni-Fusion-SU performs much better than 3DGenZ.
In addition, our Uni-Fusion-SU demonstrates high precision in classes such as sofa and Toilet, even when compared to the fully supervised model.

\begin{table}[htbp]
	\caption{Classwise GZSL semantic segmentation performance (\%) on the ScanNet unseen split.}
	\centering
	\newcommand*\rotext{\multicolumn{1}{R{45}{1em}}}
	\setlength{\tabcolsep}{1.7pt}
	\begin{tabular}{@{}l@{~}c|rrrr|r@{}}
		\toprule		
		& &
		{\textbf{Bookshelf}} & {\textbf{Desk}} & {\textbf{Sofa}} & {\textbf{Toilet}} & \stackbox{mean} \\
		
		\midrule
		FSL (Fully supervise) & IoU & 	56.9&	30.0&	57.4&	63.4 & 51.9
		\\ 
		3DGenZ (Zero-shot) & IoU & 	6.3&	3.3&	13.1&	8.1 & 7.7
		\\
		Uni-Fusion-SU (Ours) & IoU &38.3&16.8&51.7&60.9&41.9
		\\ \midrule 
		3DGenZ (Zero-shot)& Acc. & 	13.4&	5.9&	49.6&	26.3 &23.8
		\\
		Uni-Fusion-SU (Ours) & Acc. &61.9&29.6&67.4&91.6& 62.6
		\\
		\bottomrule
	\end{tabular}
	
	\label{sup:tab:sn_acc_miou}
\end{table}

However, in the S3DIS dataset, our model does not outperform 3DGenZ and other methods as shown in~\cref{tab:sem_seg_overview}.

Even in the result of unsceened data, as presented in \cref{sup:tab:s3dis_acc_miou}, we observe that Uni-Fusion-SU hardly finds some classed, e.g. Beam and Column, which are not commonly annotated objects. 
However, for common objects like sofa and window, our model performs much better.

\begin{table}[htbp]
	\caption{Classwise GZSL semantic segmentation performance (\%) on the S3DIS unseen split.}
	\centering
	\newcommand*\rotext{\multicolumn{1}{R{45}{1em}}}
	\setlength{\tabcolsep}{1.7pt}
	\begin{tabular}{@{}l@{~}c|rrrr|r@{}}
		\toprule		
		& &
		{\textbf{Beam}} & {\textbf{Column}} & {\textbf{Sofa}} & {\textbf{Window}} & \stackbox{mean} \\
		
		\midrule
		FSL (Fully supervise) & IoU & 	63.1&	10.2&	54.1&	72.4 & 50.0
		\\ 
		3DGenZ (Zero-shot) & IoU & 	13.9&	2.4&4.9&	8.1 &7.3
		\\
		Uni-Fusion-SU (Ours) & IoU &5.5&0.02&57.4&32.9&	24.0
		\\ \midrule 
		3DGenZ (Zero-shot) & Acc. & 	20.0&	9.1&	62.4&	23.7 &28.8
		\\
		Uni-Fusion-SU (Ours) & Acc. &41.5&0.02&78.3&42.1& 40.5
		\\	
		\bottomrule
	\end{tabular}
	
	\label{sup:tab:s3dis_acc_miou}
\end{table}

We present the results of the semantic segmentation in~\cref{fig:segmentation_demo}. 
It is evident that, 3DGenZ's result contains more noise, as seen in the spotted sofa, bed and other objects.
Conversely, Uni-Fusion-SU's results are generally smoother and more precise.

%
%
%
%

\subsubsection{Scene Understanding with Different Properties}

\begin{figure*}[t!]
	\centering
	\setlength{\tabcolsep}{0.1em}
	\renewcommand{\arraystretch}{.1}
	\resizebox{.98\textwidth}{!}{\begin{tabular}{|c | c | c | c | c | c|}
			\toprule 
			& \textbf{scene0568\_00} & \textbf{scene0249\_00} & \textbf{scene0435\_00} & \textbf{office3} & \textbf{room0}\\
			\midrule
			{} &
			\raisebox{-.5\height}{\includegraphics[width=\fabImSize\linewidth]{im/exp/fab/scannet/0568_color.png}} & 
			\raisebox{-.5\height}{\includegraphics[width=\fabImSize\linewidth]{im/exp/fab/scannet/0249_color.png}} & \raisebox{-.5\height}{\includegraphics[width=\fabImSize\linewidth]{im/exp/fab/scannet/0435_color.png}}
			&
			\raisebox{-.5\height}{\includegraphics[width=\fabImSize\linewidth]{im/exp/fab/replica/office3_color.png}}
			&
			\raisebox{-.5\height}{\includegraphics[width=\fabImSize\linewidth]{im/exp/fab/replica/room0_color.png}}\\ 
			\\
			\textbf{Desk}  &
			\raisebox{-.5\height}{\includegraphics[width=\fabImSize\linewidth]{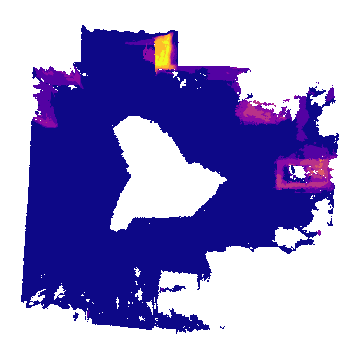}}&
			\raisebox{-.5\height}{\includegraphics[width=\fabImSize\linewidth]{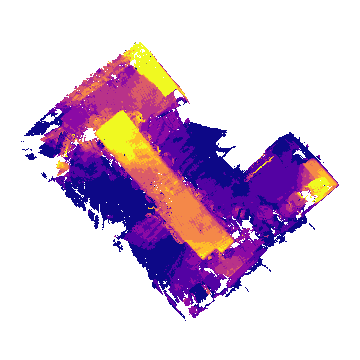}}&
			\raisebox{-.5\height}{\includegraphics[width=\fabImSize\linewidth]{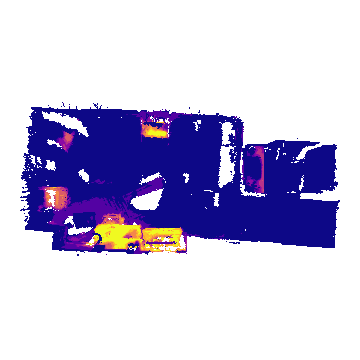}}
			&
			\raisebox{-.5\height}{\includegraphics[width=\fabImSize\linewidth]{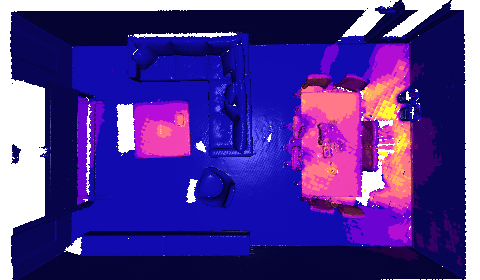}}
			&
			\raisebox{-.5\height}{\includegraphics[width=\fabImSize\linewidth]{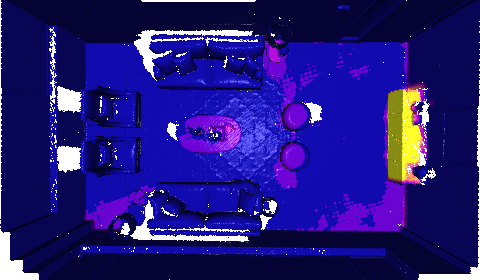}}\\
			\\
			
			\textbf{Sofa} &
			\raisebox{-.5\height}{\includegraphics[width=\fabImSize\linewidth]{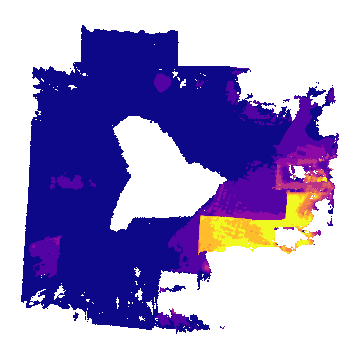}} &
			\raisebox{-.5\height}{\includegraphics[width=\fabImSize\linewidth]{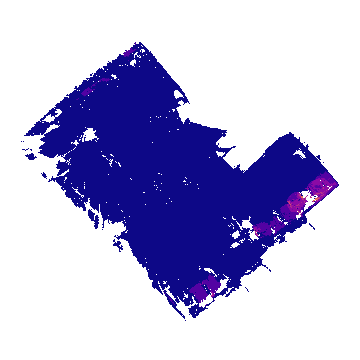}} &
			\raisebox{-.5\height}{\includegraphics[width=\fabImSize\linewidth]{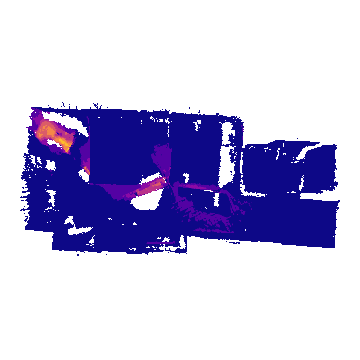}}&
			\raisebox{-.5\height}{\includegraphics[width=\fabImSize\linewidth]{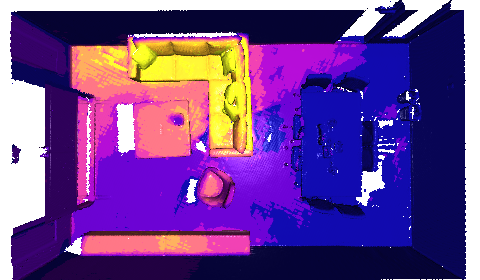}}
			&
			\raisebox{-.5\height}{\includegraphics[width=\fabImSize\linewidth]{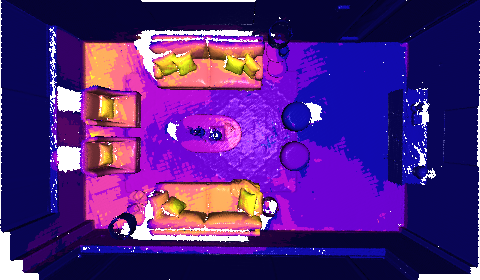}}\\
			\\
			\textbf{Work} &
			\raisebox{-.5\height}{\includegraphics[width=\fabImSize\linewidth]{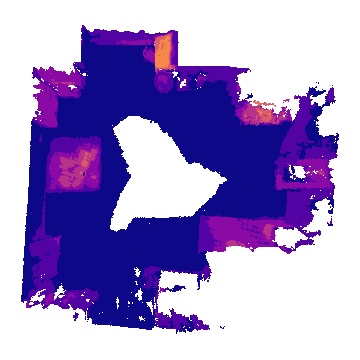}} &
			\raisebox{-.5\height}{\includegraphics[width=\fabImSize\linewidth]{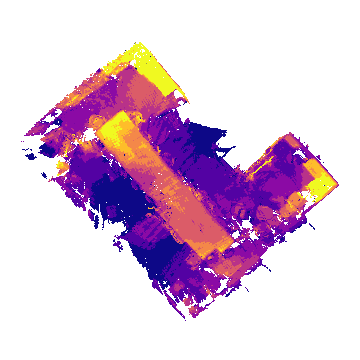}} &
			\raisebox{-.5\height}{\includegraphics[width=\fabImSize\linewidth]{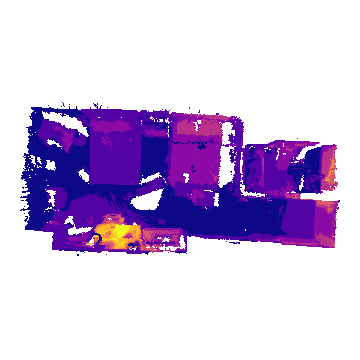}}&
			\raisebox{-.5\height}{\includegraphics[width=\fabImSize\linewidth]{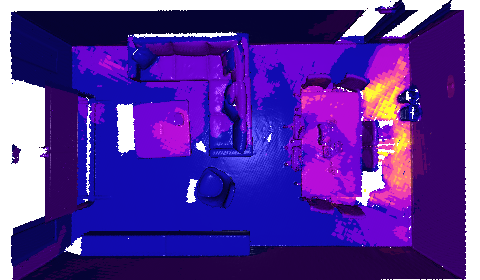}}
			&
			\raisebox{-.5\height}{\includegraphics[width=\fabImSize\linewidth]{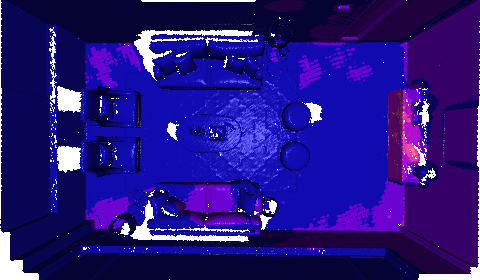}}\\
			\\
			\textbf{Sittable} &
			\raisebox{-.5\height}{\includegraphics[width=\fabImSize\linewidth]{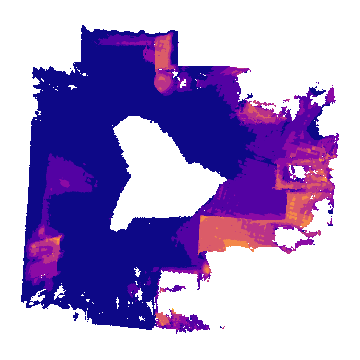}} &
			\raisebox{-.5\height}{\includegraphics[width=\fabImSize\linewidth]{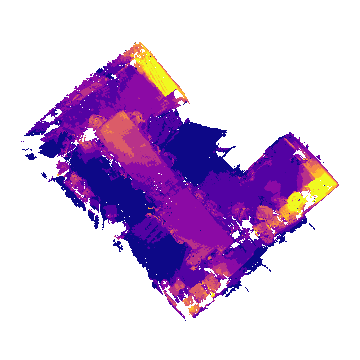}} &
			\raisebox{-.5\height}{\includegraphics[width=\fabImSize\linewidth]{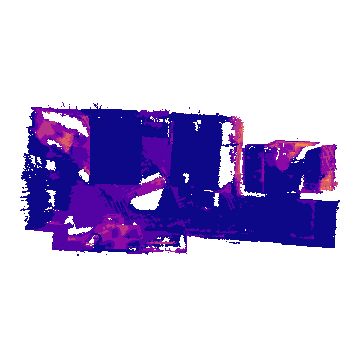}}&
			\raisebox{-.5\height}{\includegraphics[width=\fabImSize\linewidth]{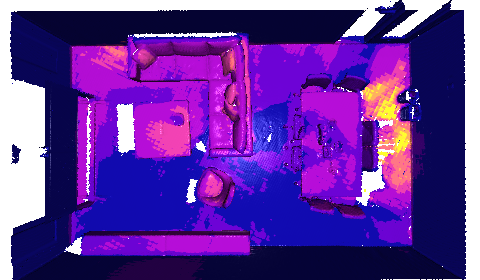}}
			&
			\raisebox{-.5\height}{\includegraphics[width=\fabImSize\linewidth]{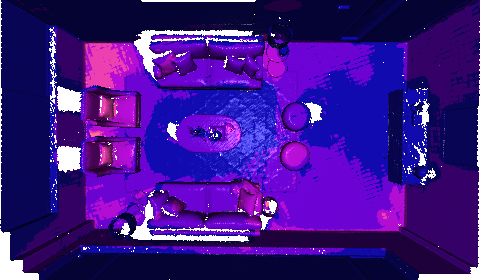}}\\
			\\
			\textbf{Wood} &
			\raisebox{-.5\height}{\includegraphics[width=\fabImSize\linewidth]{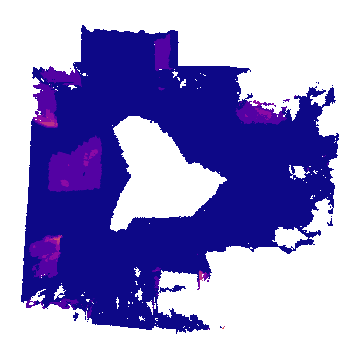}} &
			\raisebox{-.5\height}{\includegraphics[width=\fabImSize\linewidth]{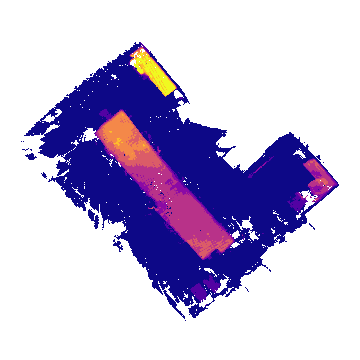}} &
			\raisebox{-.5\height}{\includegraphics[width=\fabImSize\linewidth]{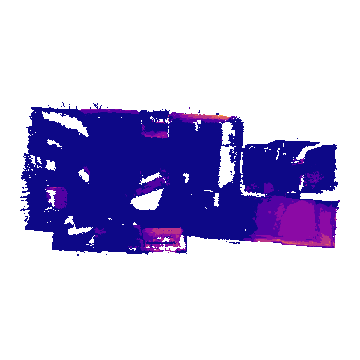}}&
			\raisebox{-.5\height}{\includegraphics[width=\fabImSize\linewidth]{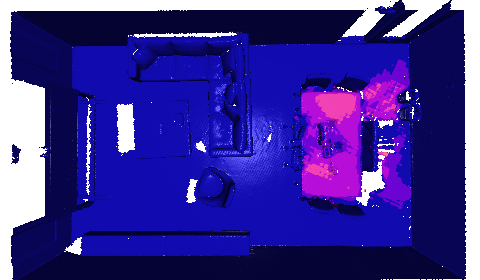}}
			&
			\raisebox{-.5\height}{\includegraphics[width=\fabImSize\linewidth]{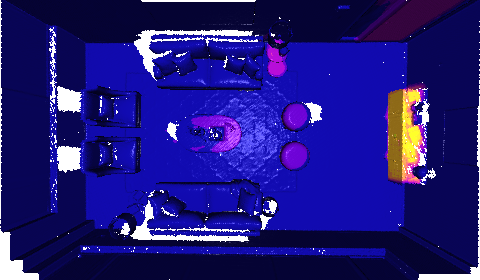}}\\
			\\

			\bottomrule
		\end{tabular}
	}
	\caption{Demonstration of the original mesh, highlighted semantic mesh given various queries.}
	\label{fig:fab_lt}
	\vspace{-.5cm}
\end{figure*}

The main contribution of this application is that, Uni-Fusion is the first model to construct a continuous mapping of high-dimensional embeddings onto the surface without the need for any training of the map representation.
In the previous experiment (\cref{sec:exp:semantic}), we evaluate the performance of generalized zero-shot semantic segmentation.
However, the potential of Uni-Fusion goes beyond semantic segmentation.
By constructing a LIM, we obtain a surface CLIP feature field.
This enables us to query various semantic categories such as 
\textbf{Object, Room Type, Material, Affordance and Activity} without requiring multiple LIMs or re-running the model.

We present the results in \cref{fig:fab_lt}, 
where we query object (desk, sofa), activity (work), affordance (sittable), and material (wood).
Uni-Fusion-SU accurately identifies and highlights the object and material regions.
However, for less specific commands such as work or sittable, the model provides a wider range of results with less confidence (indicated by dull yellow).
Nevertheless, the suggested options are also roughly correct.

\subsection{Time}

We run all of the applications in a single pass using our captured office sequences and evaluate the time cost of construction and fusion of each LIM. 
The average time cost across frames is shown in~\cref{tab:time}.

\begin{table}[htbp]
	\caption{Time required for each frame.
	}
	\centering
	\footnotesize
	\setlength{\tabcolsep}{0.7em}
	\resizebox{\linewidth}{!}{
		\begin{tabular}{l|ccccccc}
			\toprule
			&Surface & Color & Infrared & Style & Saliency & Latent&Internal Track \\ \midrule
			Time ($\si{\second}$)&0.100 & 0.038 & 0.045 & 0.048 & 0.045 &0.011 &0.225 \\ \bottomrule
	\end{tabular}}
	
	\label{tab:time}
\end{table}

\newcommand{\mineImSize}{.32}

\begin{figure*}[]
	\centering
	\setlength{\tabcolsep}{0.1em}
	\renewcommand{\arraystretch}{.1}
	\begin{tabular}{|c | c |c |}
		\hline \hline
		\includegraphics[width=\mineImSize\linewidth]{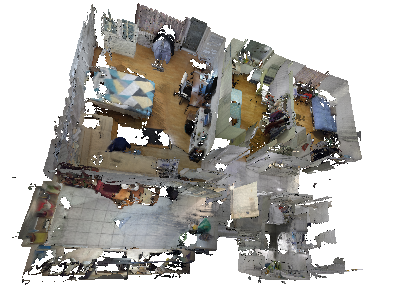}&	\includegraphics[width=\mineImSize\linewidth]{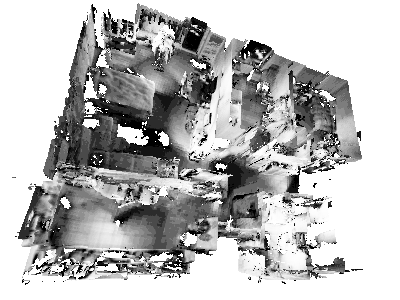}&	\includegraphics[width=\mineImSize\linewidth]{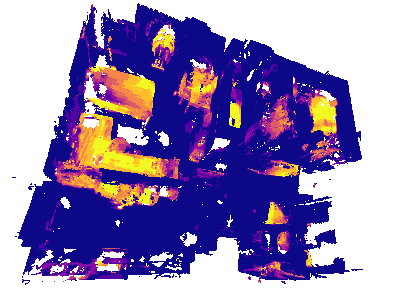}\\
		{Color} &{Infrared} & {Saliency}\\
		\includegraphics[width=\mineImSize\linewidth]{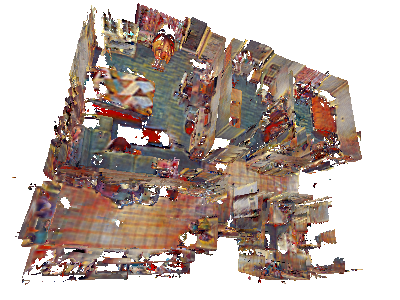}
		&\includegraphics[width=\mineImSize\linewidth]{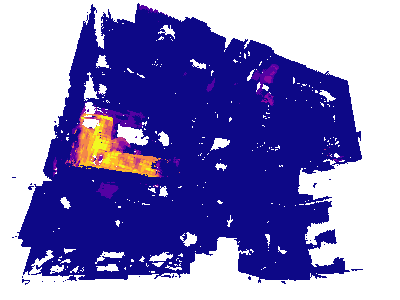}
		&\includegraphics[width=\mineImSize\linewidth]{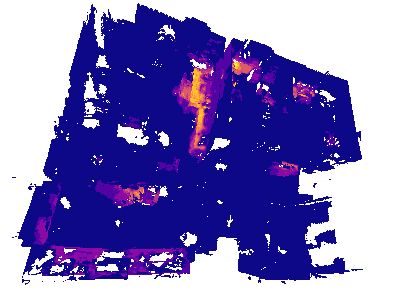}
		\\
		{Style} & {Object-sofa} & {Object-desk}\\
		\includegraphics[width=\mineImSize\linewidth]{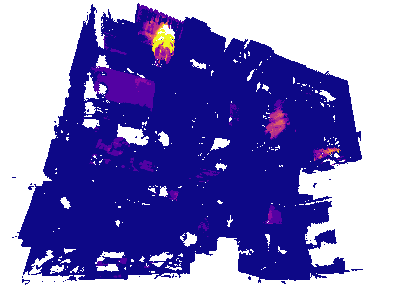}
		&\includegraphics[width=\mineImSize\linewidth]{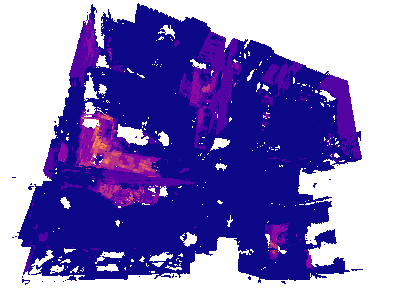}
		&\includegraphics[width=\mineImSize\linewidth]{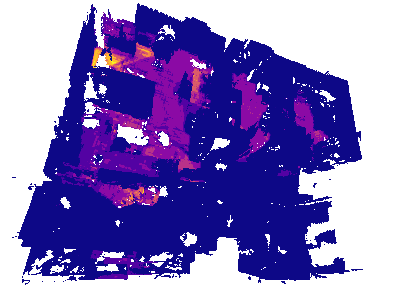}\\
		{Object-coat} & {Affordance-sit} & {Material-wood} \\\hline
	\end{tabular}
	\caption{Demonstration on the captured apartment data.}
	\label{fig:appartment}
\end{figure*}

Using depth and property images of size $720\times1280$ as input, it is evident from the table, that our model operates at a frequency of $\sim10\si{\hertz}$ for  surface (sample mode) LIM construction and integration. 
It alse achieves a frequency of over $20\si{\hertz}$ for color, infrared, style, and saliency.
These results demonstrate the suitability of Uni-Fusion for real-time applications.

However, our internal tracking process takes around $0.225\si{\second}$ per frame, which is relatively slower compared to the mapping module. 
Nevertheless, Uni-Fusion uses external tracking to prevent tracking loss, enabling our internal tracking and mapping to operate at a lower frequency.
As a result, the entire model can be effectively applied in real-time in various scenarios.

\section{Extensive experiment on our own dataset}

In previous experiments, we evaluate the capabilities of Uni-Fusion in different applications. 
To further demonstrate its effectiveness in robotic environmental understanding, we capture our own dataset to show all applications together.

We capture two scenes: The office and apartment of the first author using a Microsoft Kinect Azure. 
RGB-D and infrared video are captured. After calibration, RGB, depth, infrared inputs have resolution of $720\times1280$.
Uni-Fusion tracks and reconstructs all applications in one pass.
While office data has been involved in ablation study (\cref{exp:surface:ablation}), we showcase all applications using the apartment dataset, as depicted in~\cref{fig:appartment}.

For better visualization, the ceiling of reconstruction is removed.
The top row of images presents the colored mesh with room details, the infrared mesh revealing the lighting effect, and the saliency reconstruction highlighting objects crucial for navigation.
Additionally, we select the second style from~\cref{fig:style} for style transfer to the apartment canvas.
As a result, the wooden floor in the room is colored with dark green.
The whole apartment is in a warm style.

The remaining results are generated from the surface field of the CLIP embeddings. 
We issue commands to locate objects, e.g., where is the sofa, desk and coat.
In addition, it easily identifies affordances such as being sittable.
For material, it successfully detects the wooden floor in each room.

\section{Limitations and Future Work}

\subsubsection{Remapping}

Uni-Fusion currently lacks support for deintegrating local LIM from global LIM, which is essential for incorporating bundle adjustment or loop closing techniques. 
In addition, the current state of Uni-Fusion does not allow the transformation of LIMs as demonstrated by NIM-REM~\cite{yuan2022algorithm}. 
To enhance a better quality and to facilitate large scale mapping, loop closing and bundle adjustment are future targets.

\subsubsection{Visual Language Navigation}

Uni-Fusion serves as a solid foundation for reconstruction and scene understanding in the context of Visual-Language Robot Navigation (VLN).
While existing work produces a 2D embedding map~\cite{huang2023visual}, Uni-Fusion excels in constructing a 3D embedding map of the scene.
As a result, Uni-Fusion empowers the robot with a deeper understanding of the scene. In our future work, we intend to explore applications such as navigation.

\section{Conclusion}

In this paper, we have introduced Uni-Fusion, a novel universal model for all continuous mapping applications.
Without any training, Uni-Fusion constructs Latent Implicit Maps that support geometry and arbitrary properties. 
Moving one step further to scene understanding, Uni-Fusion is also the first model that is capable of constructing continuous maps with high-dimensional embeddings without the training of map representation.
With such a basis, we have implemented several applications, including a high-quality incremental surface and color reconstruction application, a 2D-to-3D transfer of fabricated properties, and an open-vocabulary scene understanding application.

	\bibliographystyle{IEEEtran}
	\bibliography{ref}

\begin{thebibliography}{10}
\providecommand{\url}[1]{#1}
\csname url@samestyle\endcsname
\providecommand{\newblock}{\relax}
\providecommand{\bibinfo}[2]{#2}
\providecommand{\BIBentrySTDinterwordspacing}{\spaceskip=0pt\relax}
\providecommand{\BIBentryALTinterwordstretchfactor}{4}
\providecommand{\BIBentryALTinterwordspacing}{\spaceskip=\fontdimen2\font plus
\BIBentryALTinterwordstretchfactor\fontdimen3\font minus
  \fontdimen4\font\relax}
\providecommand{\BIBforeignlanguage}[2]{{%
\expandafter\ifx\csname l@#1\endcsname\relax
\typeout{** WARNING: IEEEtran.bst: No hyphenation pattern has been}%
\typeout{** loaded for the language `#1'. Using the pattern for}%
\typeout{** the default language instead.}%
\else
\language=\csname l@#1\endcsname
\fi
#2}}
\providecommand{\BIBdecl}{\relax}
\BIBdecl

\bibitem{o2012gaussian}
S.~T. O’Callaghan and F.~T. Ramos, ``Gaussian process occupancy maps,''
  \emph{The Intl. Journal of Robotics Research}, vol.~31, no.~1, pp. 42--62,
  2012.

\bibitem{kim2013continuous}
S.~Kim and J.~Kim, ``Continuous occupancy maps using overlapping local gaussian
  processes,'' in \emph{2013 IEEE/RSJ international conference on intelligent
  robots and systems}.\hskip 1em plus 0.5em minus 0.4em\relax IEEE, 2013, pp.
  4709--4714.

\bibitem{ghaffari2018gaussian}
M.~Ghaffari~Jadidi, J.~Valls~Miro, and G.~Dissanayake, ``Gaussian processes
  autonomous mapping and exploration for range-sensing mobile robots,''
  \emph{Autonomous Robots}, vol.~42, pp. 273--290, 2018.

\bibitem{yuan2018fast}
Y.~Yuan, H.~Kuang, and S.~Schwertfeger, ``Fast gaussian process occupancy
  maps,'' in \emph{2018 15th Intl. Conf. on Control, Automation, Robotics and
  Vision (ICARCV)}.\hskip 1em plus 0.5em minus 0.4em\relax IEEE, 2018, pp.
  1502--1507.

\bibitem{martens2016geometric}
W.~Martens, Y.~Poffet, P.~R. Soria, R.~Fitch, and S.~Sukkarieh, ``Geometric
  priors for gaussian process implicit surfaces,'' \emph{IEEE Robotics and
  Automation Letters}, vol.~2, no.~2, pp. 373--380, 2016.

\bibitem{lee2019online}
B.~Lee, C.~Zhang, Z.~Huang, and D.~D. Lee, ``Online continuous mapping using
  gaussian process implicit surfaces,'' in \emph{2019 Intl. Conf. on Robotics
  and Automation (ICRA)}.\hskip 1em plus 0.5em minus 0.4em\relax IEEE, 2019,
  pp. 6884--6890.

\bibitem{wu2021faithful}
L.~Wu, K.~M.~B. Lee, L.~Liu, and T.~Vidal-Calleja, ``Faithful euclidean
  distance field from log-gaussian process implicit surfaces,'' \emph{IEEE
  Robotics and Automation Letters}, vol.~6, no.~2, pp. 2461--2468, 2021.

\bibitem{ivan2022online}
J.-P.~A. Ivan, T.~Stoyanov, and J.~A. Stork, ``Online distance field priors for
  gaussian process implicit surfaces,'' \emph{IEEE Robotics and Automation
  Letters}, vol.~7, no.~4, pp. 8996--9003, 2022.

\bibitem{curless1996volumetric}
B.~Curless and M.~Levoy, ``A volumetric method for building complex models from
  range images,'' in \emph{Proc. of the 23rd annual conf. on Computer graphics
  and interactive techniques}, 1996, pp. 303--312.

\bibitem{izadi2011kinectfusion}
S.~Izadi, D.~Kim, O.~Hilliges, D.~Molyneaux, R.~Newcombe, P.~Kohli, J.~Shotton,
  S.~Hodges, D.~Freeman, A.~Davison \emph{et~al.}, ``Kinectfusion: real-time 3d
  reconstruction and interaction using a moving depth camera,'' in \emph{Proc.
  of the 24th annual ACM symposium on User interface software and technology},
  2011, pp. 559--568.

\bibitem{dai2017bundlefusion}
A.~Dai, M.~Nie{\ss}ner, M.~Zollh{\"o}fer, S.~Izadi, and C.~Theobalt,
  ``Bundlefusion: Real-time globally consistent 3d reconstruction using
  on-the-fly surface reintegration,'' \emph{ACM Transactions on Graphics
  (ToG)}, vol.~36, no.~4, p.~1, 2017.

\bibitem{huang2021di}
J.~Huang, S.-S. Huang, H.~Song, and S.-M. Hu, ``Di-fusion: Online implicit 3d
  reconstruction with deep priors,'' in \emph{Proc. of the IEEE/CVF Conf. on
  Computer Vision and Pattern Recognition}, 2021, pp. 8932--8941.

\bibitem{yuan2022algorithm}
Y.~Yuan and A.~N{\"u}chter, ``An algorithm for the se (3)-transformation on
  neural implicit maps for remapping functions,'' \emph{IEEE Robotics and
  Automation Letters}, vol.~7, no.~3, pp. 7763--7770, 2022.

\bibitem{sucar2021imap}
E.~Sucar, S.~Liu, J.~Ortiz, and A.~J. Davison, ``imap: Implicit mapping and
  positioning in real-time,'' in \emph{Proc. of the IEEE/CVF Intl. Conf. on
  Computer Vision}, 2021, pp. 6229--6238.

\bibitem{zhu2022nice}
Z.~Zhu, S.~Peng, V.~Larsson, W.~Xu, H.~Bao, Z.~Cui, M.~R. Oswald, and
  M.~Pollefeys, ``Nice-slam: Neural implicit scalable encoding for slam,'' in
  \emph{Proc. of the IEEE/CVF Conf. on Computer Vision and Pattern
  Recognition}, 2022, pp. 12\,786--12\,796.

\bibitem{lorensen1987marching}
W.~E. Lorensen and H.~E. Cline, ``Marching cubes: A high resolution 3d surface
  construction algorithm,'' \emph{ACM siggraph computer graphics}, vol.~21,
  no.~4, pp. 163--169, 1987.

\bibitem{li2022bnv}
K.~Li, Y.~Tang, V.~A. Prisacariu, and P.~H. Torr, ``Bnv-fusion: Dense 3d
  reconstruction using bi-level neural volume fusion,'' in \emph{Proc. of the
  IEEE/CVF Conf. on Computer Vision and Pattern Recognition}, 2022, pp.
  6166--6175.

\bibitem{rosinol2022nerf}
A.~Rosinol, J.~J. Leonard, and L.~Carlone, ``Nerf-slam: Real-time dense
  monocular slam with neural radiance fields,'' \emph{arXiv preprint
  arXiv:2210.13641}, 2022.

\bibitem{ghiasi2022scaling}
G.~Ghiasi, X.~Gu, Y.~Cui, and T.-Y. Lin, ``Scaling open-vocabulary image
  segmentation with image-level labels,'' in \emph{Computer Vision--ECCV 2022:
  17th European Conf., Tel Aviv, Israel, October 23--27, 2022, Proc., Part
  XXXVI}.\hskip 1em plus 0.5em minus 0.4em\relax Springer, 2022, pp. 540--557.

\bibitem{senanayake2017bayesian}
R.~Senanayake and F.~Ramos, ``Bayesian hilbert maps for dynamic continuous
  occupancy mapping,'' in \emph{Conf. on Robot Learning}.\hskip 1em plus 0.5em
  minus 0.4em\relax PMLR, 2017, pp. 458--471.

\bibitem{zhi2019continuous}
W.~Zhi, L.~Ott, R.~Senanayake, and F.~Ramos, ``Continuous occupancy map fusion
  with fast bayesian hilbert maps,'' in \emph{2019 Intl. Conf. on Robotics and
  Automation (ICRA)}.\hskip 1em plus 0.5em minus 0.4em\relax IEEE, 2019, pp.
  4111--4117.

\bibitem{park2019deepsdf}
J.~J. Park, P.~Florence, J.~Straub, R.~Newcombe, and S.~Lovegrove, ``Deepsdf:
  Learning continuous signed distance functions for shape representation,'' in
  \emph{Proc. of the IEEE/CVF conf. on computer vision and pattern
  recognition}, 2019, pp. 165--174.

\bibitem{mescheder2019occupancy}
L.~Mescheder, M.~Oechsle, M.~Niemeyer, S.~Nowozin, and A.~Geiger, ``Occupancy
  networks: Learning 3d reconstruction in function space,'' in \emph{Proc. of
  the IEEE/CVF conf. on computer vision and pattern recognition}, 2019, pp.
  4460--4470.

\bibitem{chabra2020deep}
R.~Chabra, J.~E. Lenssen, E.~Ilg, T.~Schmidt, J.~Straub, S.~Lovegrove, and
  R.~Newcombe, ``Deep local shapes: Learning local sdf priors for detailed 3d
  reconstruction,'' in \emph{Computer Vision--ECCV 2020: 16th European Conf.,
  Glasgow, UK, August 23--28, 2020, Proc., Part XXIX 16}.\hskip 1em plus 0.5em
  minus 0.4em\relax Springer, 2020, pp. 608--625.

\bibitem{jiang2020local}
C.~Jiang, A.~Sud, A.~Makadia, J.~Huang, M.~Nie{\ss}ner, T.~Funkhouser
  \emph{et~al.}, ``Local implicit grid representations for 3d scenes,'' in
  \emph{Proc. of the IEEE/CVF Conf. on Computer Vision and Pattern
  Recognition}, 2020, pp. 6001--6010.

\bibitem{peng2020convolutional}
S.~Peng, M.~Niemeyer, L.~Mescheder, M.~Pollefeys, and A.~Geiger,
  ``Convolutional occupancy networks,'' in \emph{Computer Vision--ECCV 2020:
  16th European Conf., Glasgow, UK, August 23--28, 2020, Proc., Part III
  16}.\hskip 1em plus 0.5em minus 0.4em\relax Springer, 2020, pp. 523--540.

\bibitem{lionar2021neuralblox}
S.~Lionar, L.~Schmid, C.~Cadena, R.~Siegwart, and A.~Cramariuc, ``Neuralblox:
  Real-time neural representation fusion for robust volumetric mapping,'' in
  \emph{2021 Intl. Conf. on 3D Vision (3DV)}.\hskip 1em plus 0.5em minus
  0.4em\relax IEEE, 2021, pp. 1279--1289.

\bibitem{deng2022neuralef}
Z.~Deng, J.~Shi, and J.~Zhu, ``Neuralef: Deconstructing kernels by deep neural
  networks,'' in \emph{International Conference on Machine Learning}.\hskip 1em
  plus 0.5em minus 0.4em\relax PMLR, 2022, pp. 4976--4992.

\bibitem{rahimi2007random}
A.~Rahimi and B.~Recht, ``Random features for large-scale kernel machines,''
  \emph{Advances in neural information processing systems}, vol.~20, 2007.

\bibitem{rahimi2008weighted}
------, ``Weighted sums of random kitchen sinks: Replacing minimization with
  randomization in learning,'' \emph{Advances in neural information processing
  systems}, vol.~21, 2008.

\bibitem{yu2016orthogonal}
F.~X.~X. Yu, A.~T. Suresh, K.~M. Choromanski, D.~N. Holtmann-Rice, and
  S.~Kumar, ``Orthogonal random features,'' \emph{Advances in neural
  information processing systems}, vol.~29, 2016.

\bibitem{munkhoeva2018quadrature}
M.~Munkhoeva, Y.~Kapushev, E.~Burnaev, and I.~Oseledets, ``Quadrature-based
  features for kernel approximation,'' \emph{Advances in neural information
  processing systems}, vol.~31, 2018.

\bibitem{francis2021major}
D.~P. Francis and K.~Raimond, ``Major advancements in kernel function
  approximation,'' \emph{Artificial Intelligence Review}, vol.~54, no.~2, pp.
  843--876, 2021.

\bibitem{williams2000using}
C.~Williams and M.~Seeger, ``Using the nystr{\"o}m method to speed up kernel
  machines,'' \emph{Advances in neural information processing systems},
  vol.~13, 2000.

\bibitem{yang2012nystrom}
T.~Yang, Y.-F. Li, M.~Mahdavi, R.~Jin, and Z.-H. Zhou, ``Nystr{\"o}m method vs
  random fourier features: A theoretical and empirical comparison,''
  \emph{Advances in neural information processing systems}, vol.~25, 2012.

\bibitem{williams2006gaussian}
C.~K. Williams and C.~E. Rasmussen, \emph{Gaussian processes for machine
  learning}.\hskip 1em plus 0.5em minus 0.4em\relax MIT press Cambridge, MA,
  2006, vol.~2, no.~3.

\bibitem{genton2001classes}
M.~G. Genton, ``Classes of kernels for machine learning: a statistics
  perspective,'' \emph{Journal of machine learning research}, vol.~2, no. Dec,
  pp. 299--312, 2001.

\bibitem{solak2002derivative}
E.~Solak, R.~Murray-Smith, W.~Leithead, D.~Leith, and C.~Rasmussen,
  ``Derivative observations in gaussian process models of dynamic systems,''
  \emph{Advances in neural information processing systems}, vol.~15, 2002.

\bibitem{park2017colored}
J.~Park, Q.-Y. Zhou, and V.~Koltun, ``Colored point cloud registration
  revisited,'' in \emph{Proc. of the IEEE intl. conf. on computer vision},
  2017, pp. 143--152.

\bibitem{peng2022openscene}
S.~Peng, K.~Genova, C.~Jiang, A.~Tagliasacchi, M.~Pollefeys, T.~Funkhouser
  \emph{et~al.}, ``Openscene: 3d scene understanding with open vocabularies,''
  \emph{arXiv preprint arXiv:2211.15654}, 2022.

\bibitem{shafiullah2022clip}
N.~M.~M. Shafiullah, C.~Paxton, L.~Pinto, S.~Chintala, and A.~Szlam,
  ``Clip-fields: Weakly supervised semantic fields for robotic memory,''
  \emph{arXiv preprint arXiv:2210.05663}, 2022.

\bibitem{huang2023visual}
C.~Huang, O.~Mees, A.~Zeng, and W.~Burgard, ``Visual language maps for robot
  navigation,'' in \emph{ICRA}, 2023.

\bibitem{dai2017scannet}
A.~Dai, A.~X. Chang, M.~Savva, M.~Halber, T.~Funkhouser, and M.~Nie{\ss}ner,
  ``Scannet: Richly-annotated 3d reconstructions of indoor scenes,'' in
  \emph{Proc. of the IEEE conf. on computer vision and pattern recognition},
  2017, pp. 5828--5839.

\bibitem{sturm2012benchmark}
J.~Sturm, N.~Engelhard, F.~Endres, W.~Burgard, and D.~Cremers, ``A benchmark
  for the evaluation of rgb-d slam systems,'' in \emph{2012 IEEE/RSJ intl.
  conf. on intelligent robots and systems}.\hskip 1em plus 0.5em minus
  0.4em\relax IEEE, 2012, pp. 573--580.

\bibitem{zhang2018multi}
H.~Zhang and K.~Dana, ``Multi-style generative network for real-time
  transfer,'' in \emph{Proc. of the European Conf. on Computer Vision (ECCV)
  Workshops}, 2018, pp. 0--0.

\bibitem{qi2017pointnet++}
C.~R. Qi, L.~Yi, H.~Su, and L.~J. Guibas, ``Pointnet++: Deep hierarchical
  feature learning on point sets in a metric space,'' \emph{Advances in neural
  information processing systems}, vol.~30, 2017.

\bibitem{armeni20163d}
I.~Armeni, O.~Sener, A.~R. Zamir, H.~Jiang, I.~Brilakis, M.~Fischer, and
  S.~Savarese, ``3d semantic parsing of large-scale indoor spaces,'' in
  \emph{Proc. of the IEEE conf. on computer vision and pattern recognition},
  2016, pp. 1534--1543.

\bibitem{occipital}
Occipital, ``Occipital: The structure sensor,'' 2016.

\bibitem{armeni2017joint}
I.~Armeni, S.~Sax, A.~R. Zamir, and S.~Savarese, ``Joint 2d-3d-semantic data
  for indoor scene understanding,'' \emph{arXiv preprint arXiv:1702.01105},
  2017.

\bibitem{rosinol2023probabilistic}
A.~Rosinol, J.~J. Leonard, and L.~Carlone, ``Probabilistic volumetric fusion
  for dense monocular slam,'' in \emph{Proc. of the IEEE/CVF Winter Conf. on
  Applications of Computer Vision}, 2023, pp. 3097--3105.

\bibitem{cheraghian2019zero}
A.~Cheraghian, S.~Rahman, and L.~Petersson, ``Zero-shot learning of 3d point
  cloud objects,'' in \emph{2019 16th Intl. Conf. on Machine Vision
  Applications (MVA)}.\hskip 1em plus 0.5em minus 0.4em\relax IEEE, 2019, pp.
  1--6.

\bibitem{frome2013devise}
A.~Frome, G.~S. Corrado, J.~Shlens, S.~Bengio, J.~Dean, M.~Ranzato, and
  T.~Mikolov, ``Devise: A deep visual-semantic embedding model,''
  \emph{Advances in neural information processing systems}, vol.~26, 2013.

\bibitem{michele2021generative}
B.~Michele, A.~Boulch, G.~Puy, M.~Bucher, and R.~Marlet, ``Generative zero-shot
  learning for semantic segmentation of 3d point clouds,'' in \emph{2021 Intl.
  Conf. on 3D Vision (3DV)}.\hskip 1em plus 0.5em minus 0.4em\relax IEEE, 2021,
  pp. 992--1002.

\bibitem{zhou2018open3d}
Q.-Y. Zhou, J.~Park, and V.~Koltun, ``Open3d: A modern library for 3d data
  processing,'' \emph{arXiv preprint arXiv:1801.09847}, 2018.

\bibitem{schops2019bad}
T.~Schops, T.~Sattler, and M.~Pollefeys, ``Bad slam: Bundle adjusted direct
  rgb-d slam,'' in \emph{Proc. of the IEEE/CVF Conf. on Computer Vision and
  Pattern Recognition}, 2019, pp. 134--144.

\bibitem{whelan2012kintinuous}
T.~Whelan, M.~Kaess, M.~Fallon, H.~Johannsson, J.~Leonard, and J.~McDonald,
  ``Kintinuous: Spatially extended kinectfusion,'' 2012.

\bibitem{mur2017orb}
R.~Mur-Artal and J.~D. Tard{\'o}s, ``Orb-slam2: An open-source slam system for
  monocular, stereo, and rgb-d cameras,'' \emph{IEEE transactions on robotics},
  vol.~33, no.~5, pp. 1255--1262, 2017.

\bibitem{teed2021droid}
Z.~Teed and J.~Deng, ``Droid-slam: Deep visual slam for monocular, stereo, and
  rgb-d cameras,'' \emph{Advances in neural information processing systems},
  vol.~34, pp. 16\,558--16\,569, 2021.

\bibitem{muller2022instant}
T.~M{\"u}ller, A.~Evans, C.~Schied, and A.~Keller, ``Instant neural graphics
  primitives with a multiresolution hash encoding,'' \emph{ACM Transactions on
  Graphics (ToG)}, vol.~41, no.~4, pp. 1--15, 2022.

\bibitem{kim2022revisiting}
T.~Kim, K.~Kim, J.~Lee, D.~Cha, J.~Lee, and D.~Kim, ``Revisiting image pyramid
  structure for high resolution salient object detection,'' in \emph{Proc. of
  the Asian Conf. on Computer Vision}, 2022, pp. 108--124.

\end{thebibliography}
\begin{IEEEbiography}[{\includegraphics[width=1in,height=1.25in,clip,keepaspectratio]{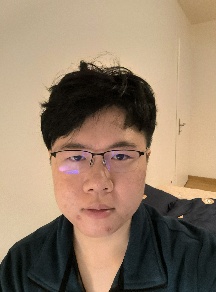}}]{Yijun Yuan} is a third year PhD student in Prof. Andreas N\"uchter's group at Julius Maximilian University Würzburg. He received Bachelor's and Master's Degree from ShanghaiTech university in 2018 and 2021.
Yijun has experience on Rescue Robotics and Metrical\&Topological Mapping and in PhD studies he focuses on dense SLAM methods using various sensors.
\end{IEEEbiography}

\begin{IEEEbiography}[{\includegraphics[width=1in,height=1.25in,clip,keepaspectratio]{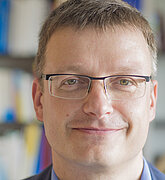}}]{Andreas N\"uchter} is professor and chair of robotics at University of Würzburg. Before fall 2022 he was associated professor (tenured) for telematics at University of Würzburg and before summer 2013 he headed as assistant professor the Automation group at Jacobs University Bremen. Prior he was a research associate at University of Osnabrück. Further past affiliations were with the Fraunhofer Institute for Autonomous Intelligent Systems (AIS, Sankt Augustin), the University of Bonn, from which he received the diploma degree in computer science in 2002 and the Washington State University. He holds a doctorate degree (Dr. rer. nat) from University of Bonn. Andreas works on robotics and automation, cognitive systems and artificial intelligence. His main research interests include reliable robot control, 3D environment mapping, 3D vision, and laser scanning technologies for various applications, e.g. for planetary exploration, safety security and rescue robotics, or underwater inspection. Andreas developed fast 3D scan matching algorithms that enable robots to perceive and map their environment in 3D representing the pose with 6 degrees of freedom. The capabilities of these robotic SLAM approaches were demonstrated at RoboCup Rescue competitions, ELROB and several other events. He is a member of the GI and the IEEE.\end{IEEEbiography}

\end{document}